\documentclass[preprint,12pt]{elsarticle}

\usepackage[T1]{fontenc}
\usepackage[utf8]{inputenc}

\usepackage{amsmath,amssymb,amsfonts}
\usepackage{amsthm}

\usepackage{graphicx}
\usepackage{subcaption}
\usepackage{booktabs}
\usepackage{multirow}
\usepackage{float}
\usepackage[section]{placeins}

\usepackage{algorithm}
\usepackage{algpseudocode}

\usepackage{microtype}
\usepackage{xurl}
\usepackage[colorlinks=true,allcolors=blue]{hyperref}
\usepackage[capitalise,nameinlink,noabbrev]{cleveref}

\biboptions{numbers,sort&compress}

\theoremstyle{plain}

\theoremstyle{remark}

\journal{Journal of Computational Physics}

\makeatletter
\def\ps@pprintTitle{%
  \let\@oddhead\@empty
  \let\@evenhead\@empty
  \def\@oddfoot{\reset@font\hfil\thepage\hfil}%
  \let\@evenfoot\@oddfoot
}
\makeatother

\begin{document}
\begin{frontmatter}

\title{A Probabilistic Framework for Solving High-Frequency Helmholtz Equations via Diffusion Models}

\author[mems]{Yicheng Zou}
\ead{yz886@duke.edu}

\author[vienna]{Samuel Lanthaler}
\ead{samuel.lanthaler@univie.ac.at}

\author[cee,mems]{Hossein Salahshoor\corref{cor1}}
\ead{hossein.salahshoor@duke.edu}
\cortext[cor1]{Corresponding author.}

\affiliation[cee]{organization={Department of Civil and Environmental Engineering},
    addressline={Duke University},
    city={Durham},
    state={NC},
    country={USA}}

\affiliation[mems]{organization={Department of Mechanical Engineering and Materials Science},
    addressline={Duke University},
    city={Durham},
    state={NC},
    country={USA}}

\affiliation[vienna]{organization={Department of Mathematics},
  addressline={University of Vienna},
  city={Vienna},
  country={Austria}}

\begin{abstract}
Deterministic neural operators perform well on many PDEs but can struggle with the approximation of high-frequency wave phenomena, where strong input-to-output sensitivity makes operator learning challenging, and spectral bias blurs oscillations. We argue for adopting a probabilistic approach for approximating waves in high-frequency regime, and develop our probabilistic framework using a score-based conditional diffusion operator. After demonstrating a stability analysis of the Helmholtz operator, we present our numerical experiments across a wide range of frequencies, benchmarked against other popular data-driven and machine learning approaches for waves. We show that our probabilistic neural operator consistently produces robust predictions with the lowest errors in $L^2$, $H^1$, and energy norms. Moreover, unlike all the other tested deterministic approaches, our framework remarkably captures uncertainties in the input sound speed map propagated to the solution field. We envision that our results position probabilistic operator learning as a principled and effective approach for solving complex PDEs such as Helmholtz in the challenging high-frequency regime.
\end{abstract}


\end{frontmatter}

\section{Introduction}
Helmholtz equation is a class of elliptic PDEs that arise in modeling time-harmonic wave propagation with a wide range of applications in applied sciences from geophysics to medical fields such as imaging and therapeutic via ultrasound \citep{amundsen2023_viscoacoustic_greens,huttunen_full_wave_helmholtz,sarvazyan_radiation_force_review,rybyanets_hifu_transducers,salahshoor_shear_waves_tFUS,juraev_helmholtz_applications}. Solving Helmholtz equation in high-frequency regimes in heterogeneous media require very fine computational grids that often renders the problem as computationally prohibitive \citep{chen2025_iterative_methods_helmholtz,bootland2021_coarse_spaces_helmholtz,bao_numerical_solution_high_wavenumbers,Cheng2026MNO,Hu2025DeepOMamba,Cho2026MBNO,Zhao2025LESnets}. This, in turn, motivates learning-based surrogates that remarkably reduce computational cost via fast inference while aiming to preserve fidelity. Among these, \emph{operator learning} targets mappings between function spaces rather than pointwise inputs and outputs, with representative approaches including DeepONet and the Fourier Neural Operator (FNO), as well as physics-tailored extensions for wave physics and elasticity \citep{lu2021deeponet,li2021fno,zhang2023fnoElastic,lehmann2024ffno3d,zou2024hno,chen2024nsno,you2024mscalefno}. However, high-frequency wavefields in heterogeneous media expose two persistent limitations of \emph{deterministic} operators: (i) \emph{operator spectral bias}, wherein models preferentially capture low-frequency content and oversmooth oscillatory structure, degrading phase accuracy and interference patterns critical to acoustics and seismics \citep{fanaskov2023_spectral_neural_operators,khodakarami2025_hfs_mitigate_spectral_bias,xu2025_overview_frequency_principle}; and (ii) \emph{sensitivity} to small perturbations in inputs (e.g., sound speed, geometry, or frequency), which can induce multi-modality or sharply varying responses that point-estimate predictors neither represent nor calibrate \citep{ivanovs2021_perturbation_survey,le2024_math_analysis_neural_operators,behroozi2025_sc_fno}. 

Probabilistic learning offers an alternative to deterministic (single-output) surrogates by modeling a distribution over solutions, thereby capturing multi-modality and input sensitivity \citep{stanziola2021_unsupervised_helmholtz,vertes2020_probabilistic_learning,wu2023_pinn_quadratic_pml,alkhalifah2022_wavefield_ml,lobato2023_learned_priors_helmholtz}. Conditional diffusion model enable a concrete tool to examine probabilistic operator learning \citep{shysheya2024cdmPDE,huang2024diffusionpde,bastek2025pidm,lim2023functionspace,wang2024fundiff,yao2025guidedfs}, and we aim to employ it for high-frequency acoustics governed by the Helmholtz equation. We instantiate a \emph{conditional diffusion operator} that maps problem inputs (sound-speed map, source mask, positional encodings) to the complex frequency-domain wavefield and evaluate it across low to high frequencies, comparing against strong deterministic baselines (FNO, HNO, and a backbone-matched U-Net). Our study revolves around two questions central to high-frequency prediction: (Q1) can a \emph{probabilistic} operator preserve \emph{stable} quantities—e.g., energies in sub-regions—more reliably than deterministic surrogates; and (Q2) can it produce \emph{calibrated} uncertainty that faithfully reflects sensitivity to input perturbations? Although the underlying forward map is deterministic, practical indeterminacy at high frequencies makes single-output surrogates brittle. Our probabilistic formulation though remarkably learns a conditional distribution over solutions, using its mean for accuracy and its dispersion to encode epistemic sensitivity.

In this paper, we use conditional diffusion models as probabilistic learning machines to: (i) learn fine-features in solutions fields of Helmholtz equation which are hard to capture, if at all possible, by any deterministic model, and (ii) to learn the push-forward of uncertainties in coefficient fields, reflected in our sensitivity study. Across all tested frequencies, diffusion achieves the lowest errors in $L^2$, $H^1$, and energy norm. We also demonstrate that diffusion model can predict the statistics of amplitudes in the far-field. Our sensitivity analysis further show that diffusion robustly mirrors the ground-truth variability and produces calibrated uncertainty, whereas deterministic operators are systematically under-dispersed and miss small-variance modes. Taken together, our results show that probabilistic operator learning is a promising approach for approximating high-frequency solutions of PDEs. 

\section{Related Work}
\vspace{-5pt}

\paragraph{Operator learning for Helmholtz}
Neural operators promise fast surrogates for PDEs at scale, but Helmholtz problems expose their core weakness: preserving high-frequency structure. Fourier Neural Operators can efficiently model elastic waves at low frequencies and tend to smooth oscillations—an instance of spectral bias \citep{zhang2023fnoElastic,lehmann2024ffno3d,lara2024_ood_bounds_neural_operators}. To alleviate this spectral bias, several Helmholtz-specific designs have emerged. Helmholtz Neural Operators (HNO) is one important example which operates in the frequency domain, exchanging time stepping for per-frequency solves and delivering substantial memory/runtime benefits while retaining accuracy on elastic wavefields \citep{zou2024_deep_helmholtz_operators_3d,zou2024hno}. Complementary efforts build analytic structure into the surrogate: Neumann–series neural operators target large wavenumbers to improve stability and error \citep{chen2024nsno}, while multi-scale architectures inject resolution where spectral bias hurts most \citep{you2024mscalefno}. We have utilized HNO in our investigations as a benchmark to compare our results with.

\paragraph{Diffusion models for PDE solution fields}
Diffusion models approach the problem from a different angle: rather than a single point estimate, they learn a distribution over solutions. Many efforts have aimed at developing physics-informed diffusion by introducing residual penalties and priors in training/sampling so that the generated fields satisfy governing equations \citep{bastek2025pidm,zhou2025_text2pde,zhang2025_piddm}. Conditional diffusion, in particular, has shown strong fidelity on approximating solutions of PDEs \citep{shysheya2024cdmPDE, haitsiukevich2024dmno,hu2025_safediffcon}. Several studies have also investigated the partial observations scenarios, aiming to infer from sparse or indirect data \citep{huang2024diffusionpde,cao2025_diffusion_pde_perspective,gao2024_pde_diffusion_iclr,gao2025_guided_diffusion_virtual_observations,bergamin2025_guided_autoregressive_diffusion_pde}. Moreover, the concept of statistical learning in turbulence  \cite{foias2001navier} has rendered probabilistic approaches such as conditional diffusion models 
a promising pathway for modeling turbulent fluids, which has recently been investigated \citep{oommen2025noDiffTurbulence,molinaro2024_gencfd}. There has also been a wide range of efforts on function-space diffusion models defining priors directly on continuous fields to achieve discretization-agnostic learning and transfer across resolutions \citep{pidstrigach2023infinite, lim2023functionspace,yao2025guidedfs,bastek2025_physics_informed_diffusion}.

\section{Theoretical Framework}\label{sec:background}

This section is devoted to the underlying theory for our framework. After describing the general setting, we focus on the Helmholtz problem, and delineate the input-sensitivity of the Helmholtz solution in the high-frequency regime. This, in turn, motivates our adoption of probabilistic learning via conditional score-based diffusion, which we subsequently describe in the last subsection. 
\subsection{Probabilistic operator learning for PDEs}
Consider a boundary–value problem on an open, connected domain \(\Omega\subset\mathbb{R}^d\) with boundary \(\partial\Omega\),
\begin{equation}
\label{eq:pde-general}
\mathcal{L}_{\kappa}\,u \;=\; f \quad \text{in }\Omega,
\qquad
\mathcal{B}\,u \;=\; g \quad \text{on }\partial\Omega,
\end{equation}
where \(\mathcal{L}_{\kappa}\) is a differential operator depending on coefficient fields \(\kappa\) (e.g., sound speed, density), \(f\) is a source term, and \(g\) encodes boundary data. Under standard well-posedness assumptions, \eqref{eq:pde-general} induces a \emph{solution operator}
\begin{equation}
\label{eq:solution-operator}
\mathcal{S}:\; \mathcal{Z}\to\mathcal{Y}, 
\qquad 
z:=(\kappa,f,g)\in\mathcal{Z} \;\mapsto\; u=\mathcal{S}(z)\in\mathcal{Y},
\end{equation}
for suitable function spaces (e.g., \(\mathcal{Z}\subset L^\infty\times L^2\times H^{1/2}\), \(\mathcal{Y}\subset H^1\)). Classical solvers approximate \(\mathcal{S}\) \emph{per query} by discretizing \eqref{eq:pde-general} and computing \(u_h\approx u\), which is expensive in many-query regimes (varying \(z\) across geometries, coefficients, and frequencies). \emph{Operator learning} seeks a data-driven surrogate \(\widehat{\mathcal{S}}_{\theta}:\mathcal{Z}\to\mathcal{Y}\) trained from pairs \(\{(z_i,u_i)\}_{i=1}^N\) with \(u_i=\mathcal{S}(z_i)\), solving
\begin{equation}
\begin{aligned}
\theta^\star &\in \arg\min_{\theta}\; 
\frac{1}{N}\sum_{i=1}^N 
\mathcal{L}\!\bigl(\widehat{\mathcal{S}}_{\theta}(z_i),\,u_i\bigr),\\
&\mathcal{L}(a,b) := \|a-b\|_{\mathcal{Y}},
\end{aligned}
\end{equation}
where $\| \cdot \|_\mathcal{Y}$ is the norm in the solution space. Popular \emph{deterministic} architectures (e.g., Fourier Neural Operators (FNOs) \citep{li2021fno}, DeepONets \citep{lu2021deeponet}) realize \(\widehat{\mathcal{S}}_{\theta}\) via learned integral kernels or spectral multipliers and return a \emph{single} prediction \(\widehat{u}_{\theta}=\widehat{\mathcal{S}}_{\theta}(z)\) for input \(z\). 

A \emph{probabilistic} operator instead models the \emph{conditional law} of the solution: for each \(z\in\mathcal{Z}\), it represents a conditional distribution \(\mathbb{P}(u\mid z)\) and enables sampling \(u\sim p_{\vartheta}(\cdot\mid z)\). We parameterize a conditional generator \(T_{\vartheta}\) and draw samples using a latent Gaussian \(\eta\sim\mathcal{N}(0,I)\),
\begin{equation}
\label{eq:generator}
u \;=\; T_{\vartheta}(\eta;z) \;\sim\; p_{\vartheta}(\,\cdot\,\mid z),
\end{equation}
where \(p_\vartheta(\,\cdot\,\mid z)=T_{\vartheta,\#}\mathcal{N}(0,I)\) is the push-forward of the reference Gaussian under the learned generator. The specific choice of \(T_{\vartheta}\) and its training objective depends on the probabilistic family, and in this paper we focus on conditional diffusion models, while other instantiations include conditional normalizing flows and function-space priors.
\citep{lim2023functionspace,wang2024fundiff,yao2025guidedfs,shysheya2024cdmPDE,huang2024diffusionpde, chen2025scale}.

\subsection{High-frequency Helmholtz and why probabilistic helps}
We consider the Helmholtz equation 
\begin{equation}
   c^2 \nabla^2 u + k^2 u = 0, 
   \label{eq:helmholtz}
\end{equation}
with inhomogeneous sound map $c = c(x)$. The Helmholtz equation underpins time-harmonic wave modeling in acoustics, elasticity, and scattering. Despite its ubiquity, its numerical solution is still facing many challenges. The reader is referred to \cite{ernst2011difficult} for a detailed description of the challenges in solving Helmholtz equation. In a nutshell, Helmholtz operator is neither Hermitian symmetric nor coercive, and is poorly conditioned, which makes it hard to solve via iterative methods. With these difficulties in place, in the high wavenumber regime additionally requires resolving short wavelengths and controlling dispersion (''pollution``) errors, which, collectively renders approximating high frequency solutions of Helmholtz a difficult problem. These difficulties have motivated studies on using neural operators for Helmholtz \citep{zhang2023fnoElastic,lehmann2024ffno3d,lara2024_ood_bounds_neural_operators,zou2024_deep_helmholtz_operators_3d,zou2024hno}.

Two obstacles for deterministic operators are aggravated in the high-frequency regime of the Helmholtz equation: (i) \emph{spectral bias}, where neural networks preferentially fit low-frequency content and oversmooth oscillations, degrading phase/interference fidelity; and (ii) \emph{input sensitivity}, where tiny coefficient perturbations induce large phase shifts in the solution. 

Let \(S:\mathcal{Z}\!\to\!\mathcal{Y}\) map a sound-speed field \(c\) to the complex wavefield \(u=S(c)\). Linearizing around a baseline \(c_0\) gives with recourse to Fr\'echet differentiability and using the operator norm of the differential:
\begin{equation}
\delta u \;\approx\; DS[c_0]\,(\delta c),\qquad
L(c_0)\;:=\;\|DS[c_0]\|_{\mathrm{op}},\qquad
\text{so}\;\; \frac{\|\delta u\|}{\|u_0\|}\;\lesssim\; L(c_0)\,\frac{\|\delta c\|}{\|c_0\|},
\end{equation}
with \(u_0:=S(c_0)\), where $L(c_0)$ is the Lipchitz constant. The growth of \(L(c_0)\) with frequency can be made explicit by a standard geometric-optics, also known as WKB approximation, argument. Let us examine a one-dimensional model, since it carries the key ideas with least complexities.  

\paragraph{Helmholtz equation in dimension one}
Consider the 1D variable-coefficient Helmholtz equation
\begin{equation}
u''(x)+k^2 n(x)^2u(x)=0,\qquad n(x):=\frac{1}{c(x)},\quad x\in[0,\ell],
\end{equation}
in the high-frequency regime \(k\gg 1\). Using the WKB ansatz \(u(x)=a(x)e^{ik\phi(x)}\) and retaining the leading-order phase (i.e. $k^2$ terms), one obtains the travel time
\(\tau(x)=\int_0^x \!ds/c(s)\) and the leading approximation
\begin{equation}
u(x)\;\approx\;A_0\,\exp\!\Big(\pm i k \!\int_{0}^{x}\!\frac{ds}{c(s)}\Big).
\label{eq:wkb_1d_field}
\end{equation}
Now perturb \(c=c_0+\delta c\) with \(\|\delta c\|_\infty/\|c_0\|_\infty\ll 1\). Since \(n=1/c\), we have \(\delta n(x)= -\delta c(x)/c_0(x)^2+\mathcal{O}(\delta c^2)\), hence the travel-time perturbation satisfies
\begin{equation}
\delta\tau(x)\;=\;\int_0^x \delta n(s)\,ds
\;=\;-\int_0^x \frac{\delta c(s)}{c_0(s)^2}\,ds+\mathcal{O}(\delta c^2).
\end{equation}
Substituting into \eqref{eq:wkb_1d_field} (and neglecting subleading amplitude changes) yields a phase-dominated relative error,
\begin{equation}
\frac{\delta u(x)}{u_0(x)}
\;\approx\; \exp\!\big(\pm i k\,\delta\tau(x)\big)-1,
\qquad
\Big|\tfrac{\delta u(x)}{u_0(x)}\Big|
\;\lesssim\;
k \int_0^{x}\frac{|\delta c(s)|}{c_0(s)^2}\,ds.
\label{eq:wkb_relerr_general}
\end{equation}
If \(c_0\) is approximately constant along the path (or locally bounded below by \(c_{0,\min}\)), this simplifies to
\begin{equation}
\Big|\tfrac{\delta u(x)}{u_0(x)}\Big|
\;\lesssim\;
\underbrace{\frac{k\,x}{c_0}}_{=:L(k,x,c_0)}\,\frac{\|\delta c\|_\infty}{c_0}
\;\le\;
\underbrace{\frac{k\,\ell}{c_0}}_{=:L(k,\ell,c_0)}\,\frac{\|\delta c\|_\infty}{c_0}.
\label{eq:wkb_1d_bound}
\end{equation}
Thus, even tiny coefficient errors can produce order-one wavefield errors once \(k\ell\gg 1\), i.e., after sufficient phase accumulation.

\paragraph{Extension to higher dimensions}
The same mechanism carries to \(d\ge 2\): under geometric-optics assumptions (single dominant ray, no caustics), the solution at a receiver \(\boldsymbol{r}\) admits the ray form
\begin{equation}
u(\boldsymbol{r}) \;\approx\; A(\boldsymbol{r})\,
\exp\!\Big(\pm i k \!\int_{\gamma(\boldsymbol{r})}\!\frac{ds}{c(s)}\Big),
\label{eq:ray_form_hd}
\end{equation}
where \(\gamma(\boldsymbol{r})\) is the ray from the source to \(\boldsymbol{r}\) with length \(\ell(\boldsymbol{r})\). A perturbation \(c=c_0+\delta c\) induces an accumulated phase shift
\begin{equation}
\delta\phi(\boldsymbol{r})
\;\approx\;
-\,k\!\int_{\gamma(\boldsymbol{r})}\!\frac{\delta c(s)}{c_0(s)^2}\,ds,
\qquad
\frac{\delta u(\boldsymbol{r})}{u_0(\boldsymbol{r})}
\;\approx\;
e^{\,\pm i\,\delta\phi(\boldsymbol{r})}-1,
\label{eq:phase_shift_hd}
\end{equation}
so the effective sensitivity grows with \(k\) and the travel distance \(\ell(\boldsymbol{r})\), and is smallest near the source---precisely the scaling formalized in Eqs.~(6)--(7).

If the induced phase perturbation \(\delta\phi(\boldsymbol{r})\) can be treated as random (reflecting unresolved input variability or modeling error), then the MSE-optimal \emph{deterministic} predictor collapses oscillations by vector-averaging on the unit circle:
\begin{equation}
\hat u_{\mathrm{MSE}}(\boldsymbol{r})
=\mathbb{E}[u(\boldsymbol{r})\,|\,z]
\;\approx\;
u_0(\boldsymbol{r})\,\mathbb{E}\!\big[e^{\,i\delta\phi(\boldsymbol{r})}\big]
\;\approx\;
u_0(\boldsymbol{r})\,e^{-\tfrac{1}{2}\mathrm{Var}[\delta\phi(\boldsymbol{r})]}.
\end{equation}
The exponential attenuation produces amplitude shrinkage, blurred interference, and loss of high-frequency contrast—exactly when the sensitivity factor scales like \(L(k,\ell)\propto k\ell\).

A \emph{probabilistic} operator instead models the conditional law \(p_{\vartheta}(u\,|\,z)\) by internalizing phase ambiguity and input sensitivity rather than collapsing them into a single output. Drawing samples \(\{u^{(s)}\}_{s=1}^{S}\sim p_{\vartheta}(\cdot\,|\,z)\) allows: (i) preserving oscillations and multi-modality at prediction time (via samples representing uncertain phase modes, not just a learned mean); (ii) quantifying \emph{calibrated uncertainty} that tracks sensitivity; and (iii) reporting \emph{stable} functionals where phase cancels. For example, the energy
\begin{equation}
E(u)\;=\;\int_{\Omega}\!\Big(\|\nabla u(\boldsymbol{r})\|^2+\tfrac{k^2}{c_0(\boldsymbol{r})^{2}}\,|u(\boldsymbol{r})|^2\Big)\,d\boldsymbol{r}
\end{equation}
can be estimated in a Bayes-optimal way for energy risk by \(\hat E=\mathbb{E}[E(u)\,|\,z]\approx \tfrac{1}{S}\sum_{s=1}^{S}E(u^{(s)})\), which remains stable even when \(\mathbb{E}[u\,|\,z]\) is attenuated by phase dispersion. Practically, \(p_{\vartheta}(u\,|\,z)\) lets us decouple \emph{fragile} quantities (phase, pointwise fields) from \emph{stable} ones (energy, band-limited summaries), deliver coverage diagnostics for sensitivity-aware calibration, and choose decision rules matched to the evaluation metric—yielding robustness exactly where deterministic surrogates struggle.

\subsection{Conditional diffusion model for probabilistic learning}
In this subsection, we describe our adopted conditional diffusion models for probabilistic learning of wavefields. Let us denote the Helmholtz solution as $u_0$, our objective is to approximate the conditional law $p_\theta(u_0\mid z)$ given the PDE inputs, where $u_0\in\mathbb{C}^{H\times W}$ and $z=[\,c,\,m,\,\mathrm{PE}_x,\,\mathrm{PE}_y\,]$ concatenates the sound-speed map $c$, a binary source mask $m$, and sinusoidal positional encodings (we are assuming a 2D square domain for the sake of simplicity). This is achieved through \emph{noising} and \emph{denoising} process. For the first step, a \emph{forward} noising chain progressively corrupts $u_0$ to $u_T$ as
\begin{equation}
q(u_t\mid u_{t-1})=\mathcal{N}\!\bigl(u_t;\sqrt{\alpha_t}\,u_{t-1},(1-\alpha_t)I\bigr),\quad
q(u_t\mid u_0)=\mathcal{N}\!\bigl(u_t;\sqrt{\bar\alpha_t}\,u_0,(1-\bar\alpha_t)I\bigr),
\end{equation}
with schedule $\{\beta_t\}_{t=1}^T\subset(0,1)$, $\alpha_t=1-\beta_t$, and $\bar\alpha_t=\prod_{s\le t}\alpha_s$. Then, for the next step we use a time-indexed Gaussian family to approximate the \emph{reverse} conditionals
\begin{equation}
\begin{aligned}
p_\theta(u_{t-1}\mid u_t,\mathrm{z})
  &= \mathcal{N}\!\bigl(u_{t-1};\,\mu_\theta(u_t,t,\mathrm{z}),\,\sigma_t^2 I\bigr),\\
\mu_\theta(u_t,t,\mathrm{z})
  &= \frac{1}{\sqrt{\alpha_t}}\!\left(u_t-\frac{\beta_t}{\sqrt{1-\bar\alpha_t}}\,
     \varepsilon_\theta(u_t,t,\mathrm{z})\right),
\end{aligned}
\end{equation}
where a U\mbox{-}Net with time embeddings parametrizes the noise predictor $\varepsilon_\theta$. We note that the conditioning $z$ is \emph{never} noised. The learning objective aligns with conditional denoising score matching is:
\begin{equation}
\label{eq:cond-noise-single}
\mathcal{L}(\theta)=\mathbb{E}_{u_0,z,t,\varepsilon}\Big\|\varepsilon-\varepsilon_\theta\!\big(u_t,t,z\big)\Big\|_2^2,\qquad u_t=\sqrt{\bar\alpha_t}\,u_0+\sqrt{1-\bar\alpha_t}\,\varepsilon,\ \ \varepsilon\sim\mathcal{N}(0,I),
\end{equation}
During the inference, we generate $S$ ancestral samples by iterating $u_T\!\sim\!\mathcal{N}(0,I)$ through $t=T\!\to\!1$ to obtain $\{u_0^{(s)}\}_{s=1}^S\sim p_\theta(\cdot\mid z)$.

\section{Details of datasets and training}
In this section we delineate the details of data generation and diffusion model training. 

\paragraph{Data generation} For a fixed frequency, we synthesize a dataset comprised of 10000 pairs of sound speed maps and Helmholtz solution, i.e. $\{c_j(x), u_j(x)\}_{j=1}^{10000}$. Each $c_j$ is generated using a family of Gaussian random fields (GRFs; see~\ref{app:grf}) where GRF parameters are randomly chosen from a range of amplitudes and correlation lengths. Out of the 10000 synthesized data, we allocate  8190, 1020, and 500 respectively for training, validation, and testing. 

\begin{figure}[H]
  \centering
  \includegraphics[width=\linewidth,height=0.6\textheight,keepaspectratio]{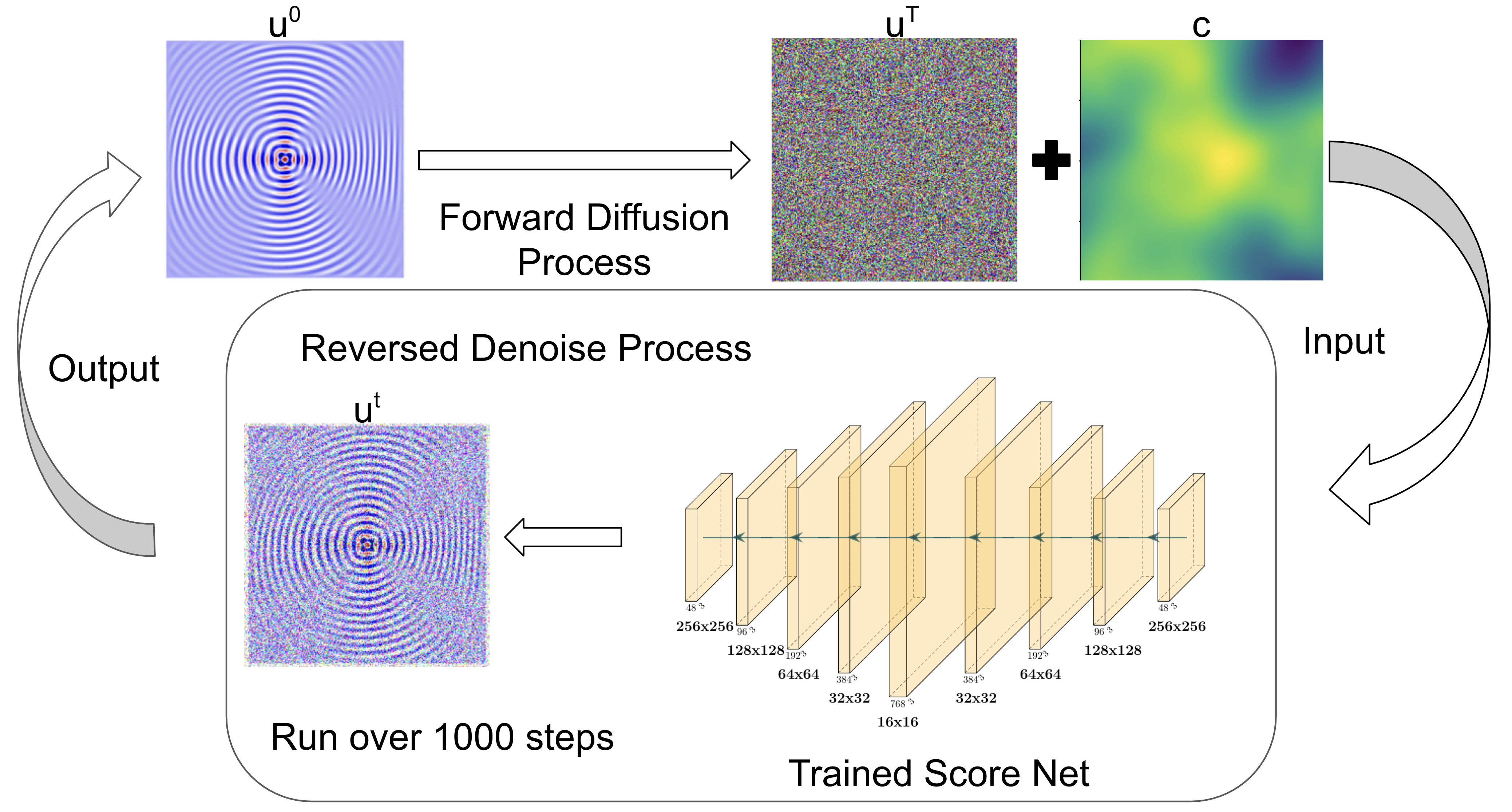}
  \caption{\textbf{Conditional diffusion for Helmholtz.} \emph{Forward diffusion (top):} the wavefield \(u_0\) is progressively noised by a fixed schedule to a Gaussian field \(u_T\). \emph{Reverse denoising (bottom):} sampling starts from pure Gaussian \(u_T\) and is \emph{conditioned} on the inputs $z$ comprised of sound-speed map \(c\), as well as source mask and positional encodings that are not shown. A time-indexed U-Net (“Trained Score Net”) predicts noise and removes it iteratively over \(T\!\approx\!1000\) steps to produce samples \(u_0^{(s)}\) that approximate the conditional distribution of solutions.}
  \label{fig:pipeline}
\end{figure}

The dataset is generated with recourse to the Helmholtz PDE solver \emph{J-Wave}, which is a powerful spectral method based solver developed via JAX \citep{stanziola2022jwave}. In particular, for each sound speed map $c_j$, we  solve the Helmholtz equation:
\begin{equation}
\label{eq:helmholtz-base}
\Bigl(\nabla^2 \;+\; \tfrac{\omega^2}{c_j(x)^2}\Bigr) u_j \;=\; F,
\end{equation}
where $u_j$ is the complex pressure field at angular frequency $\omega=2\pi f$ in a square domain with a $256\times256$ Cartesian grid, $F$ is a masked source term with a disk mask of radius $r_s{=}10$ in terms of the grid points, and perfectly matched layers (PML) is implemented in the boundaries to eliminate the effects of boundary reflections. We repeat the aforementioned to generate six datasets for six different frequencies, ranging from $1.5{\times}10^5$ to $2.5{\times}10^6$~Hz. Without loss of generality, and to reflect a concrete application, both frequencies and sound speed ranges are chosen to reflect modeling ultrasonic waves in tissues. During the learning process, we expand the input to include normalized $c$, source mask, and sinusoidal positional encodings, with the wavefield solutions as target.

\paragraph{Conditional diffusion for Helmholtz operator} Unlike existing application of Diffusion models in time-dependent PDEs, such as PDE\textendash DIFF~\citep{shysheya2024cdmPDE},  we learn a \emph{solution operator} that corresponds to steady state waves, which means there is no temporal stacking and each training example is a single instance at a fixed frequency. We adopt the conditional denoising formulation, where only the target wavefield is noised, while the inputs are kept clean and given as conditioning. The backbone score net is a 2D U\mbox{-}Net with five down/up stages (widths $[48,96,192,384,768]$), three residual blocks per stage, LayerNorm, SiLU, and circular padding (Fig. \ref{fig:pipeline}). A sinusoidal time embedding is passed through an MLP ($128\!\to\!256\!\to\!64$) to produce a 64-D context vector. The conditioning tensor $C$ is used two ways: (i) concatenated to the input image stack, and (ii) mapped (via a linear layer) from the 64-D context to the current block’s channel width to generate FiLM scale/shift $(\gamma,\beta)\in\mathbb{R}^{f}$, applied as $y=\gamma\odot h+\beta$. Lastly, the network predicts a single output channel (wavefield amplitude). See Figure~\ref{fig:pipeline} for an overview of the training/sampling pipeline and the FiLM-conditioned U-Net. For further details, the reader is referred to~\ref{app:methods}).

\section{Results}\label{sec:results}
We organize our experimental results into three parts: (i) \emph{error analysis}, where we quantify the performance of our probabilistic framework and compare it against existing neural-operator baselines; (ii) \emph{sensitivity analysis}, where we examine how uncertainty in the sound-speed field propagates to variability in the predicted wavefields; and (iii) \emph{preliminary 3D results}, where we extend the method to fully 3D Helmholtz problems and evaluate an enhanced diffusion backbone based on a U-shaped vision transformer (UViT). For clarity and worst-case visualization, we present sensitivity results primarily at the highest frequency \(f=2.5\times10^{6}\,\mathrm{Hz}\); additional frequencies and extended panels are provided in the Appendix.

\begin{table*}[t]
\centering
\scriptsize
\setlength{\tabcolsep}{4pt}
\caption{\textbf{Relative errors across frequencies.} Diffusion reports mean$\pm$std over $K{=}10$ samples.}
\label{tab:all_by_freq}
\begin{tabular}{llrrrr}
\toprule
\textbf{Frequency (Hz)} & \textbf{Metric} & \textbf{U-Net} & \textbf{FNO} & \textbf{HNO} & \textbf{Diffusion} \\
\midrule
\multirow{3}{*}{1.5e5}
  & $L^2$   & 0.040 & 0.167 & 0.115 & \textbf{0.027} $\pm$ 0.004 \\
  & $H^1$   & 0.071 & 0.219 & 0.164 & \textbf{0.045} $\pm$ 0.004 \\
  & Energy  & 0.022 & 0.114 & 0.071 & \textbf{0.016} $\pm$ 0.003 \\
\midrule[0.3pt]  
\multirow{3}{*}{2.5e5}
  & $L^2$   & 0.077 & 0.190 & 0.153 & \textbf{0.028} $\pm$ 0.001 \\
  & $H^1$   & 0.121 & 0.278 & 0.219 & \textbf{0.044} $\pm$ 0.001 \\
  & Energy  & 0.026 & 0.086 & 0.075 & \textbf{0.013} $\pm$ 0.002 \\
\midrule[0.3pt]
\multirow{3}{*}{5e5}
  & $L^2$   & 0.083 & 0.176 & 0.133 & \textbf{0.018} $\pm$ 0.002 \\
  & $H^1$   & 0.122 & 0.223 & 0.189 & \textbf{0.035} $\pm$ 0.002 \\
  & Energy  & 0.036 & 0.105 & 0.069 & \textbf{0.013} $\pm$ 0.005 \\
\midrule[0.3pt]
\multirow{3}{*}{1e6}
  & $L^2$   & 0.101 & 0.306 & 0.177 & \textbf{0.025} $\pm$ 0.005 \\
  & $H^1$   & 0.150 & 0.352 & 0.254 & \textbf{0.040} $\pm$ 0.007 \\
  & Energy  & 0.033 & 0.091 & 0.082 & \textbf{0.017} $\pm$ 0.008 \\
\midrule[0.3pt]
\multirow{3}{*}{1.5e6}
  & $L^2$   & 0.464 & 0.363 & 0.398 & \textbf{0.046} $\pm$ 0.011 \\
  & $H^1$   & 0.639 & 0.434 & 0.557 & \textbf{0.070} $\pm$ 0.015 \\
  & Energy  & 0.249 & 0.103 & 0.162 & \textbf{0.026} $\pm$ 0.015 \\
\midrule[0.3pt]
\multirow{3}{*}{2.5e6}
  & $L^2$   & 0.767 & 0.412 & 0.802 & \textbf{0.095} $\pm$ 0.019 \\
  & $H^1$   & 1.000 & 0.494 & 0.996 & \textbf{0.135} $\pm$ 0.028 \\
  & Energy  & 0.514 & 0.141 & 0.590 & \textbf{0.036} $\pm$ 0.016 \\
\bottomrule
\end{tabular}
\end{table*}

\subsection{Error analysis}
We have investigated different diffusion frameworks with different noise scheduling, where we ablate diffusion samplers (computed \emph{once} per setting) across step budgets $N_t\!\in\!\{10,50,100,1000\}$: DDPM with both \emph{linear} and \emph{cosine} noise schedules, DDIM (cosine), and SDE (cosine). Among the samplers we tested, \emph{DDPM with a cosine noise schedule} achieves the highest accuracy at our largest step budget (\(N_t = 1000\)). We therefore adopt DDPM as our default sampler and provide the full ablation details—showing that DDPM consistently outperforms the alternatives—in~\ref{app:sampler}. To compare our results against deterministic frameworks, we have adopted Fourier Neural Operator (FNO) and the more recent Helmholtz Neural Operator (HNO), and trained them on the identical datasets used for the diffusion models (details of FNO and HNO can be found in~\ref{app:methods}). We have computed and interrogated FNO, HNO, and backbone U\mbox{-}Net results as baseline deterministic approaches against our probabilistic framework. Table \ref{tab:all_by_freq} summarizes the relative $L^2$ and $H^1$ errors for six different frequencies, ranging from low 150 kHz to 2.5 MHz. Since often in practice one is interested in energy, we have also quantified the relative energy errors defined as $\mathcal{E}_{\text{energy}}(\hat u,u)=|E(\hat u)-E(u)|/E(u)$ with $E(u)=\int_{\Omega}(\|\nabla u\|^2+\tfrac{k^2}{c(x)^2}|u|^2)\,dx$. 

\clearpage
\begin{figure}[p]
  \centering
  \captionsetup{font=small,skip=6pt}
  \captionsetup[subfigure]{labelformat=empty,justification=centering,aboveskip=2pt,belowskip=2pt}

  \newlength{\rowH}
  \setlength{\rowH}{0.20\textheight}

  \begin{subfigure}[t]{0.98\textwidth}
    \centering
    \caption{\small ($f=1.5\times10^{5}$ Hz)}
    \includegraphics[width=\linewidth,height=\rowH,keepaspectratio]{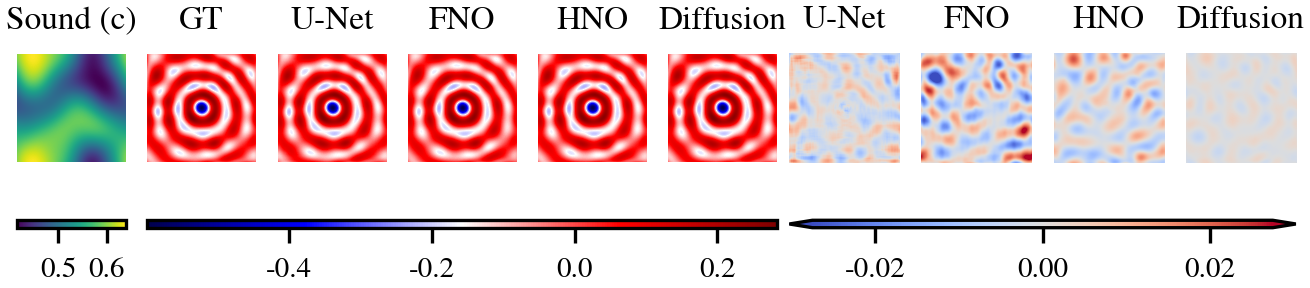}
  \end{subfigure}

  \vspace{4pt}

  \begin{subfigure}[t]{0.98\textwidth}
    \centering
    \caption{\small ($f=5\times10^{5}$ Hz)}
    \includegraphics[width=\linewidth,height=\rowH,keepaspectratio]{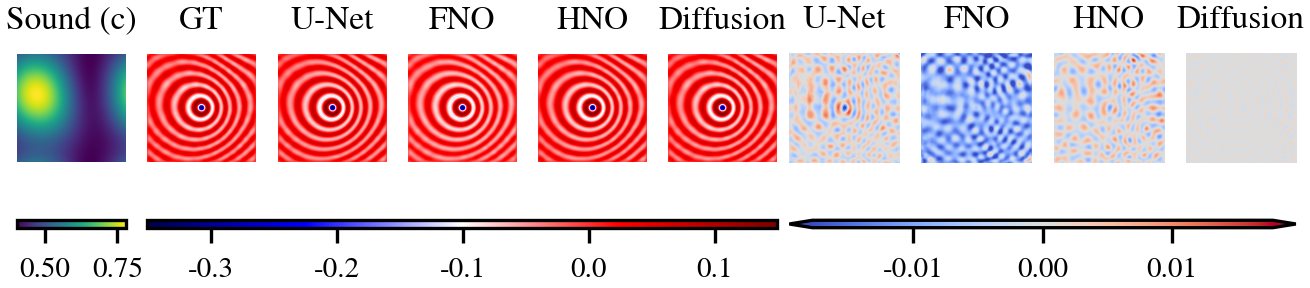}
  \end{subfigure}

  \vspace{4pt}

  \begin{subfigure}[t]{0.98\textwidth}
    \centering
    \caption{\small ($f=1\times10^{6}$ Hz)}
    \includegraphics[width=\linewidth,height=\rowH,keepaspectratio]{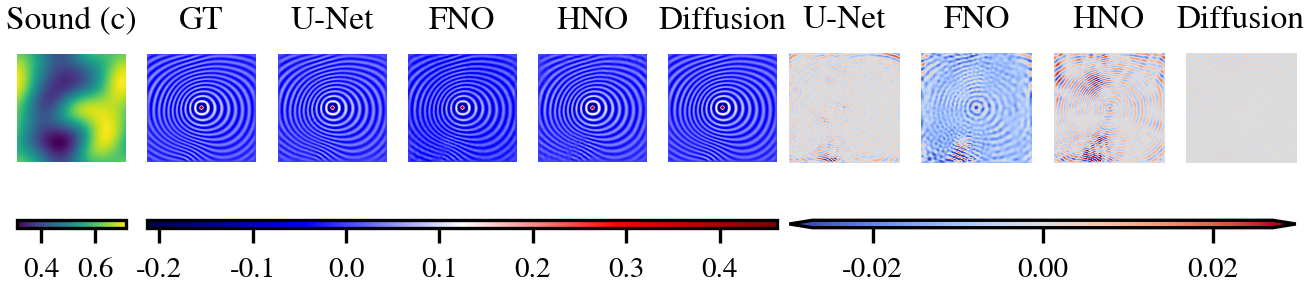}
  \end{subfigure}

  \vspace{4pt}

  \begin{subfigure}[t]{0.98\textwidth}
    \centering
    \caption{\small ($f=2.5\times10^{6}$ Hz)}
    \includegraphics[width=\linewidth,height=\rowH,keepaspectratio]{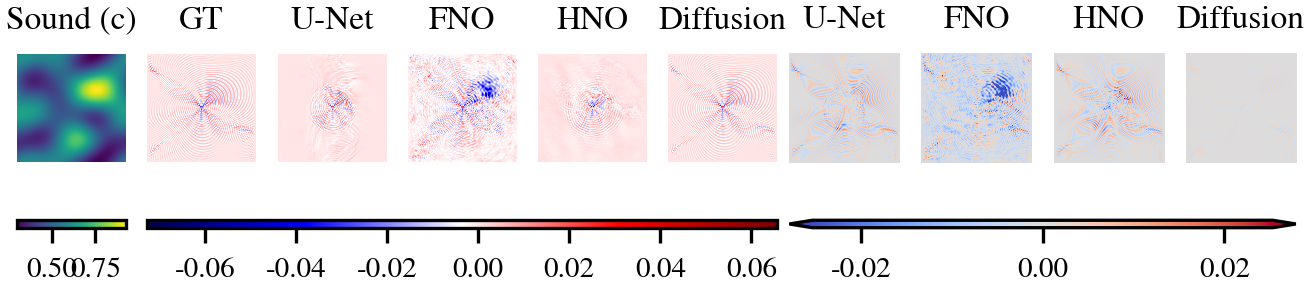}
    \vspace{2pt}
    \subcaption*{\small%
    \makebox[0pt][l]{\hspace*{0.22\linewidth}\emph{Prediction vs.\ Ground Truth}}%
    \makebox[\linewidth][l]{\hspace*{0.70\linewidth}\emph{Residual (Pred$-$GT)}}%
    }
    \vspace{-5pt}
  \end{subfigure}

  \caption{\textbf{Qualitative comparisons across selected frequencies.}
  Each row shows, left-to-right, Sound Map (c), Ground Truth (GT), U-Net, FNO, HNO, Diffusion, followed by the comparison to GT.}

  \label{fig:qual-multifreq-2col}
\end{figure}
\clearpage

We note that diffusion results are obtained by drawing $K=10$ samples per test input, and mean$\pm$std are reported. Our results demonstrate that diffusion models consistently yield smaller errors, and as frequency is ramped up, our probabilistic approach exhibits an order of magnitude lower errors compared to the best of three deterministic approaches. Notably, at 2.5 MHz, relative $L^2$ error for diffusion model is $0.095 \pm 0.019$, while the same quantity is computed for U-Net, FNO, and HNO as 0.767, 0.412, and 0.802, respectively. We further illustrate the computed wavefield solutions for a randomly chosen sound speed map. Figure. \ref{fig:qual-multifreq-2col} shows the ground truth (GT), U-Net, FNO, HNO, and Diffusion results and their corresponding errors at a subset of four frequencies (see~\ref{app:qual} for more results). We observe that the Diffusion model outperforms the other deterministic approaches in capturing wave patterns and fine features of solutions.

\begin{figure}[H]
  \centering
  \includegraphics[width=\linewidth]{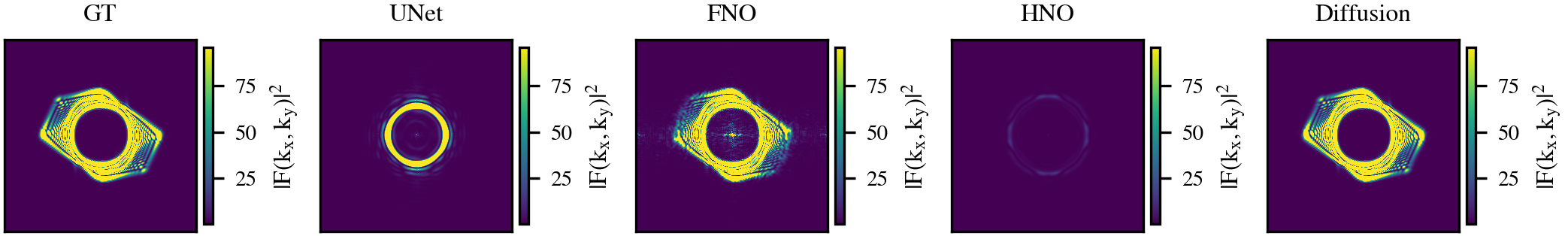}
  \caption{\textbf{2D spectral power comparison at $f=2.5\times10^6$ Hz.}
  Example Fourier power spectra $|F(k_x,k_y)|^2$ for GT and each model on a single test sample.}
  \label{fig:spectrum-2d}
\end{figure}

To make these observations more quantitative, we also analyze the 2D Fourier power spectra of the predicted wavefields. For a fixed high frequency (e.g., $f = 2.5\times 10^{6}\,\text{Hz}$), we compute the 2D FFT of GT and each model prediction and visualize the corresponding power $|F(k_x,k_y)|^2$ over the full $(k_x,k_y)$ domain. As illustrated in Fig.~\ref{fig:spectrum-2d}, the deterministic baselines (in particular the U\mbox{-}Net backbone and HNO) exhibit clear spectral bias: they oversmooth or misallocate energy at large wavenumbers, deviating substantially from the ground-truth spectrum. In contrast, the Diffusion model closely tracks the GT power spectrum across both low and high spatial frequencies, indicating that it captures the full range of oscillatory content rather than collapsing toward low-$k$ modes.

\subsection{Sensitivity analysis}
\begin{figure}[H]
  \centering
  \includegraphics[trim=0cm 2.2cm 0cm 0.5cm, clip, width=0.7\linewidth]{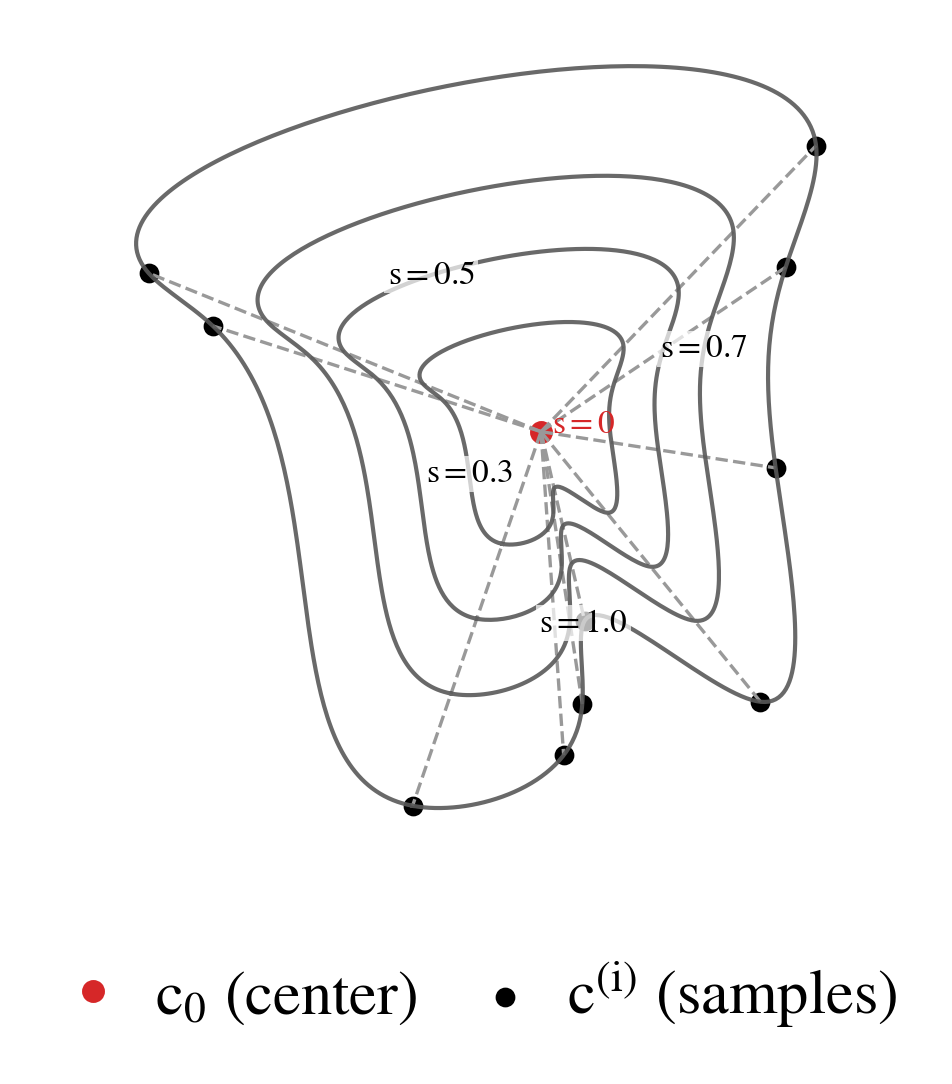}
  \caption{\textbf{Illustration of coefficient-function space and interpolation paths.}
  The red point denotes the reference (baseline) medium \(c_0\). The surrounding black points \(\{c^{(d)}\}_{d=1}^{D}\) indicate distinct GRF directions/realizations in coefficient space. For each direction \(d\), we define a linear homotopy (dashed ray)
  \(c^{(d)}(s)=(1-s)c_0+s\,c^{(d)}\), which traces a path from the center \(c_0\) toward the endpoint \(c^{(d)}\).
  The concentric contour circles visualize fixed interpolation levels \(s\in\{0.3,0.5,0.7,1.0\}\): moving outward corresponds to increasing deviation from \(c_0\), with \(s=1.0\) reaching the full GRF realization.}
  \label{fig:funcspace}
\end{figure}

To quantify how uncertainty in the sound-speed map is pushed forward to uncertainty in the wavefield, we probe the solution map along controlled perturbation paths in coefficient-function space (Fig.~\ref{fig:funcspace}). We first select a reference medium \(c_0\) (the red point) and draw \(D{=}100\) GRF realizations \(\{c^{(d)}\}_{d=1}^{D}\) that define distinct perturbation directions/endpoints (black points). For each direction \(d\), we construct a linear homotopy
\begin{equation}
c^{(d)}(s) \;=\; (1-s)c_0 + s\,c^{(d)}, \qquad s\in[0,1],
\end{equation}
discretized into \(S{=}100\) interpolation steps. In the geometric view of Fig.~\ref{fig:funcspace}, each dashed ray is one path from \(c_0\) toward a GRF endpoint, and the concentric contour circles represent fixed interpolation radii (e.g., \(s=0.3,0.5,0.7,1.0\)), i.e., increasing deviation from the baseline.

For every coefficient realization \(c^{(d)}(s)\), we evaluate the model prediction \(u_M(\cdot;\,c^{(d)}(s))\) at \(8\) probe pixels: \(4\) \emph{near} the source region and \(4\) \emph{far} near the PML boundary. For small \(s\), the initial slope along each dashed path provides a local, directional measure of sensitivity,
\begin{equation}
\left.\frac{d}{ds}u(\cdot;\,c^{(d)}(s))\right|_{s=0}
\;\approx\;
DS[c_0]\,[c^{(d)}-c_0],
\end{equation}
which directly connects to the high-frequency scaling
\(|\delta u/u_0|\lesssim (k\,r/c_0)\,(\|\delta c\|/c_0)\) (Sec.~\ref{sec:background}). As \(s\) increases (moving outward across the contour circles), the medium departs further from \(c_0\), and the responses increasingly reflect accumulated phase error and interference variability rather than purely local linear behavior.

\begin{figure}[t]
  \centering
  \captionsetup{skip=2pt}
  \captionsetup[sub]{font=small}

  \begin{subfigure}[t]{0.48\linewidth}
    \centering
    \includegraphics[width=\linewidth,height=0.21\textheight,keepaspectratio]{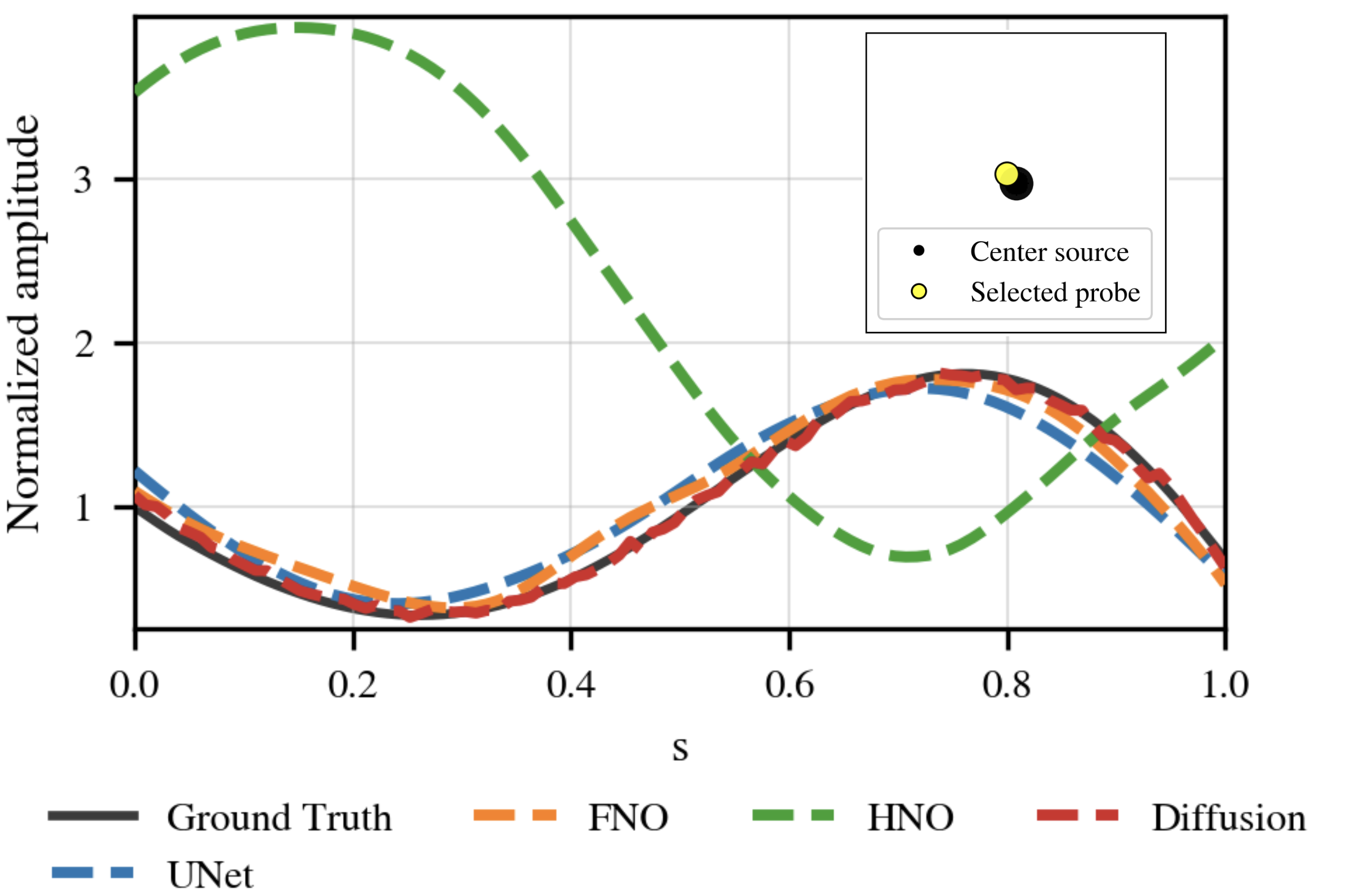}
  \end{subfigure}\hfill
  \begin{subfigure}[t]{0.48\linewidth}
    \centering
    \includegraphics[width=\linewidth,height=0.21\textheight,keepaspectratio]{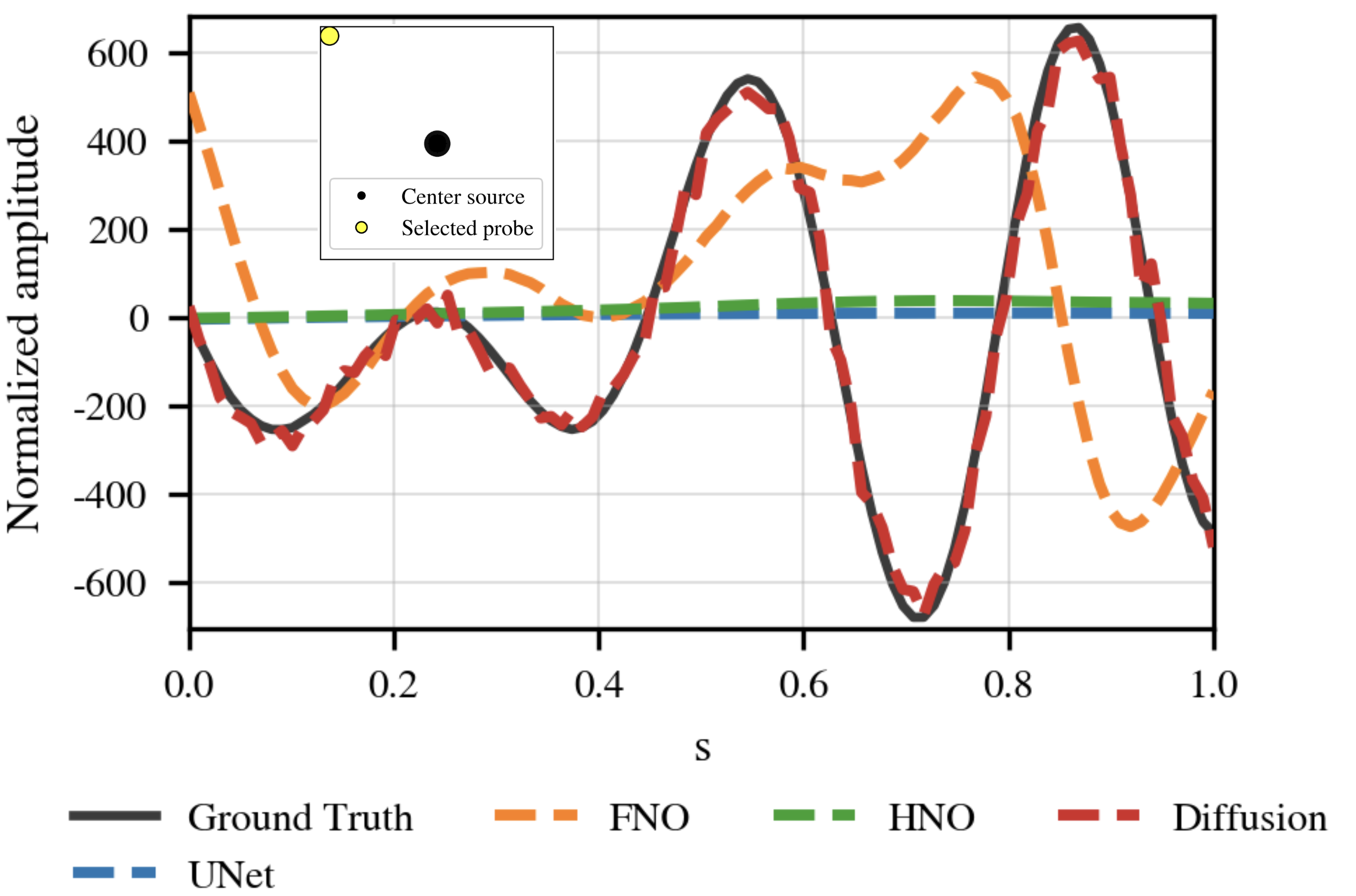}
  \end{subfigure}

  \vspace{2pt}

  \begin{subfigure}[t]{0.48\linewidth}
    \centering
    \includegraphics[width=\linewidth,height=0.21\textheight,keepaspectratio]{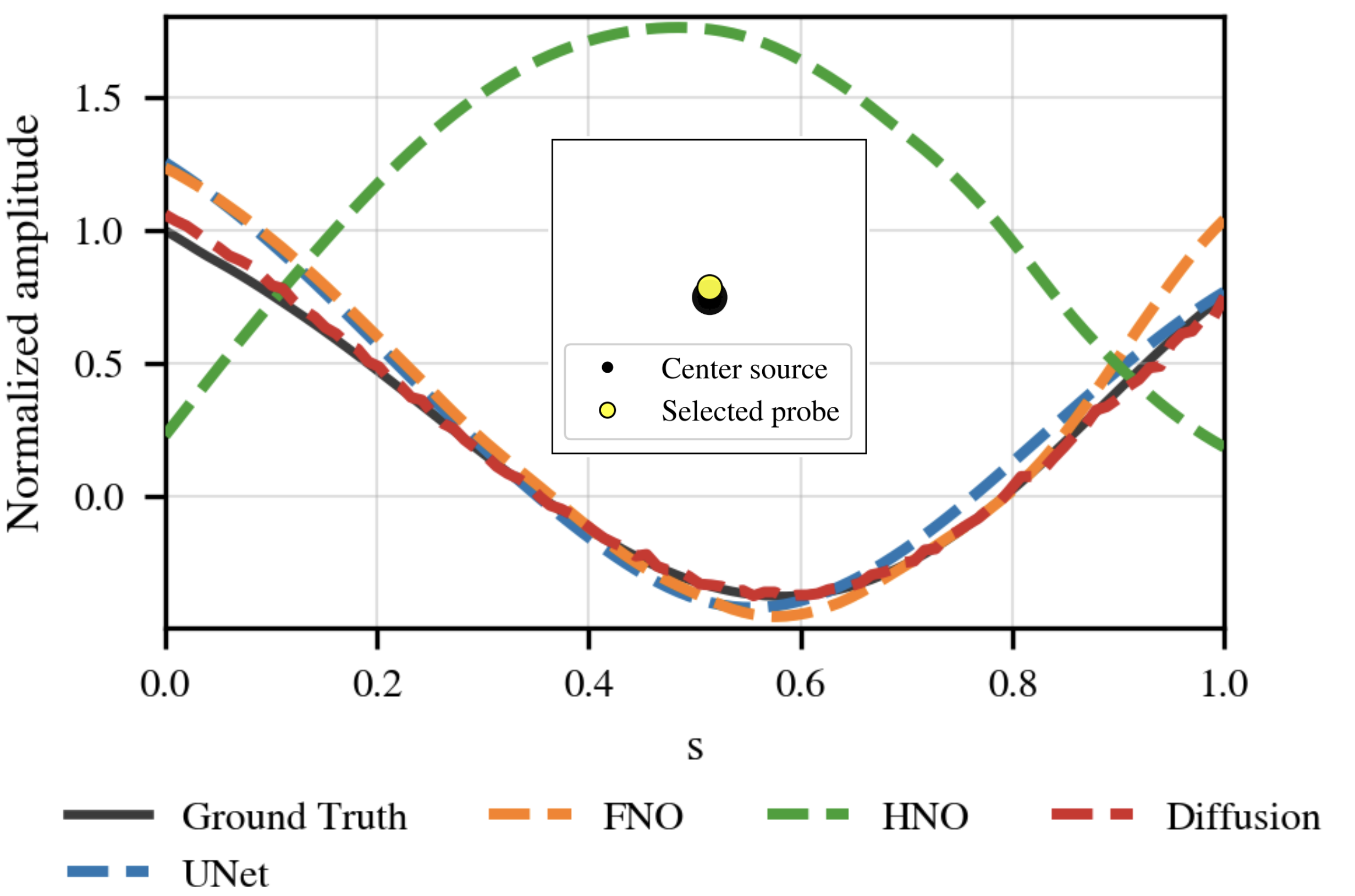}
  \end{subfigure}\hfill
  \begin{subfigure}[t]{0.48\linewidth}
    \centering
    \includegraphics[width=\linewidth,height=0.21\textheight,keepaspectratio]{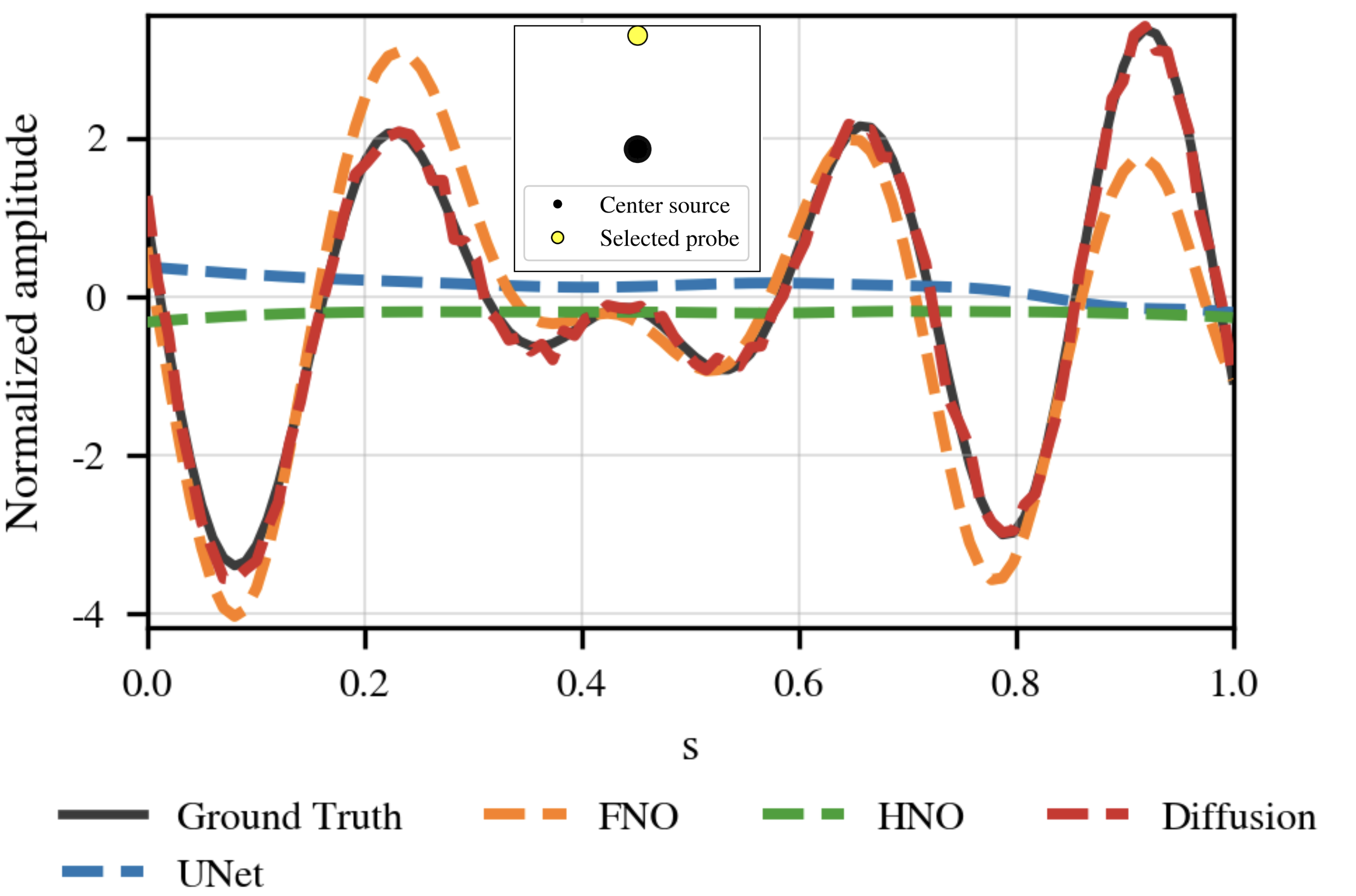}
  \end{subfigure}

  \vspace{2pt}

  \begin{subfigure}[t]{0.48\linewidth}
    \centering
    \includegraphics[width=\linewidth,height=0.21\textheight,keepaspectratio]{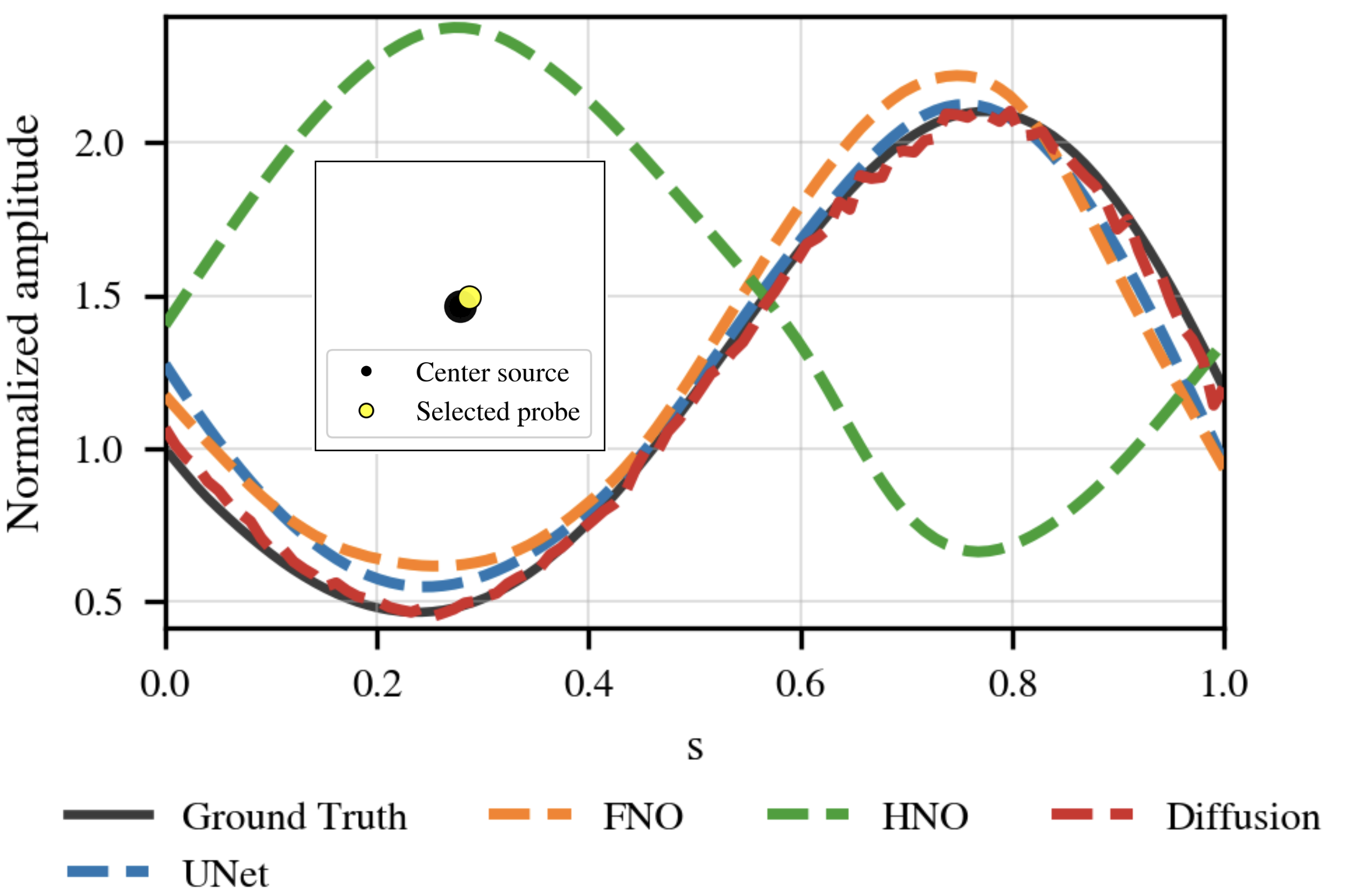}
  \end{subfigure}\hfill
  \begin{subfigure}[t]{0.48\linewidth}
    \centering
    \includegraphics[width=\linewidth,height=0.21\textheight,keepaspectratio]{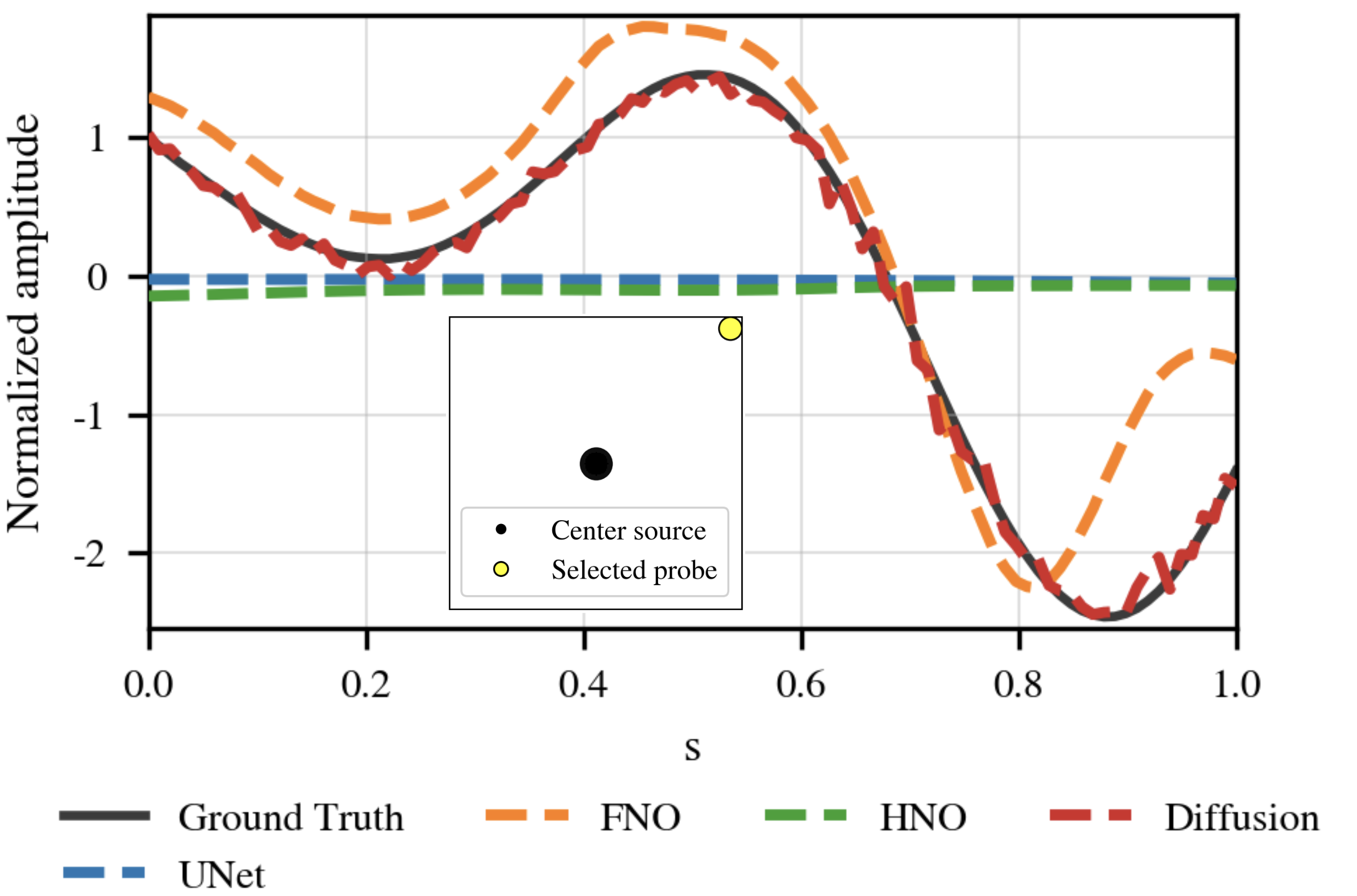}
  \end{subfigure}

  \vspace{2pt}

  \begin{subfigure}[t]{0.48\linewidth}
    \centering
    \includegraphics[width=\linewidth,height=0.21\textheight,keepaspectratio]{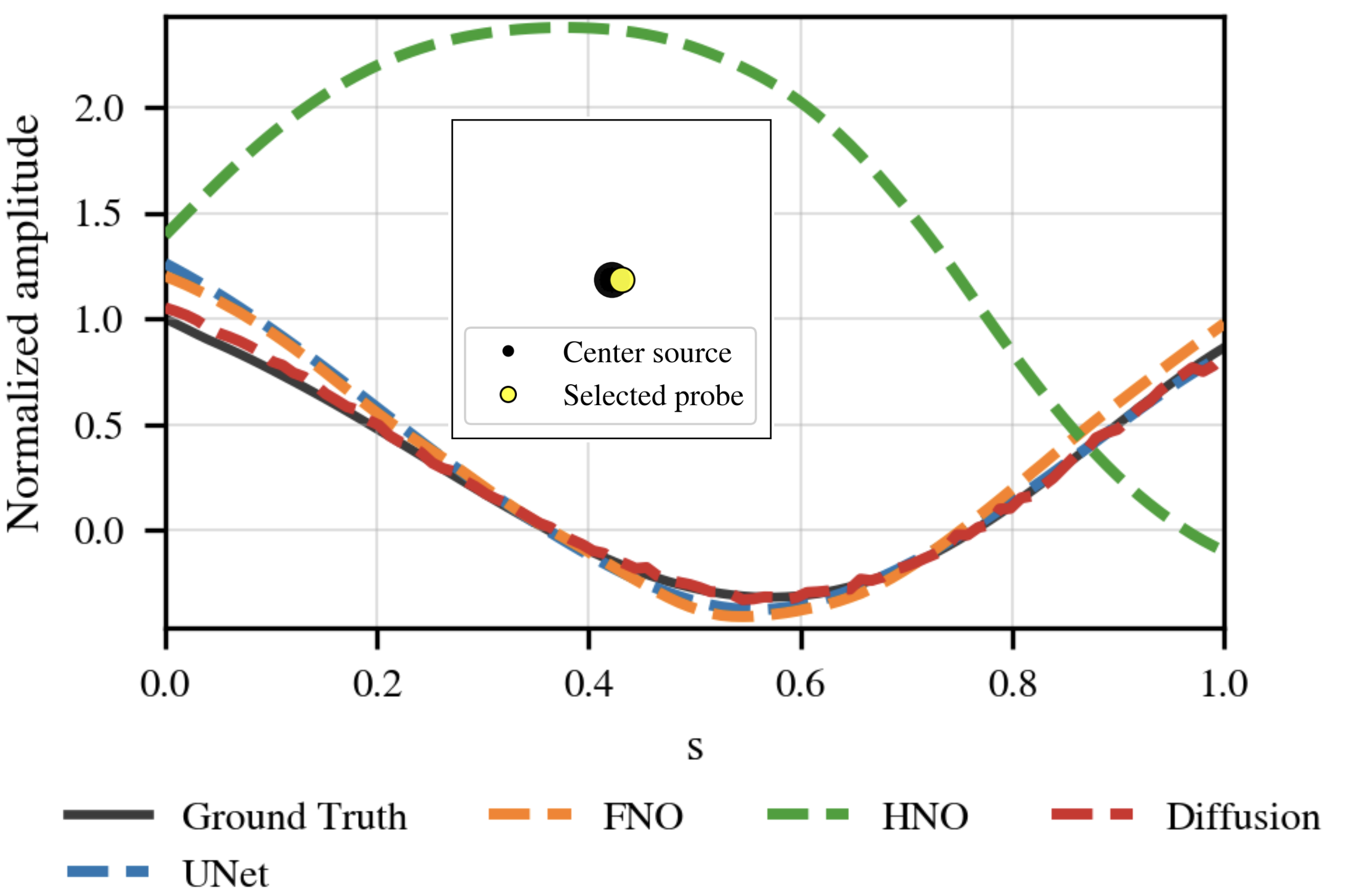}
    \caption*{Near source}
  \end{subfigure}\hfill
  \begin{subfigure}[t]{0.48\linewidth}
    \centering
    \includegraphics[width=\linewidth,height=0.21\textheight,keepaspectratio]{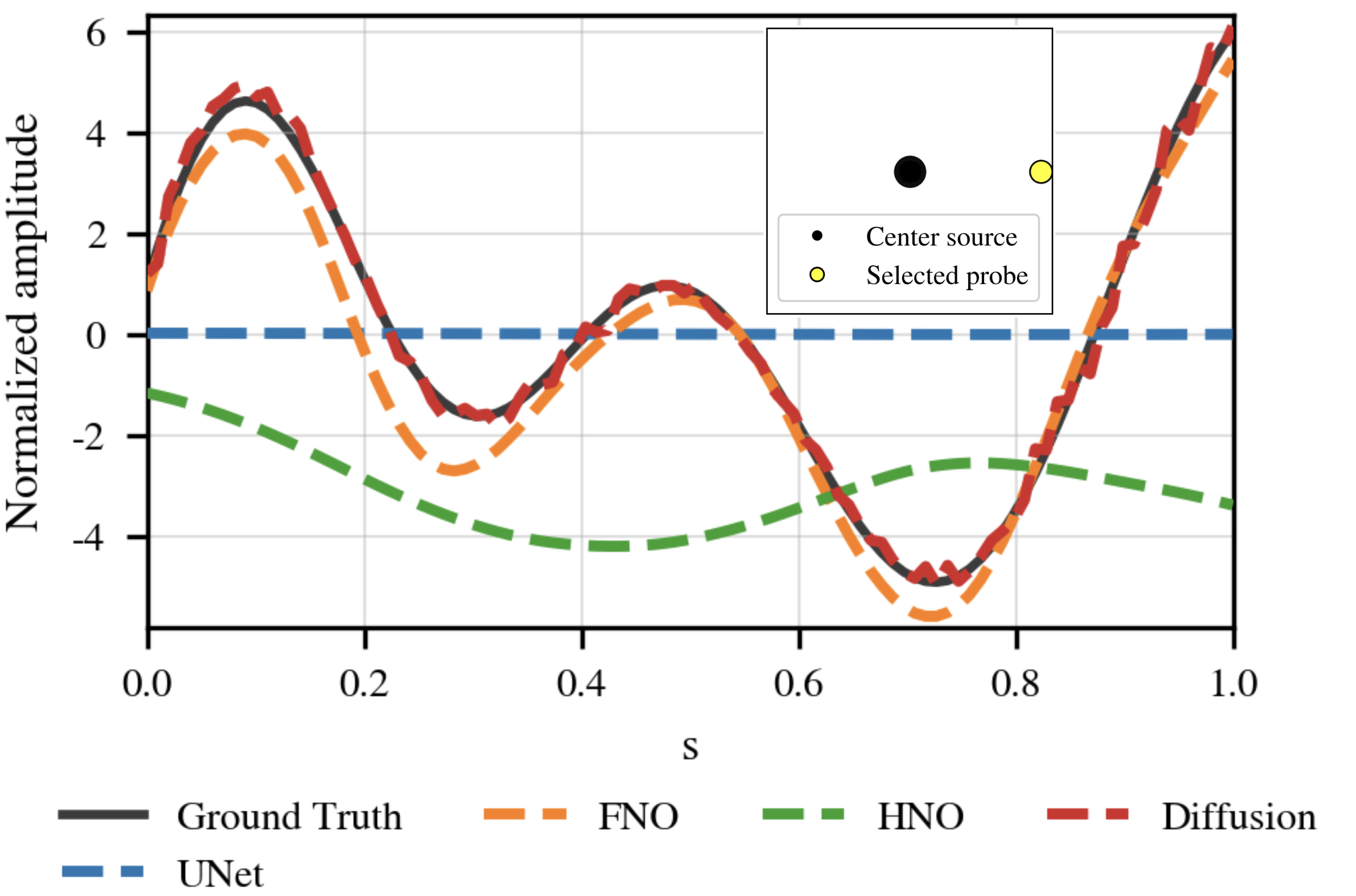}
    \caption*{Near boundary}
  \end{subfigure}

  \caption{\textbf{Sampling along a coefficient path  \(\mathbf{d{=}1}\) (all 4 near vs. all 4 far).}
  For each pair of media, we linearly interpolate the sound speed \(c_s=(1-s)c_0+s\,c_1\) with \(s\in[0,1]\) and track the wavefield amplitude at fixed probe pixels. Left column: near-source probes. Right column: near-boundary probes.}
  \label{fig:cross-sampling-dir1}
\end{figure}

Placing probes both near and far allows a direct examination of the distance-to-source dependence $L\propto k\,r/c_0$. We utilize two complementary visualizations: first, sampling along a fixed direction $d$ to show rate-of-change along a single path, and second, kernel density estimate (KDE) across directions at fixed $s$ to estimate the pushforward variability of $u$ under random coefficient perturbations. Hence, we interrogate sensitivity locally near $c_0$ (at $s{=}0$ and $s{=}0.1$) and at finite amplitudes (around $s\approx 1$) within one unified experimental design. At the highest frequency ($2.5{\times}10^6$ Hz), sampling plot reveals a clear separation: at the near-source probe, both diffusion model and majority of deterministic models can follow the ground-truth trajectory $u^\star(s,d^\star;y,x)$ across the entire path, matching both amplitude and phase; but at the near-boundary probe the discrepancy amplifies—only diffusion reproduces the rapid, non-monotone oscillations and turning points of the complex trajectory, while most deterministic models drift or flatten, consistent with the boundary oversmoothing in \cref{fig:qual-multifreq-2col} (\cref{fig:cross-sampling-dir1}). 

\begin{figure}[t]
  \centering
  \captionsetup{skip=2pt}
  \captionsetup[sub]{font=small}

  \begin{subfigure}[t]{0.48\linewidth}
    \centering
    \includegraphics[width=\linewidth,height=0.21\textheight,keepaspectratio]{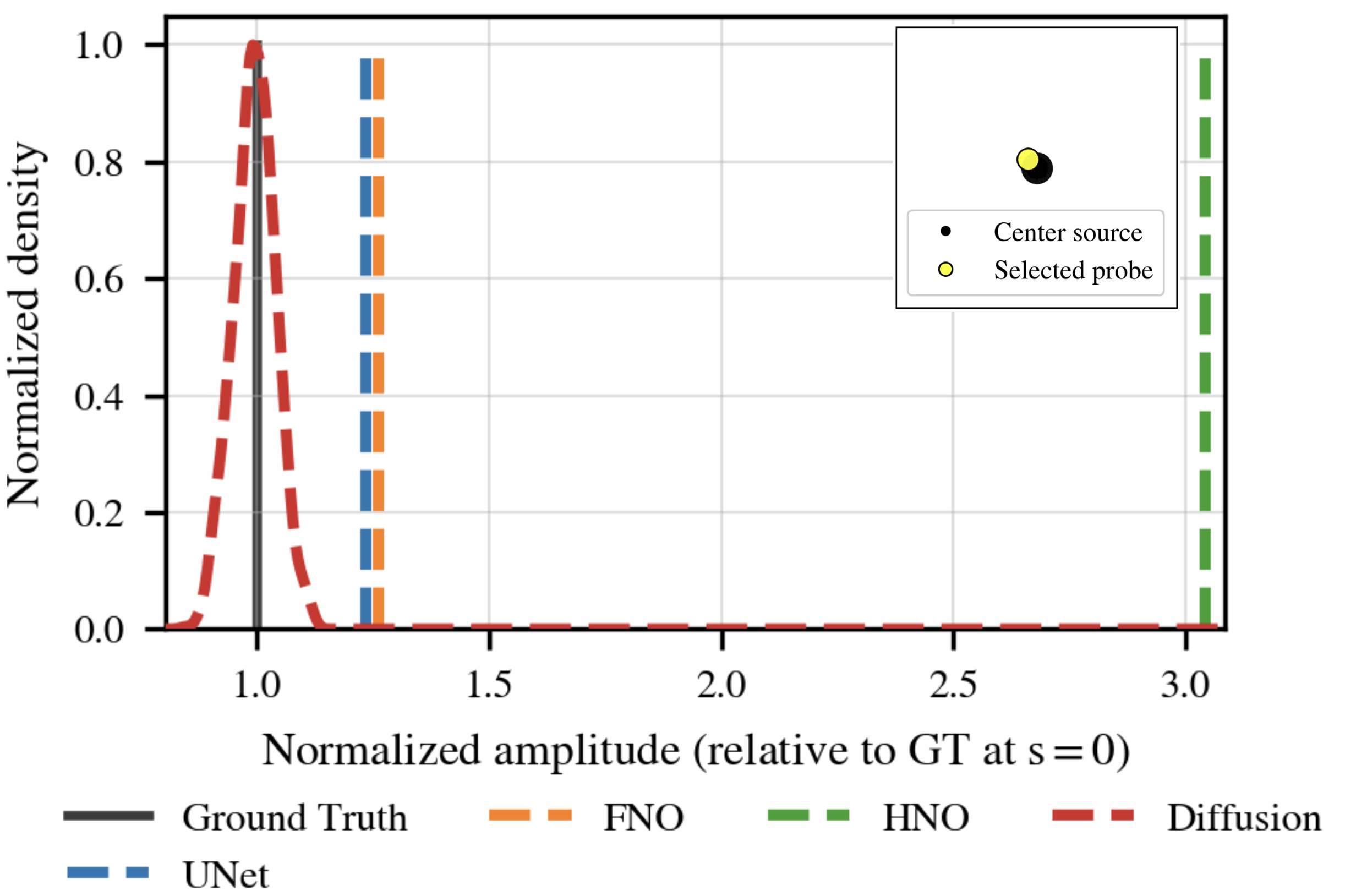}
  \end{subfigure}\hfill
  \begin{subfigure}[t]{0.48\linewidth}
    \centering
    \includegraphics[width=\linewidth,height=0.21\textheight,keepaspectratio]{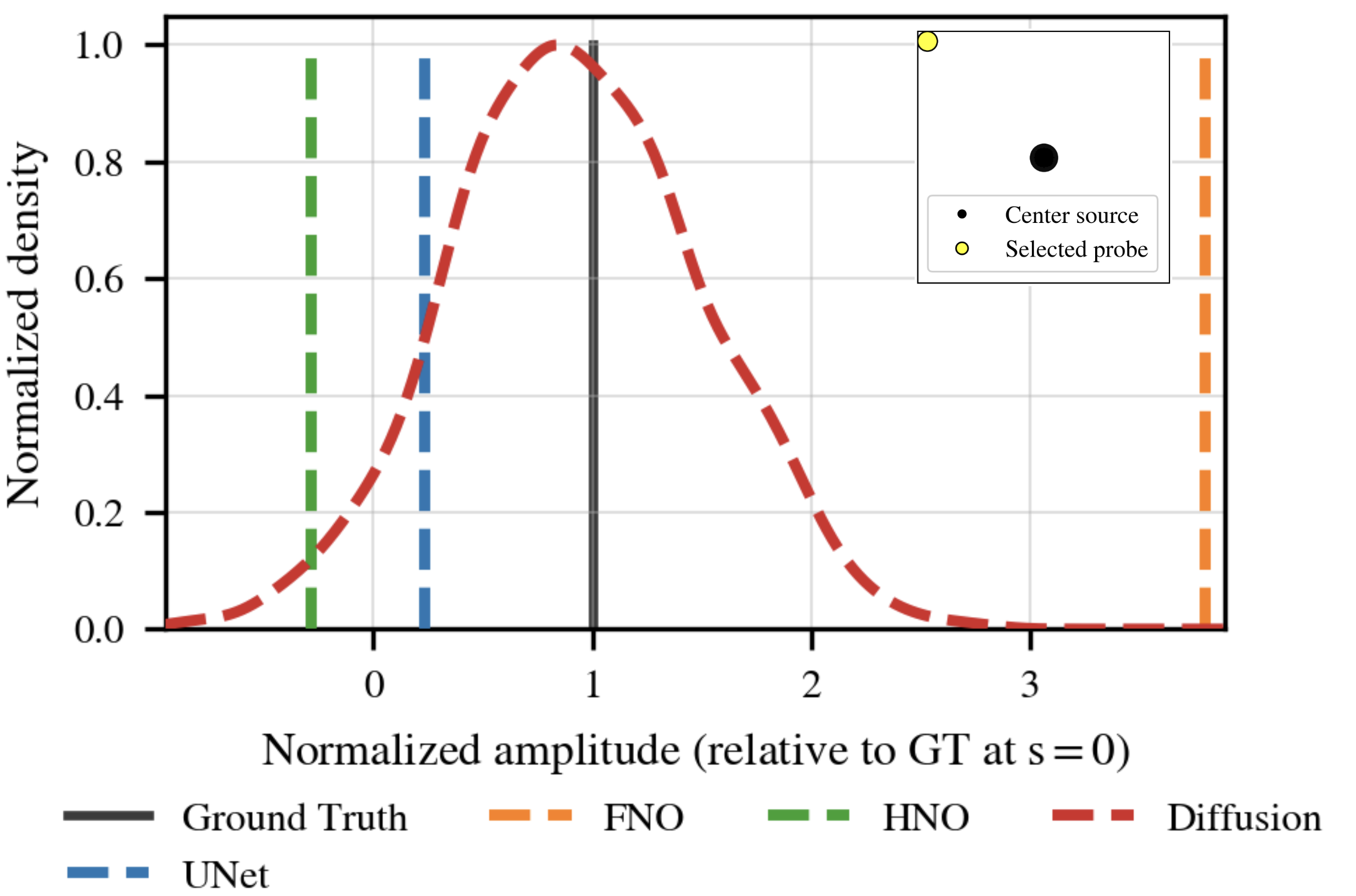}
  \end{subfigure}

  \vspace{2pt}

  \begin{subfigure}[t]{0.48\linewidth}
    \centering
    \includegraphics[width=\linewidth,height=0.21\textheight,keepaspectratio]{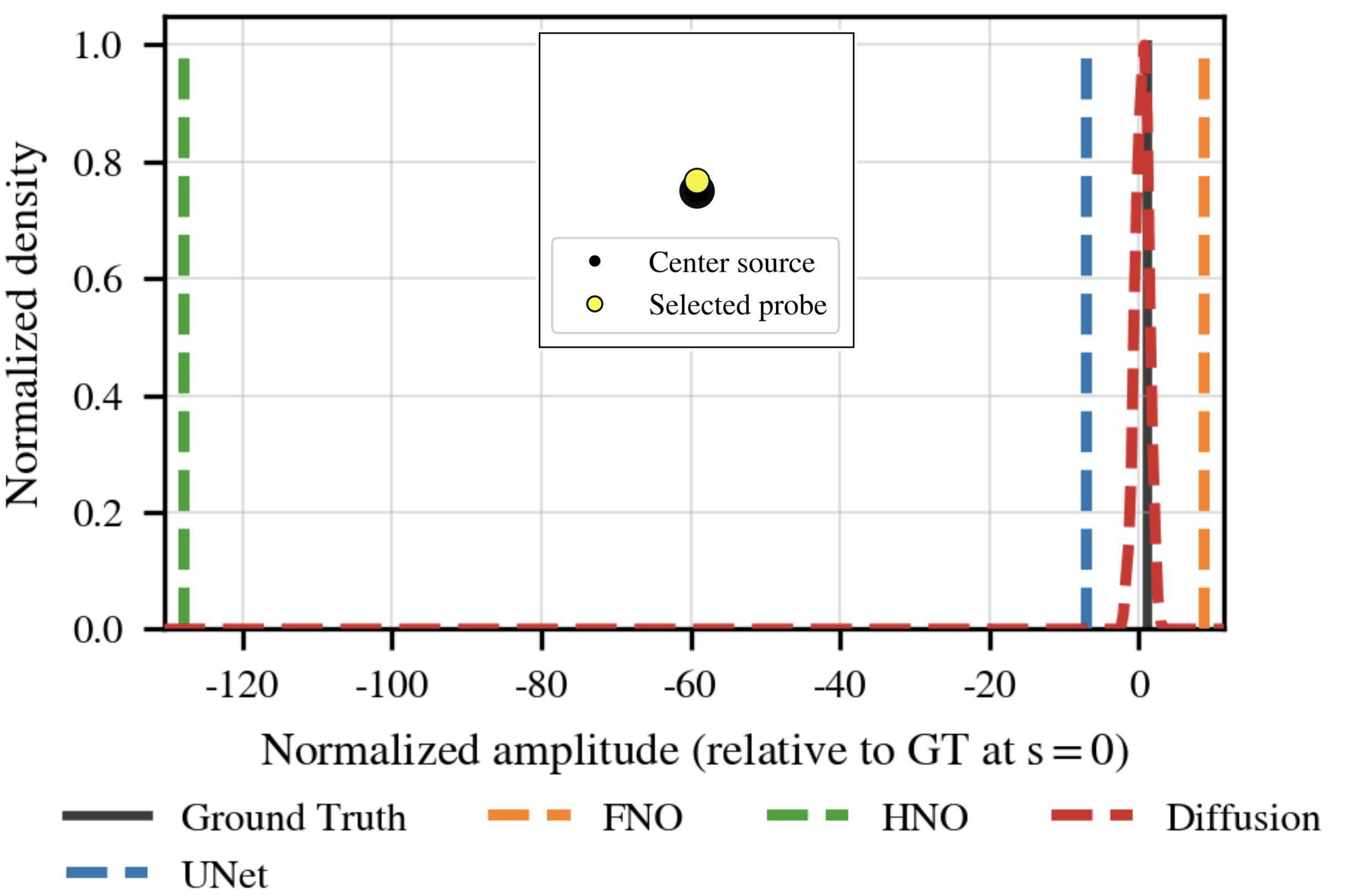}
  \end{subfigure}\hfill
  \begin{subfigure}[t]{0.48\linewidth}
    \centering
    \includegraphics[width=\linewidth,height=0.21\textheight,keepaspectratio]{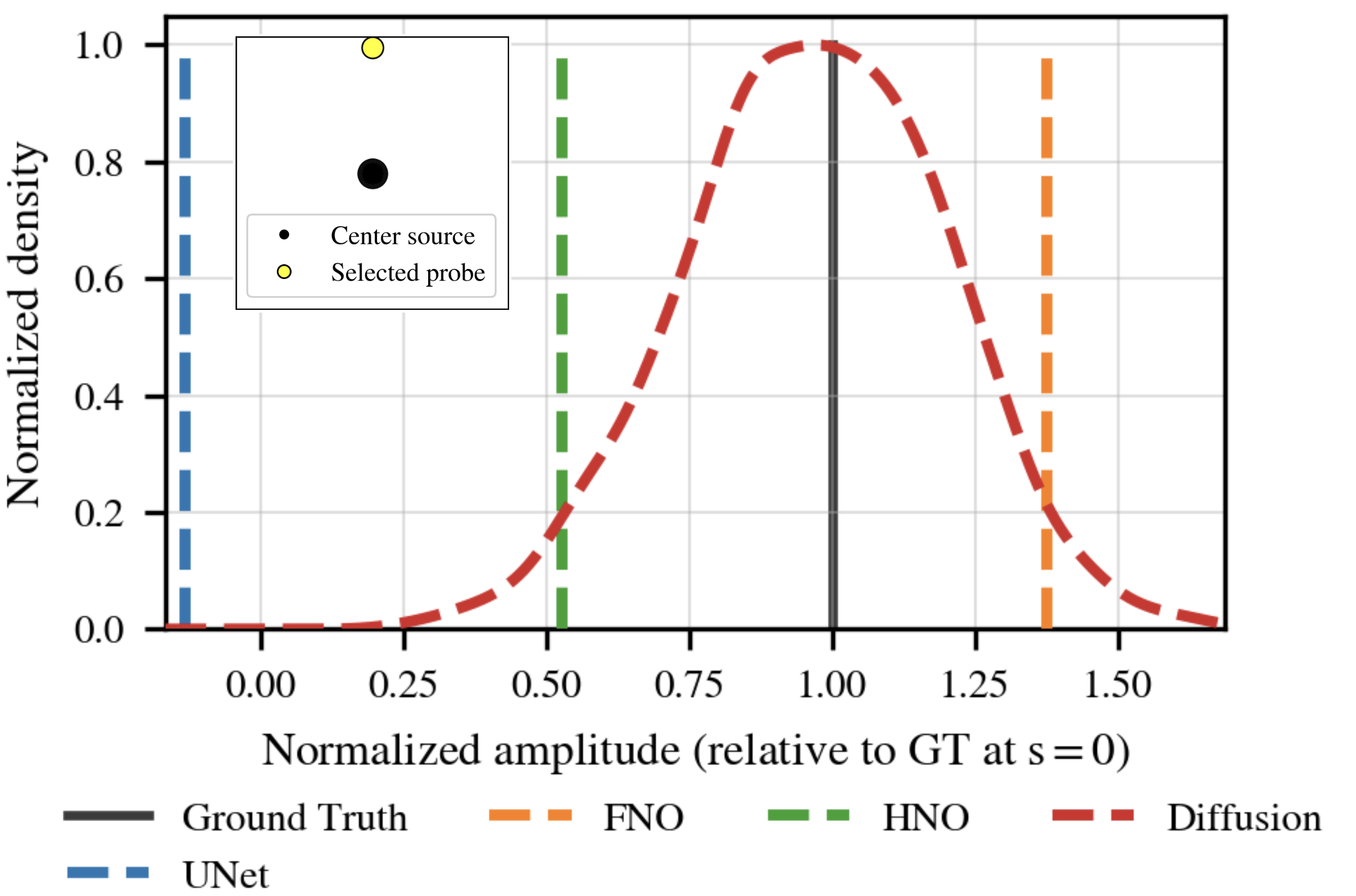}
  \end{subfigure}

  \vspace{2pt}

  \begin{subfigure}[t]{0.48\linewidth}
    \centering
    \includegraphics[width=\linewidth,height=0.21\textheight,keepaspectratio]{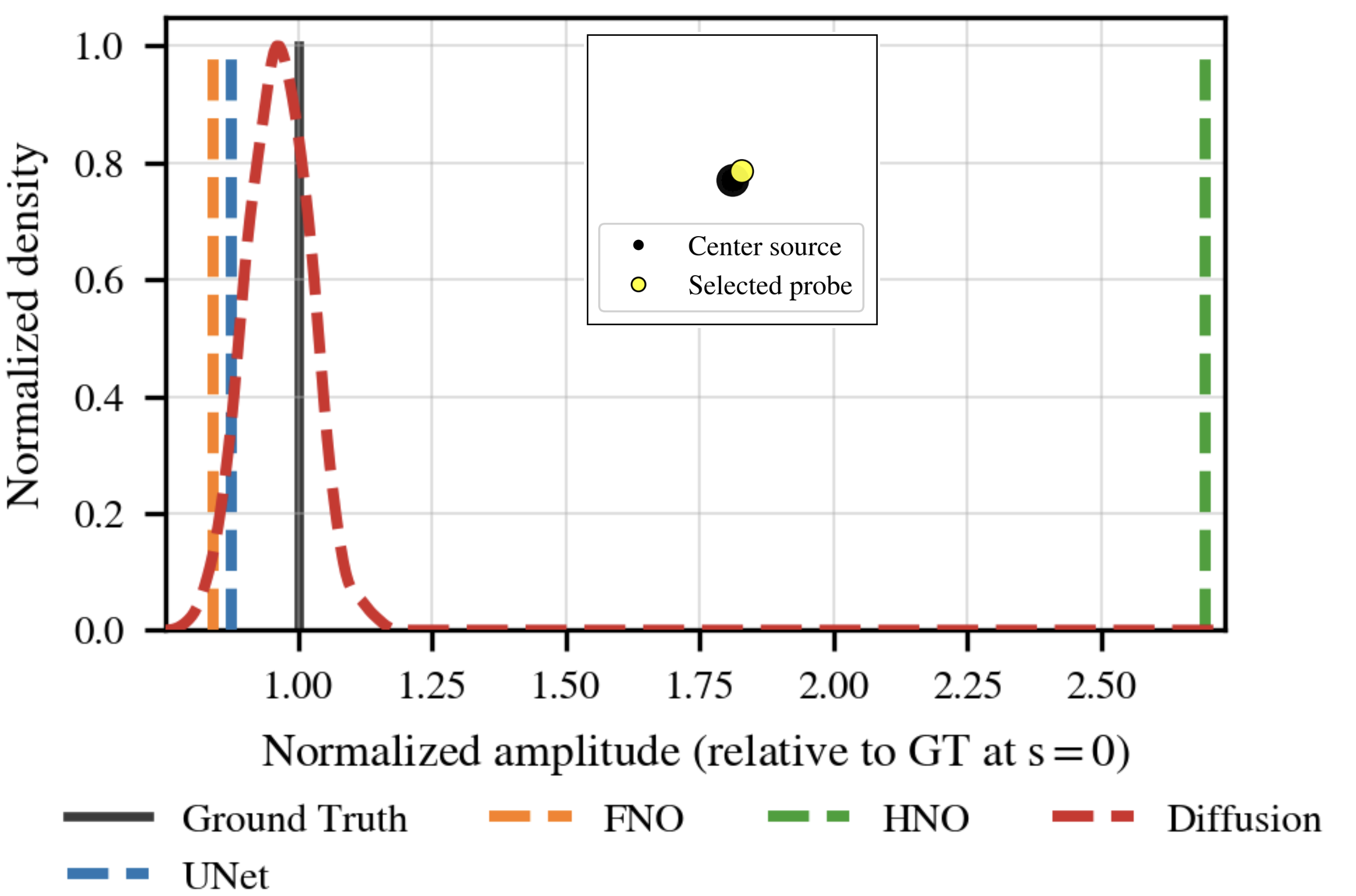}
  \end{subfigure}\hfill
  \begin{subfigure}[t]{0.48\linewidth}
    \centering
    \includegraphics[width=\linewidth,height=0.21\textheight,keepaspectratio]{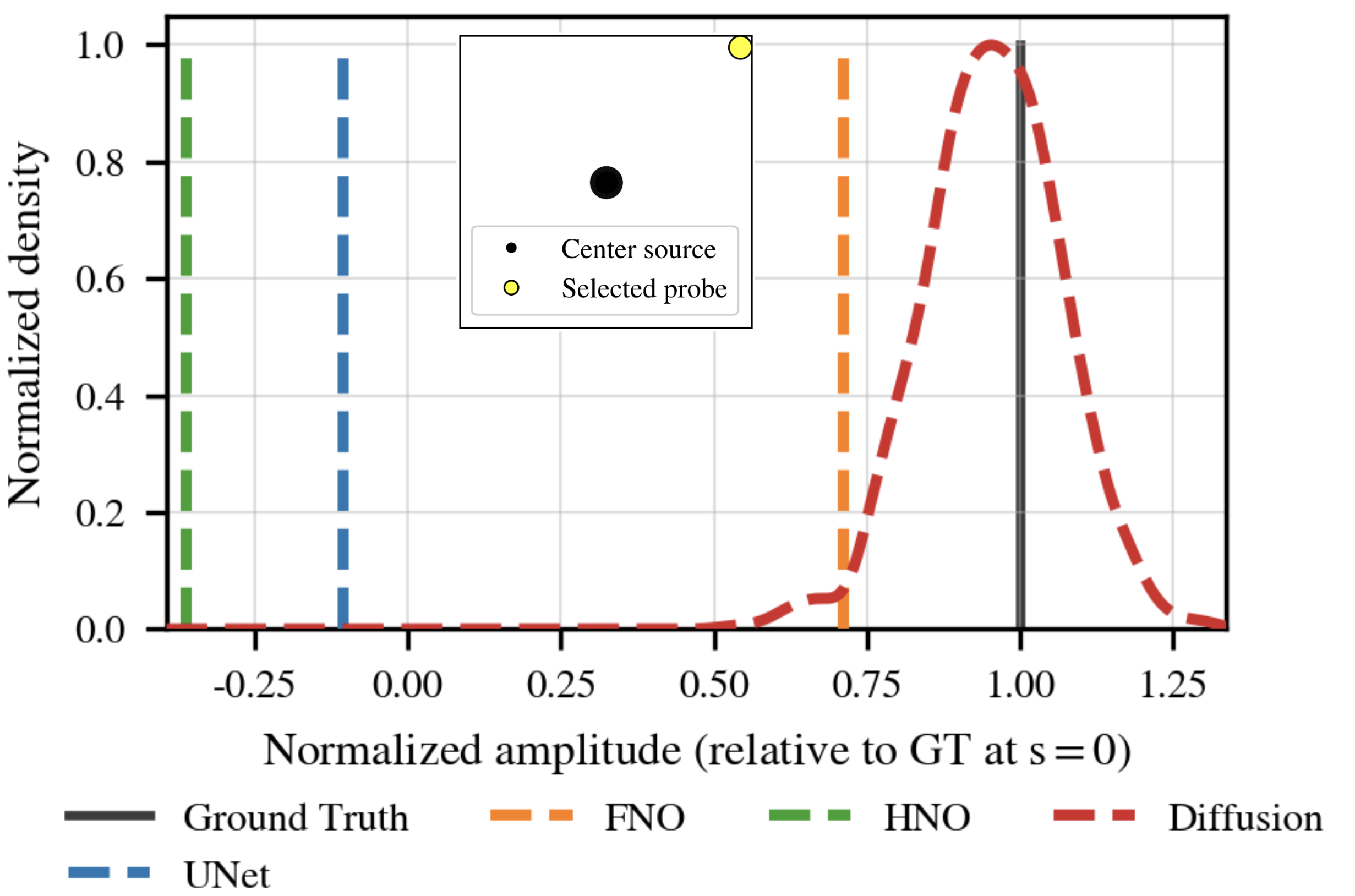}
  \end{subfigure}

  \vspace{2pt}

  \begin{subfigure}[t]{0.48\linewidth}
    \centering
    \includegraphics[width=\linewidth,height=0.21\textheight,keepaspectratio]{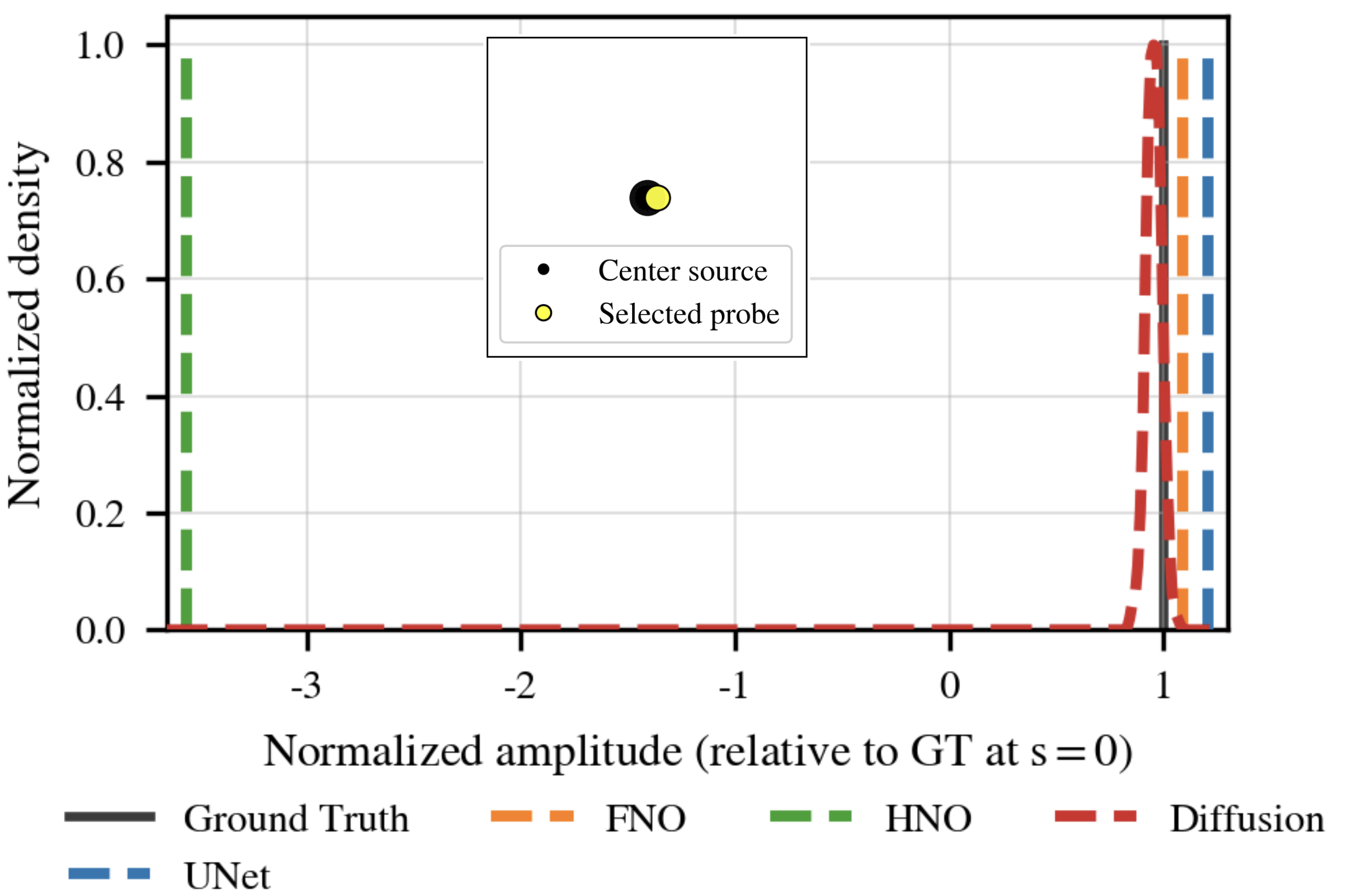}
    \caption*{Near source}
  \end{subfigure}\hfill
  \begin{subfigure}[t]{0.48\linewidth}
    \centering
    \includegraphics[width=\linewidth,height=0.21\textheight,keepaspectratio]{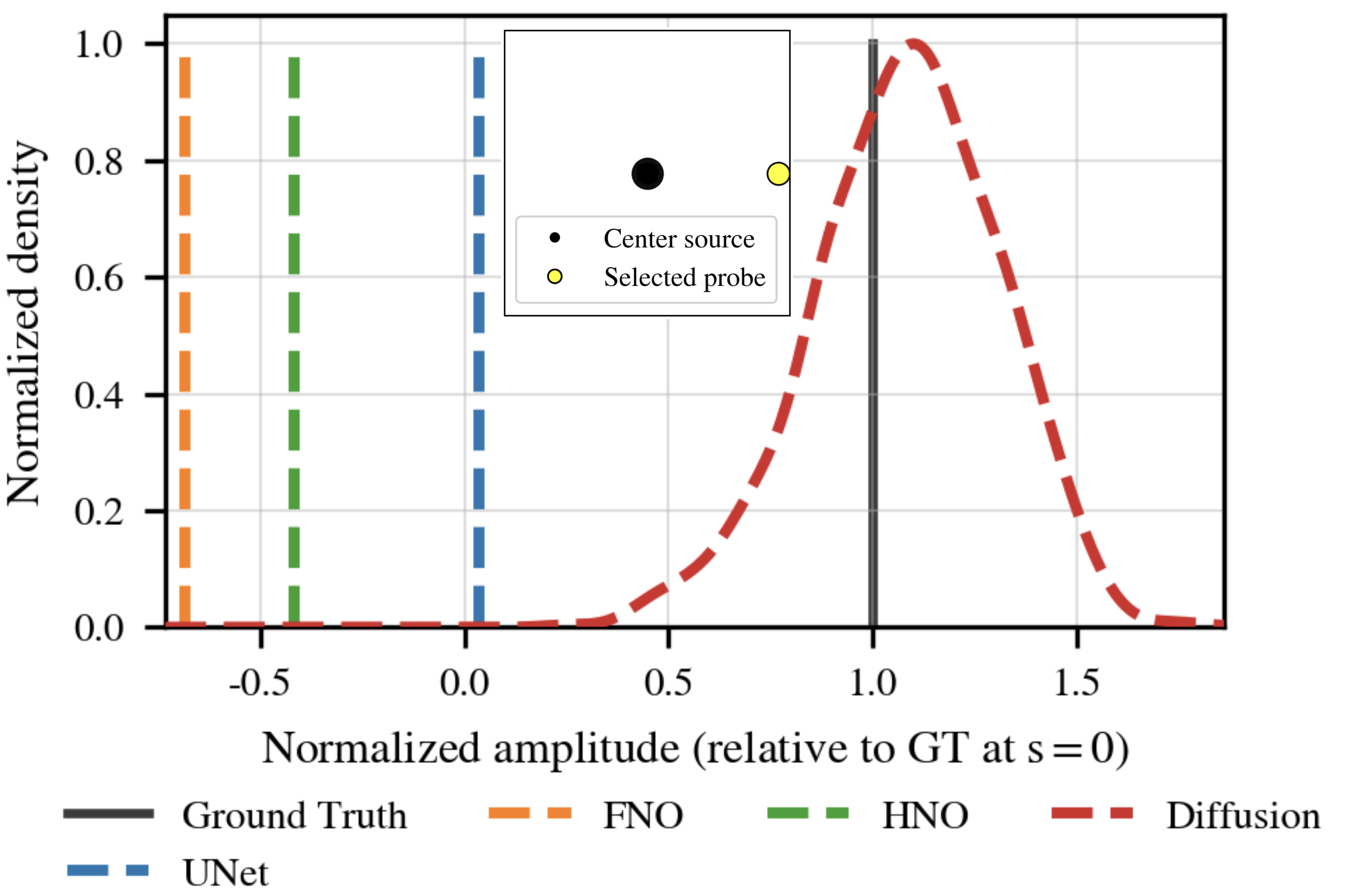}
    \caption*{Near boundary}
  \end{subfigure}

  \caption{\textbf{Kernel density estimates across directions for \(\mathbf{s{=}0}\) (all 4 near vs.\ all 4 far).}
  Left column: near the source. Right column: near the boundary.}
  \label{fig:s0-all}
\end{figure}

\begin{figure}[t]
  \centering
  \captionsetup{skip=2pt}
  \captionsetup[sub]{font=small}

  \begin{subfigure}[t]{0.48\linewidth}
    \centering
    \includegraphics[width=\linewidth,height=0.21\textheight,keepaspectratio]{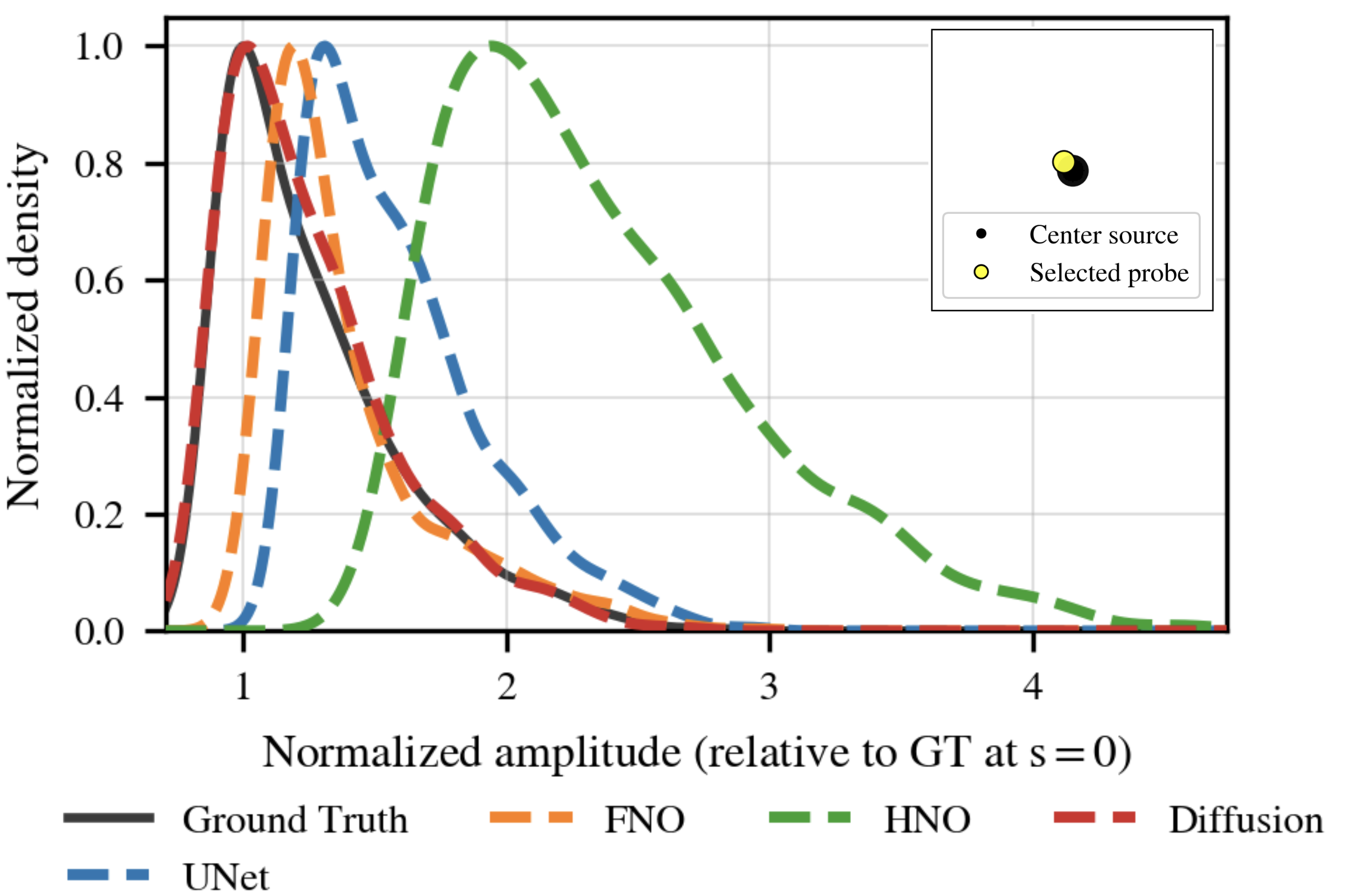}
  \end{subfigure}\hfill
  \begin{subfigure}[t]{0.48\linewidth}
    \centering
    \includegraphics[width=\linewidth,height=0.21\textheight,keepaspectratio]{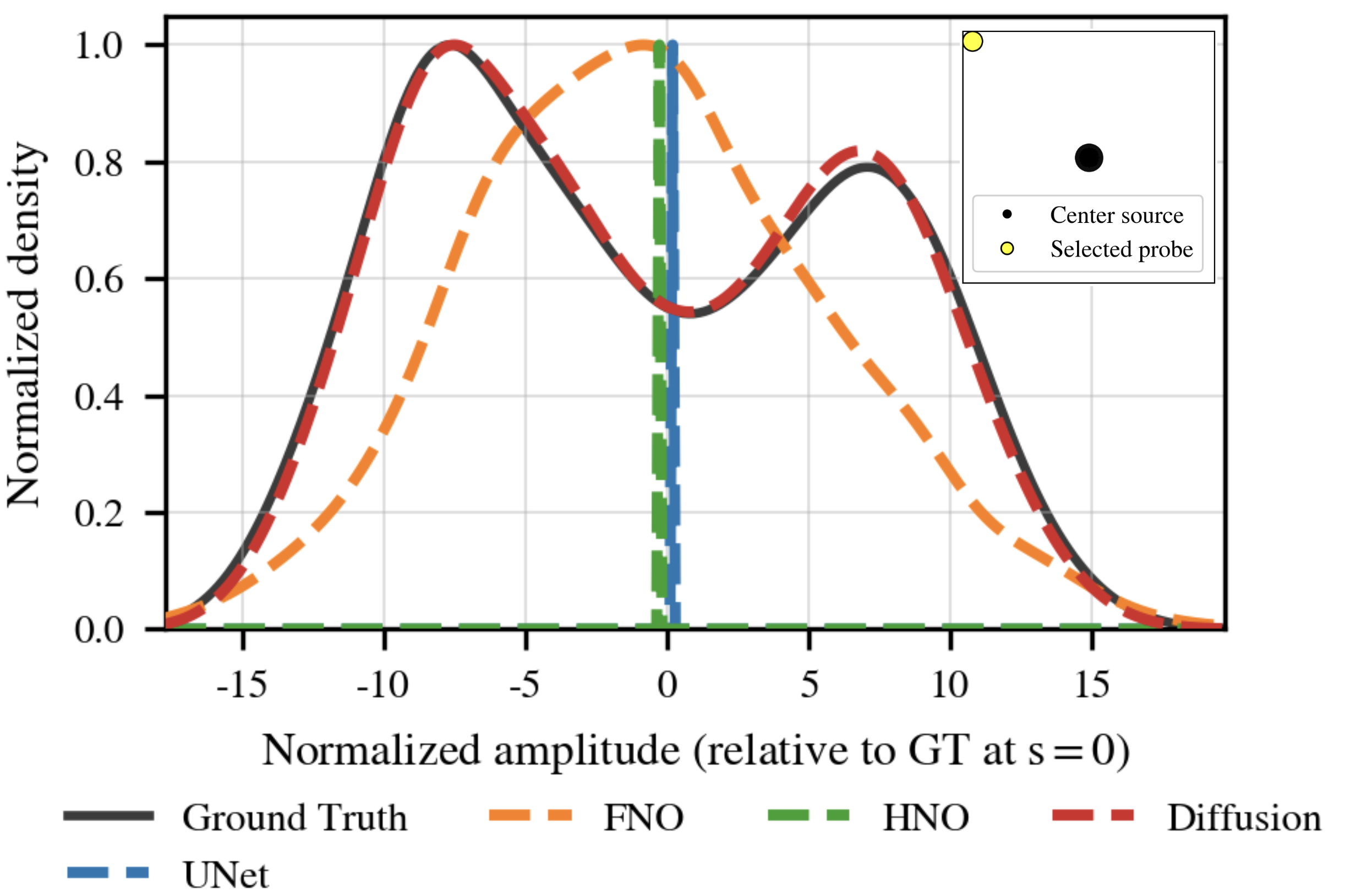}
  \end{subfigure}

  \vspace{2pt}

  \begin{subfigure}[t]{0.48\linewidth}
    \centering
    \includegraphics[width=\linewidth,height=0.21\textheight,keepaspectratio]{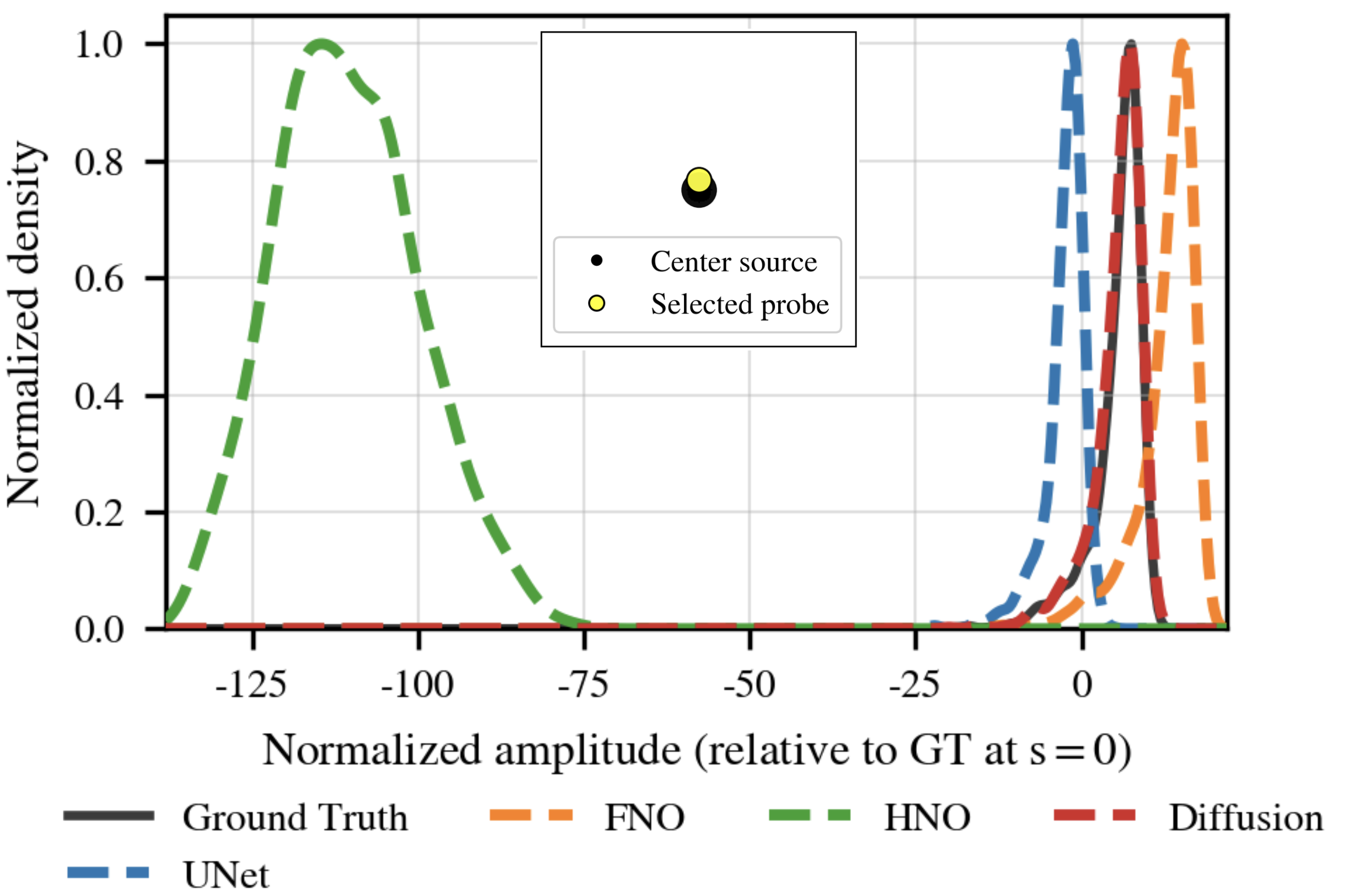}
  \end{subfigure}\hfill
  \begin{subfigure}[t]{0.48\linewidth}
    \centering
    \includegraphics[width=\linewidth,height=0.21\textheight,keepaspectratio]{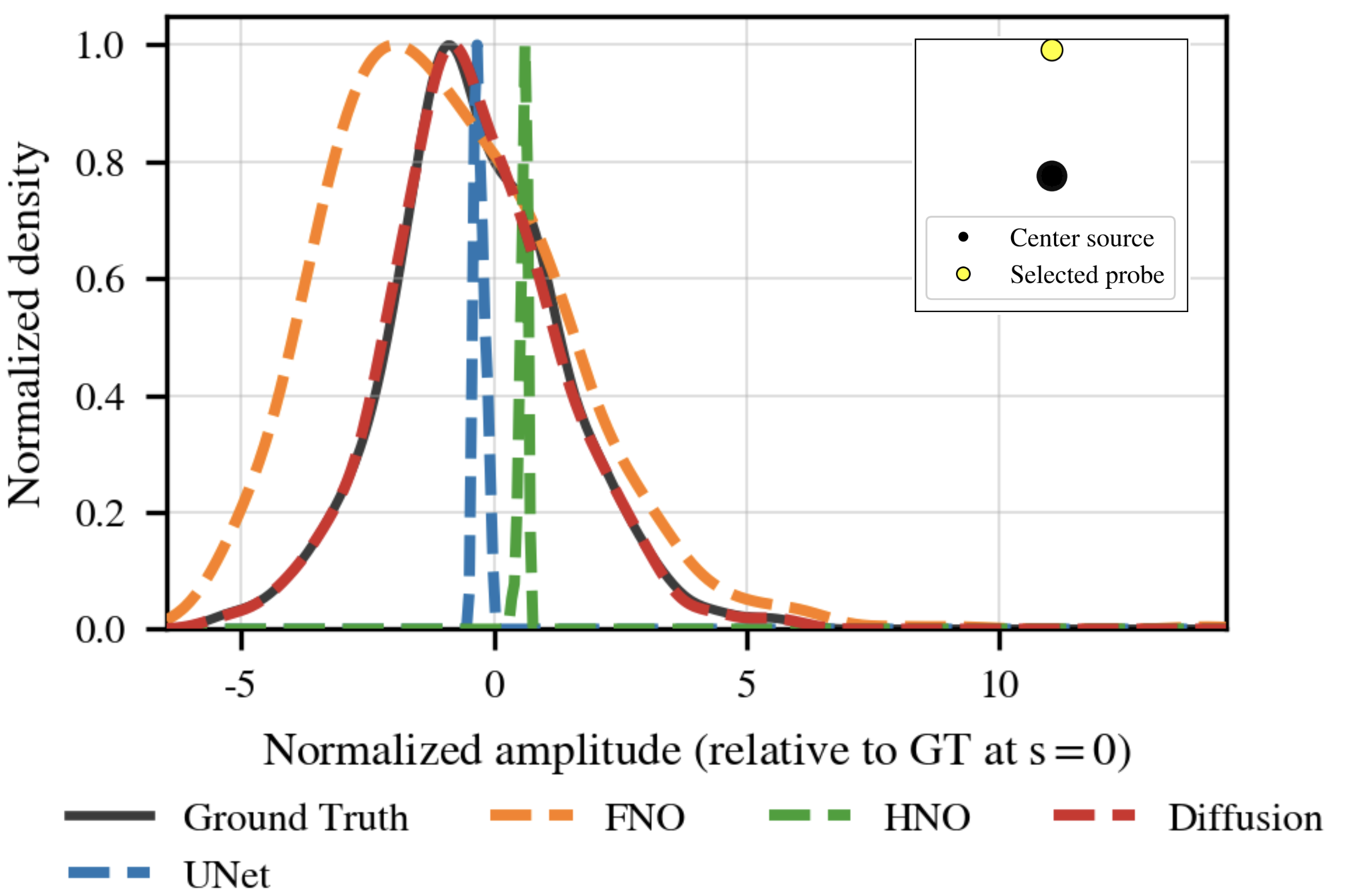}
  \end{subfigure}

  \vspace{2pt}

  \begin{subfigure}[t]{0.48\linewidth}
    \centering
    \includegraphics[width=\linewidth,height=0.21\textheight,keepaspectratio]{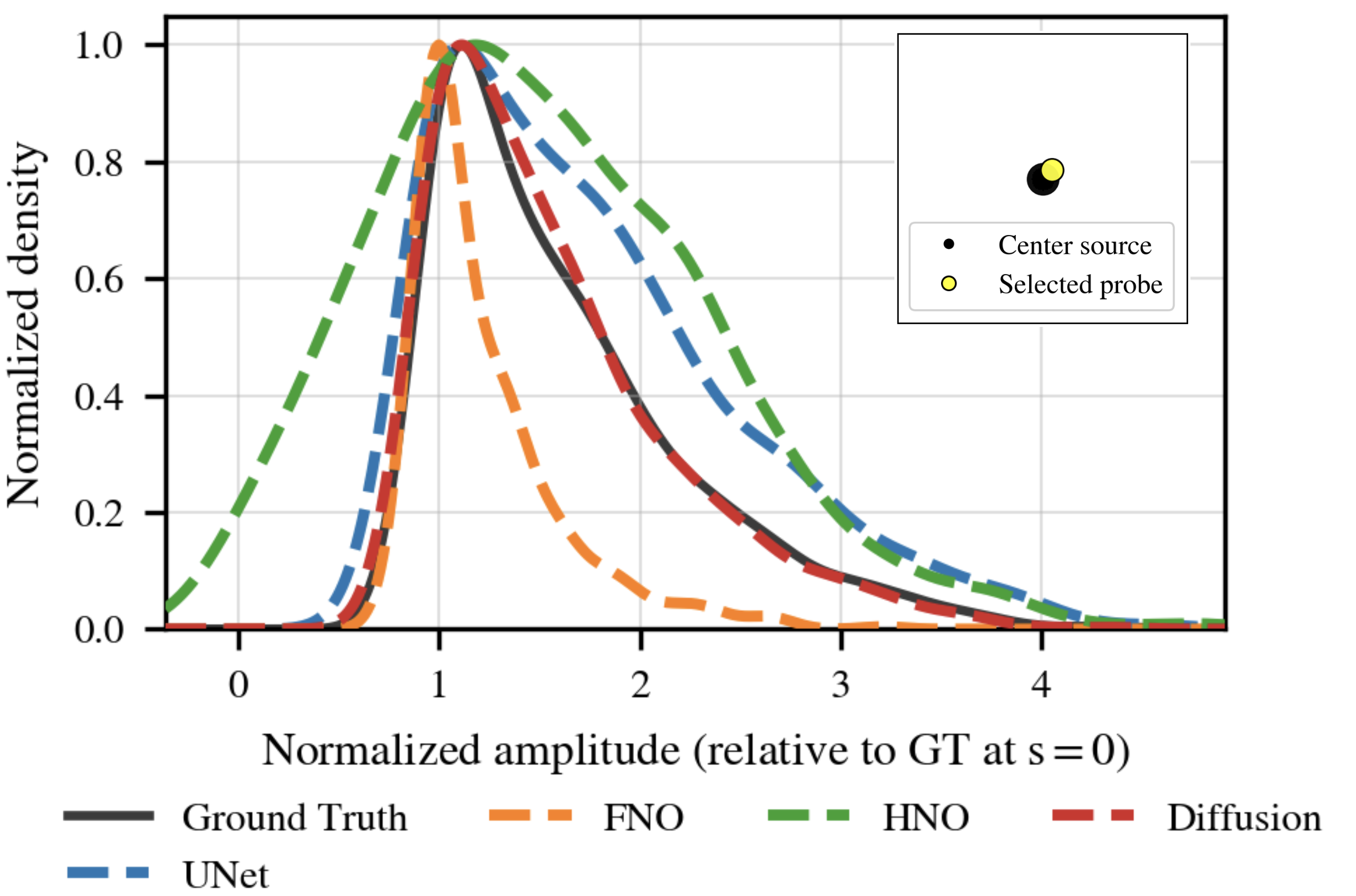}
  \end{subfigure}\hfill
  \begin{subfigure}[t]{0.48\linewidth}
    \centering
    \includegraphics[width=\linewidth,height=0.21\textheight,keepaspectratio]{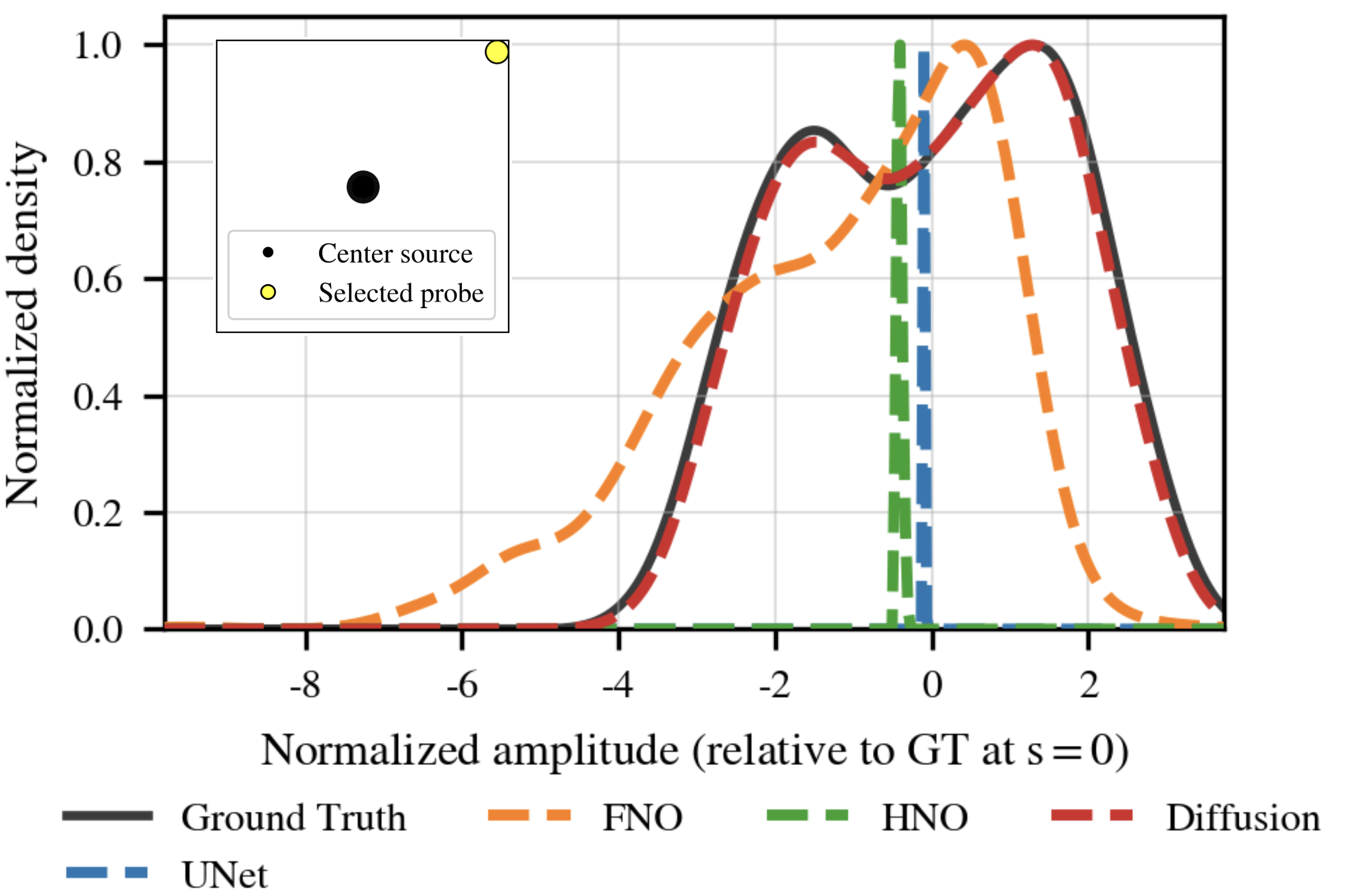}
  \end{subfigure}

  \vspace{2pt}

  \begin{subfigure}[t]{0.48\linewidth}
    \centering
    \includegraphics[width=\linewidth,height=0.21\textheight,keepaspectratio]{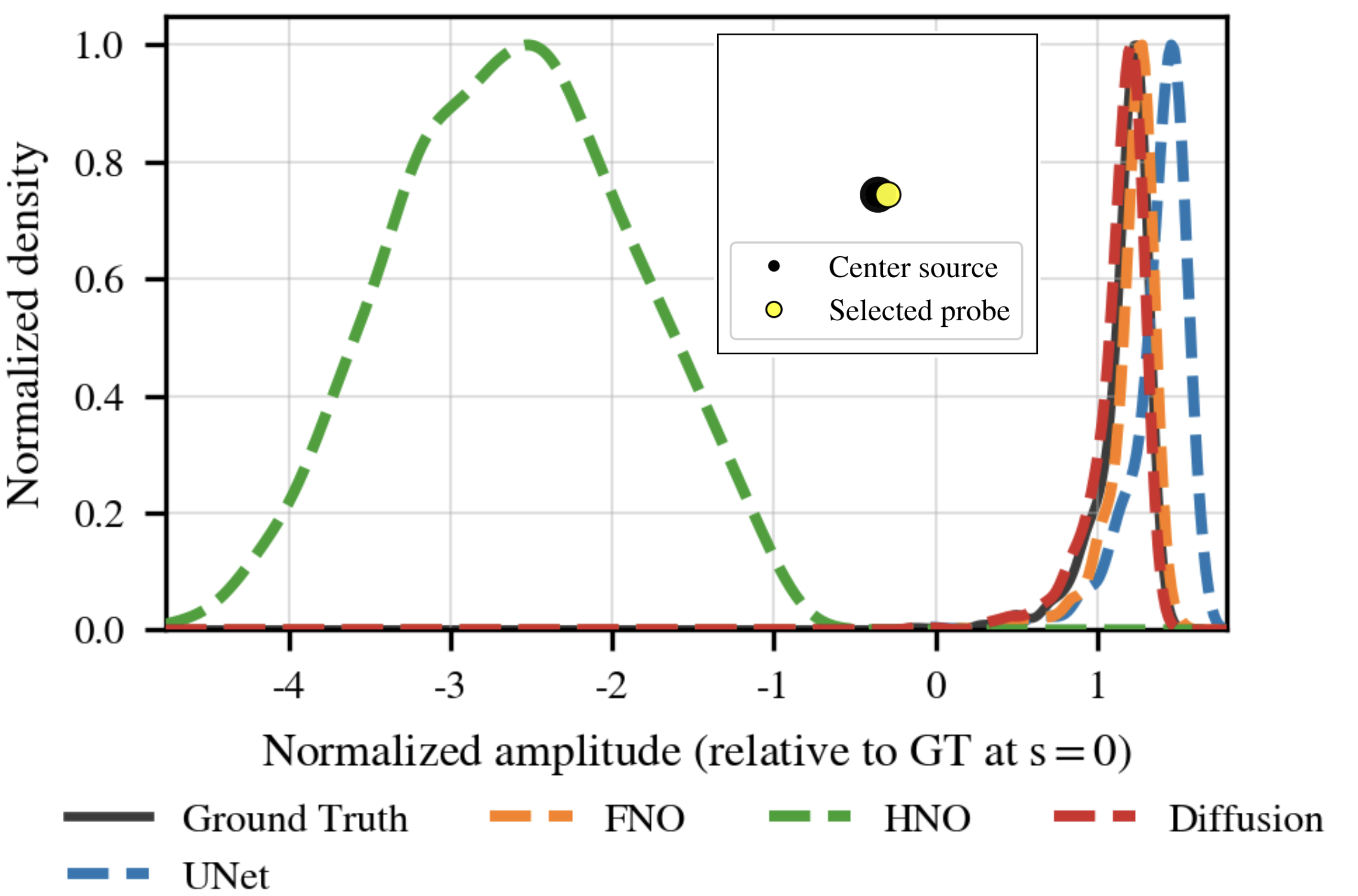}
    \caption*{Near source}
  \end{subfigure}\hfill
  \begin{subfigure}[t]{0.48\linewidth}
    \centering
    \includegraphics[width=\linewidth,height=0.21\textheight,keepaspectratio]{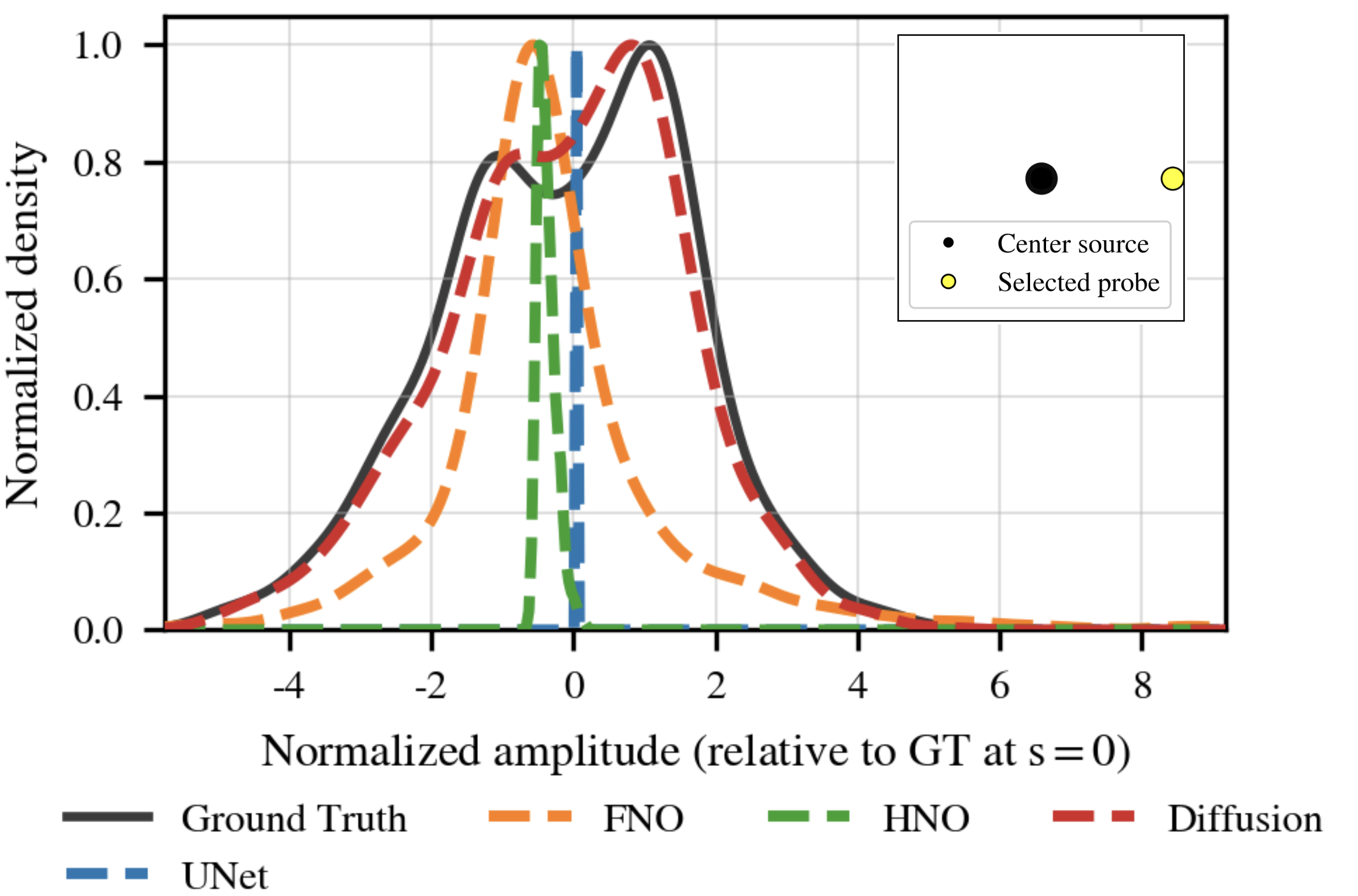}
    \caption*{Near boundary}
  \end{subfigure}

  \caption{\textbf{Kernel density estimates across directions for \(\mathbf{s{=}0.1}\) (all 4 near vs.\ all 4 far).}
  Left column: near the source. Right column: near the boundary.}
  \label{fig:s1-all}
\end{figure}

\begin{figure}[t]
  \centering
  \captionsetup{skip=2pt}
  \captionsetup[sub]{font=small}

  \begin{subfigure}[t]{0.48\linewidth}
    \centering
    \includegraphics[width=\linewidth,height=0.21\textheight,keepaspectratio]{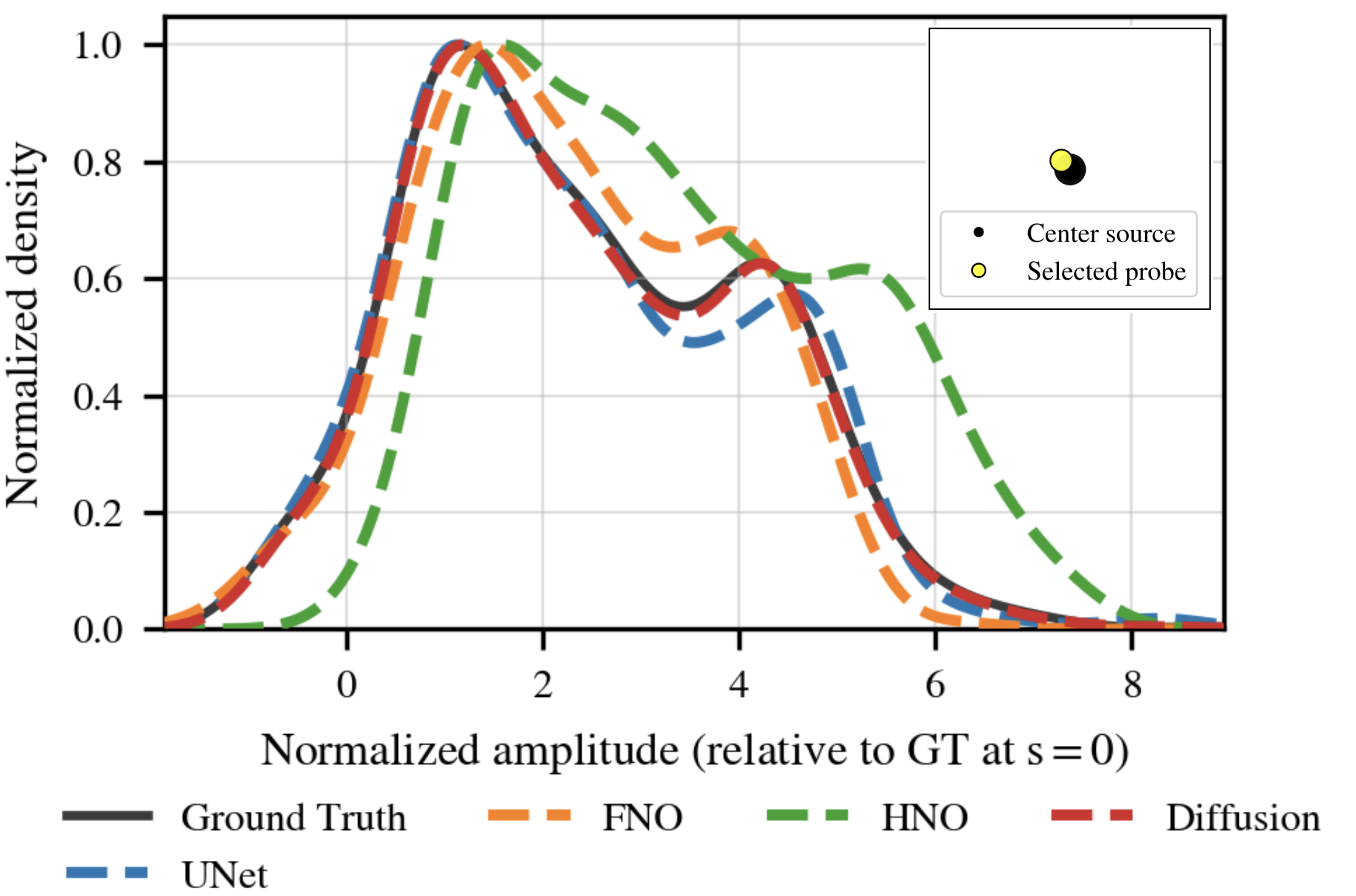}
  \end{subfigure}\hfill
  \begin{subfigure}[t]{0.48\linewidth}
    \centering
    \includegraphics[width=\linewidth,height=0.21\textheight,keepaspectratio]{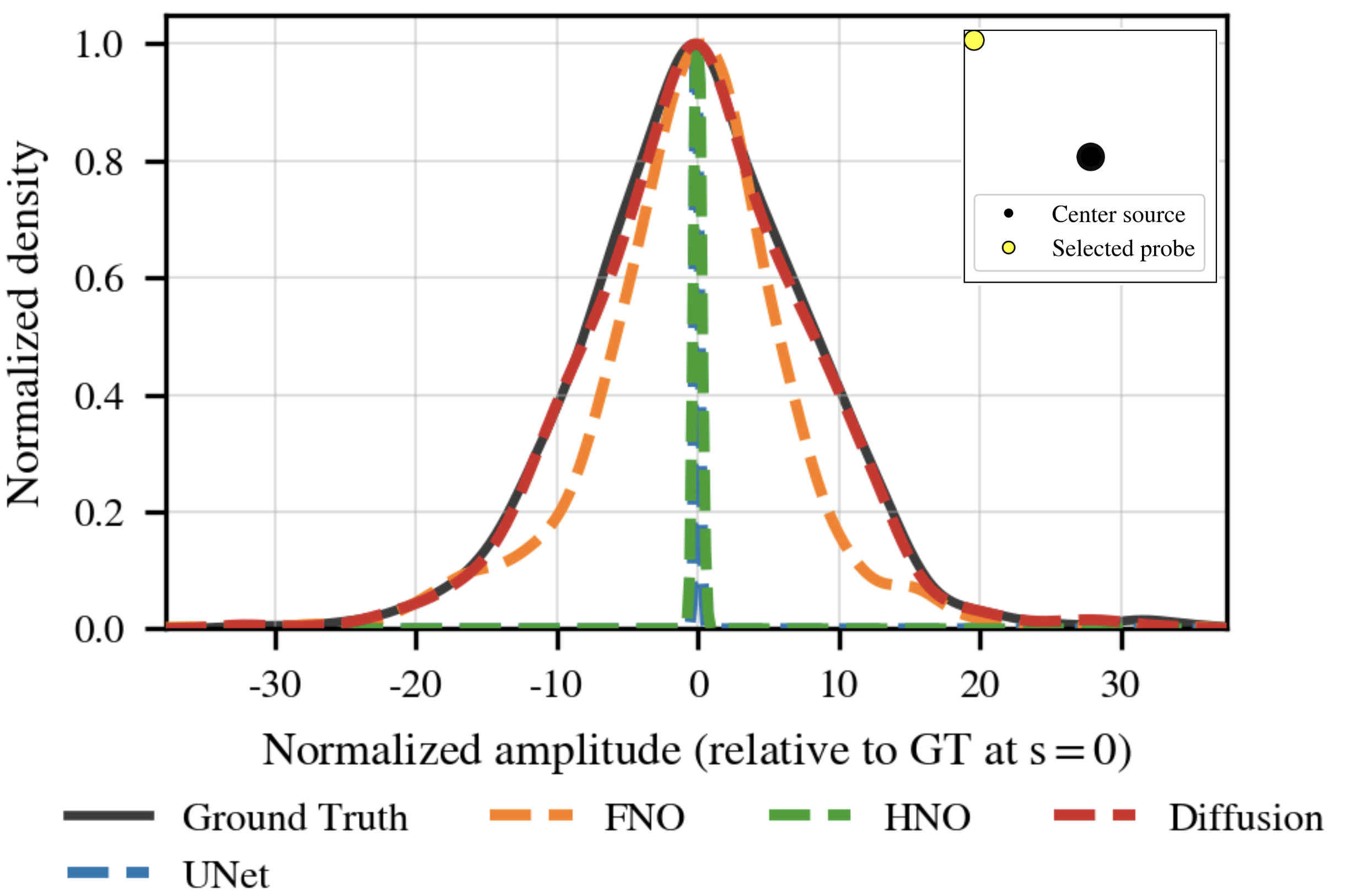}
  \end{subfigure}

  \vspace{2pt}

  \begin{subfigure}[t]{0.48\linewidth}
    \centering
    \includegraphics[width=\linewidth,height=0.21\textheight,keepaspectratio]{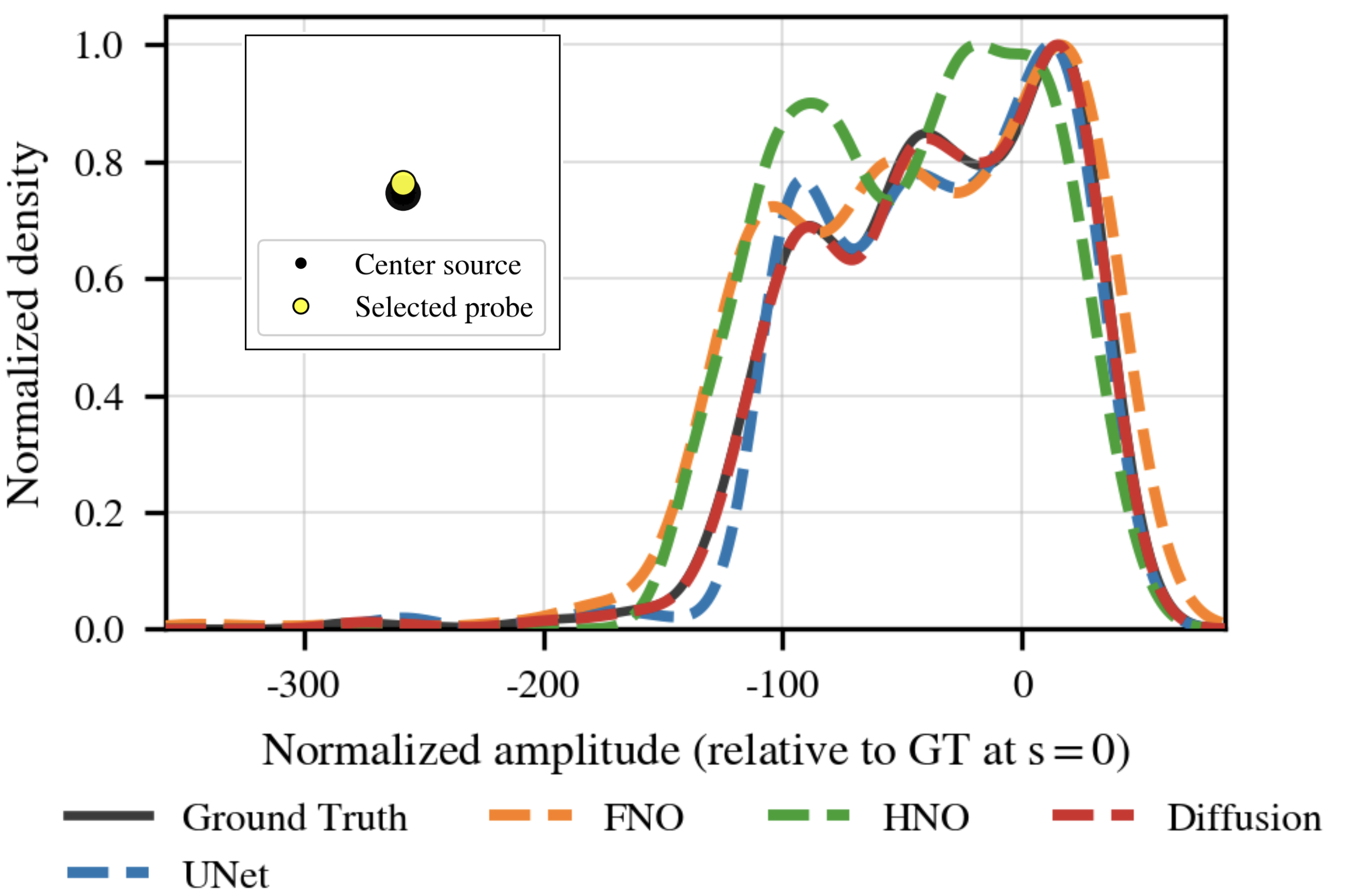}
  \end{subfigure}\hfill
  \begin{subfigure}[t]{0.48\linewidth}
    \centering
    \includegraphics[width=\linewidth,height=0.21\textheight,keepaspectratio]{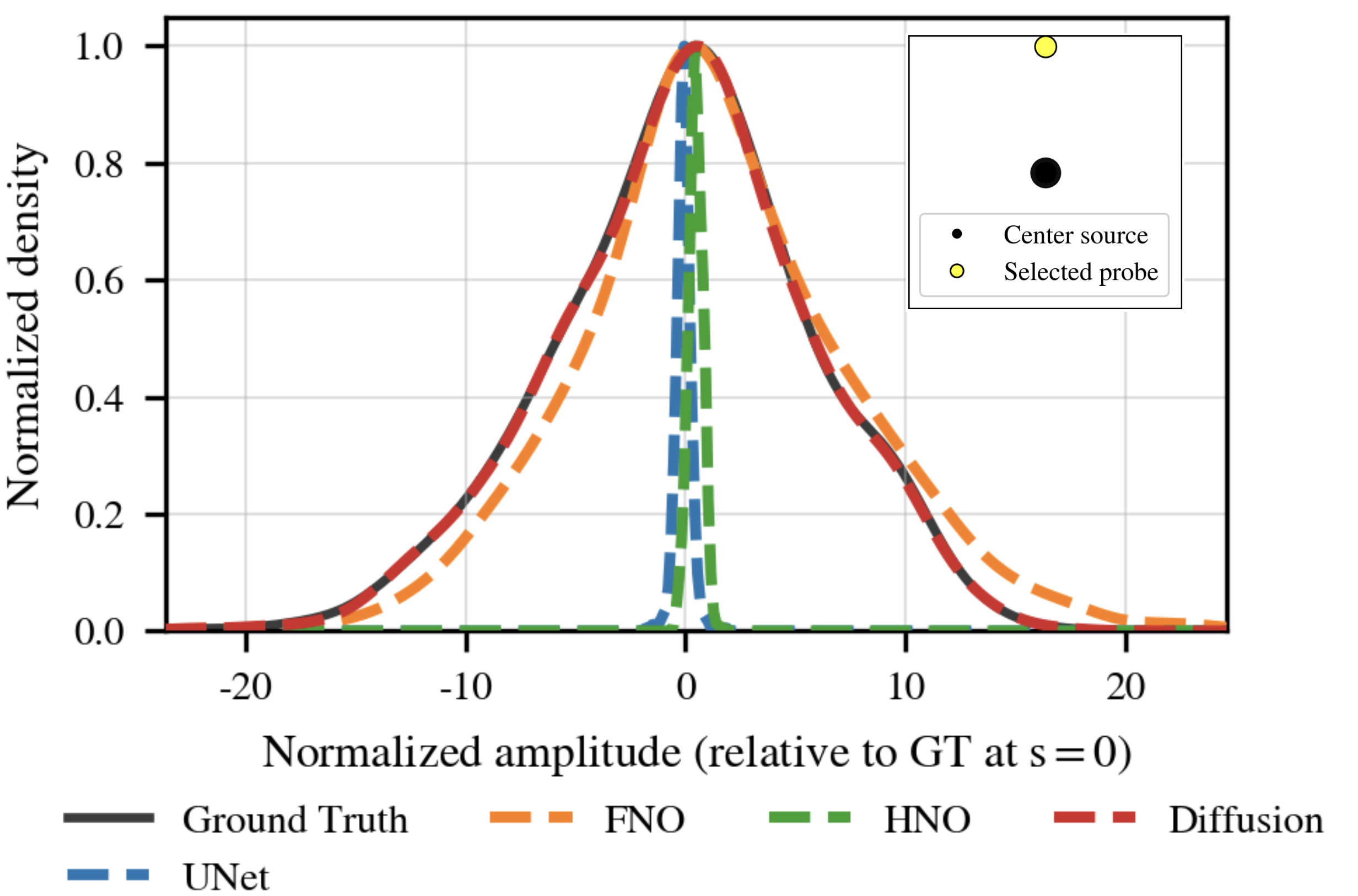}
  \end{subfigure}

  \vspace{2pt}

  \begin{subfigure}[t]{0.48\linewidth}
    \centering
    \includegraphics[width=\linewidth,height=0.21\textheight,keepaspectratio]{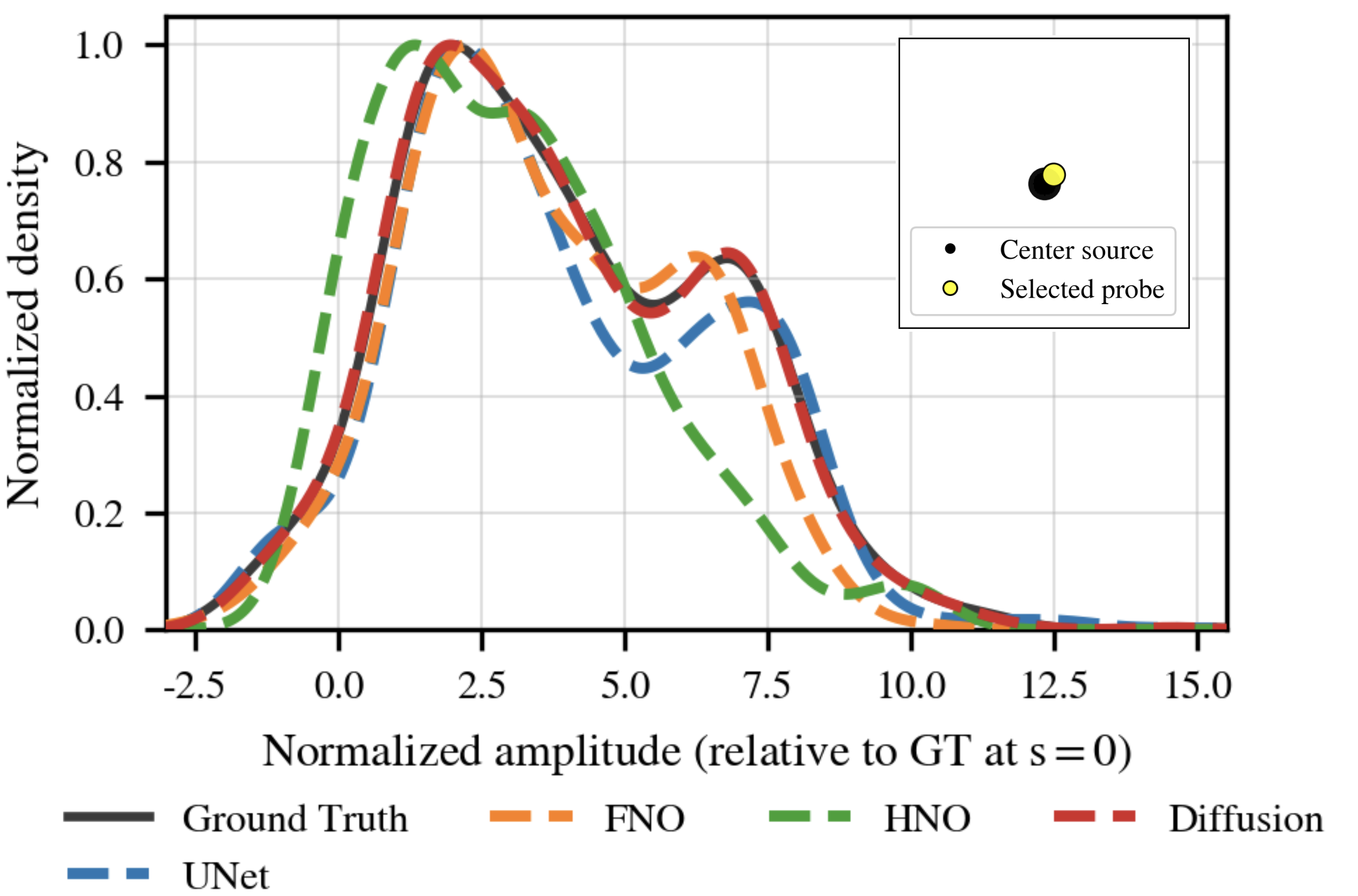}
  \end{subfigure}\hfill
  \begin{subfigure}[t]{0.48\linewidth}
    \centering
    \includegraphics[width=\linewidth,height=0.21\textheight,keepaspectratio]{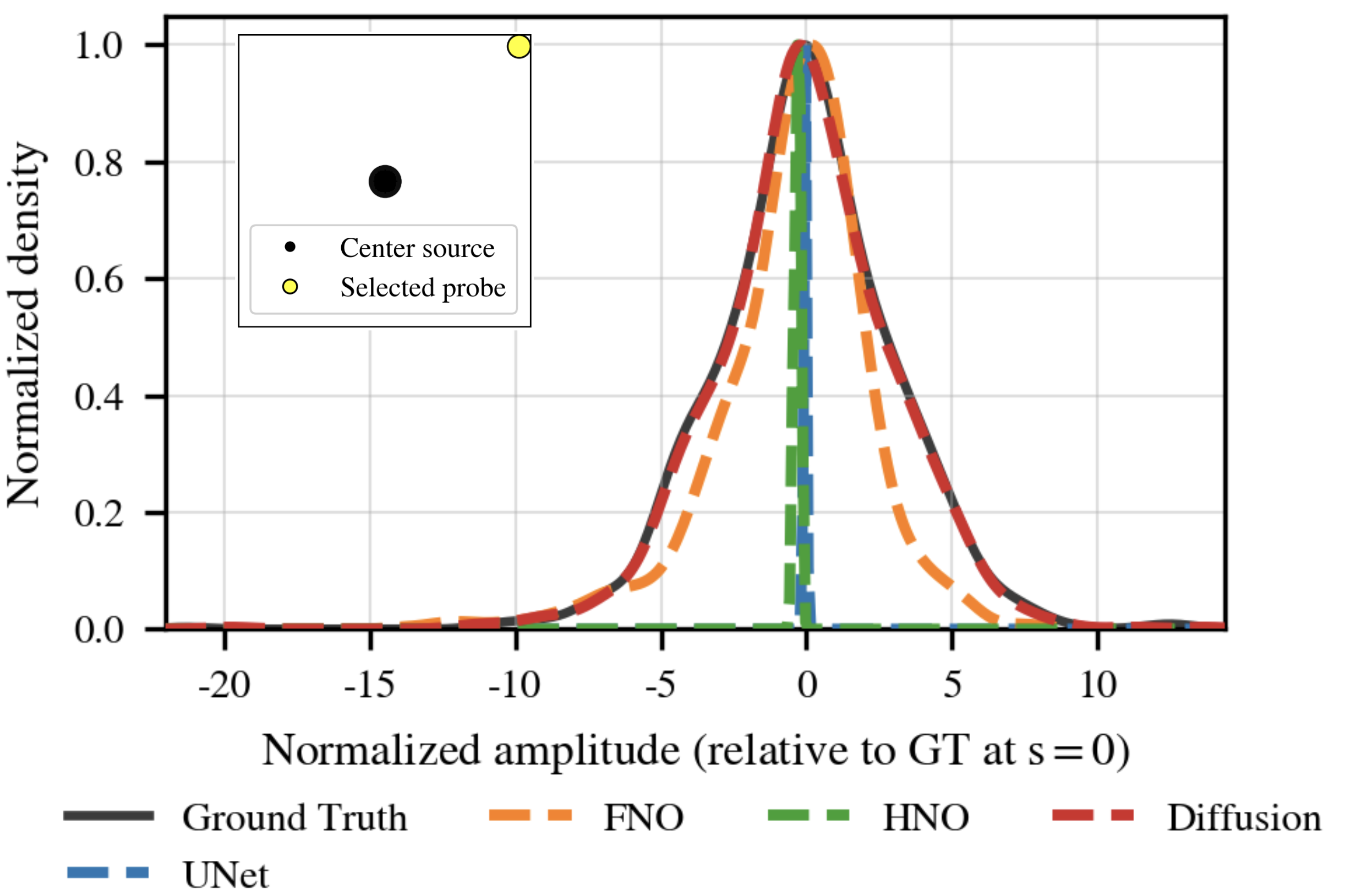}
  \end{subfigure}

  \vspace{2pt}

  \begin{subfigure}[t]{0.48\linewidth}
    \centering
    \includegraphics[width=\linewidth,height=0.21\textheight,keepaspectratio]{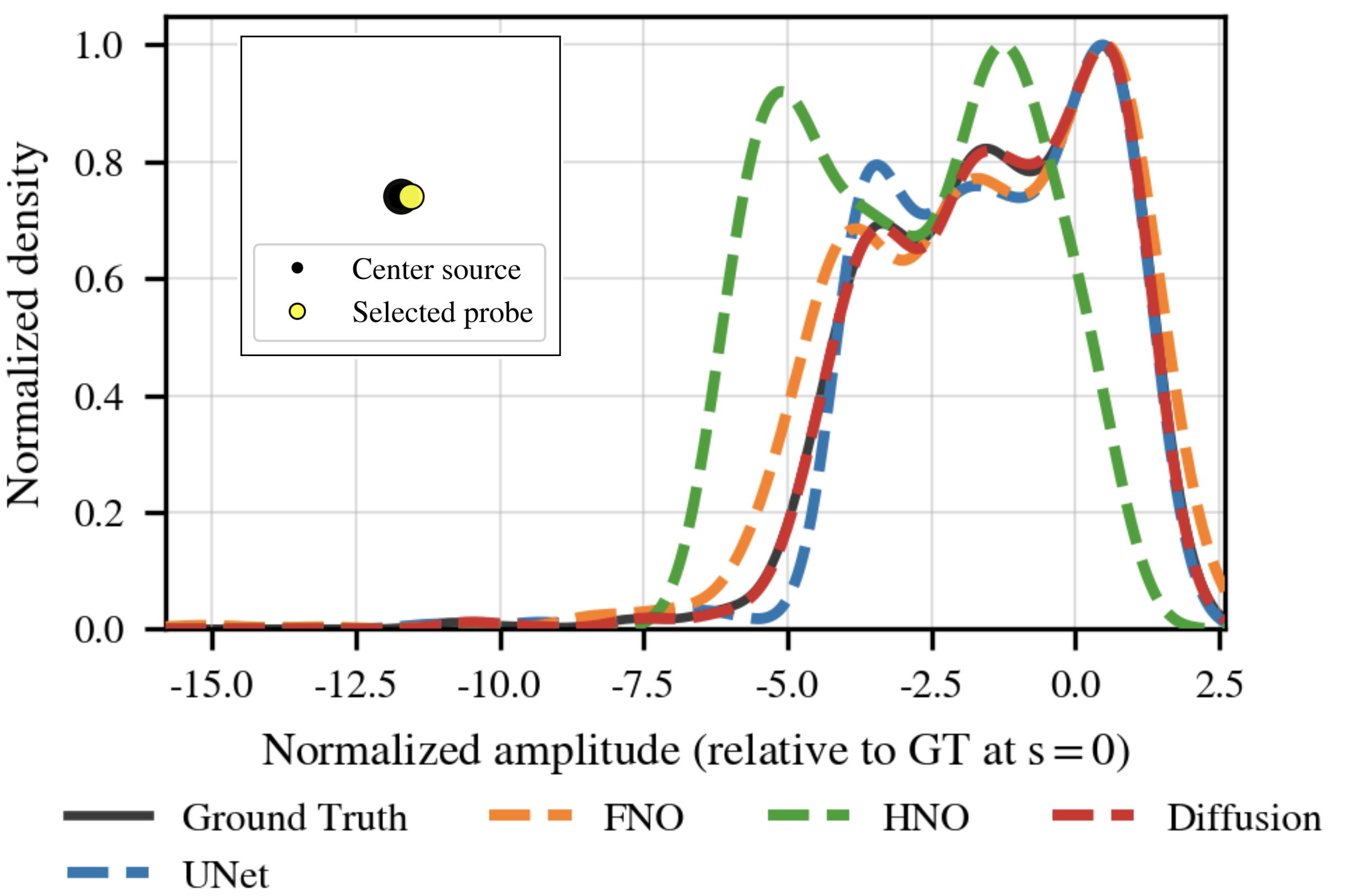}
    \caption*{Near source}
  \end{subfigure}\hfill
  \begin{subfigure}[t]{0.48\linewidth}
    \centering
    \includegraphics[width=\linewidth,height=0.21\textheight,keepaspectratio]{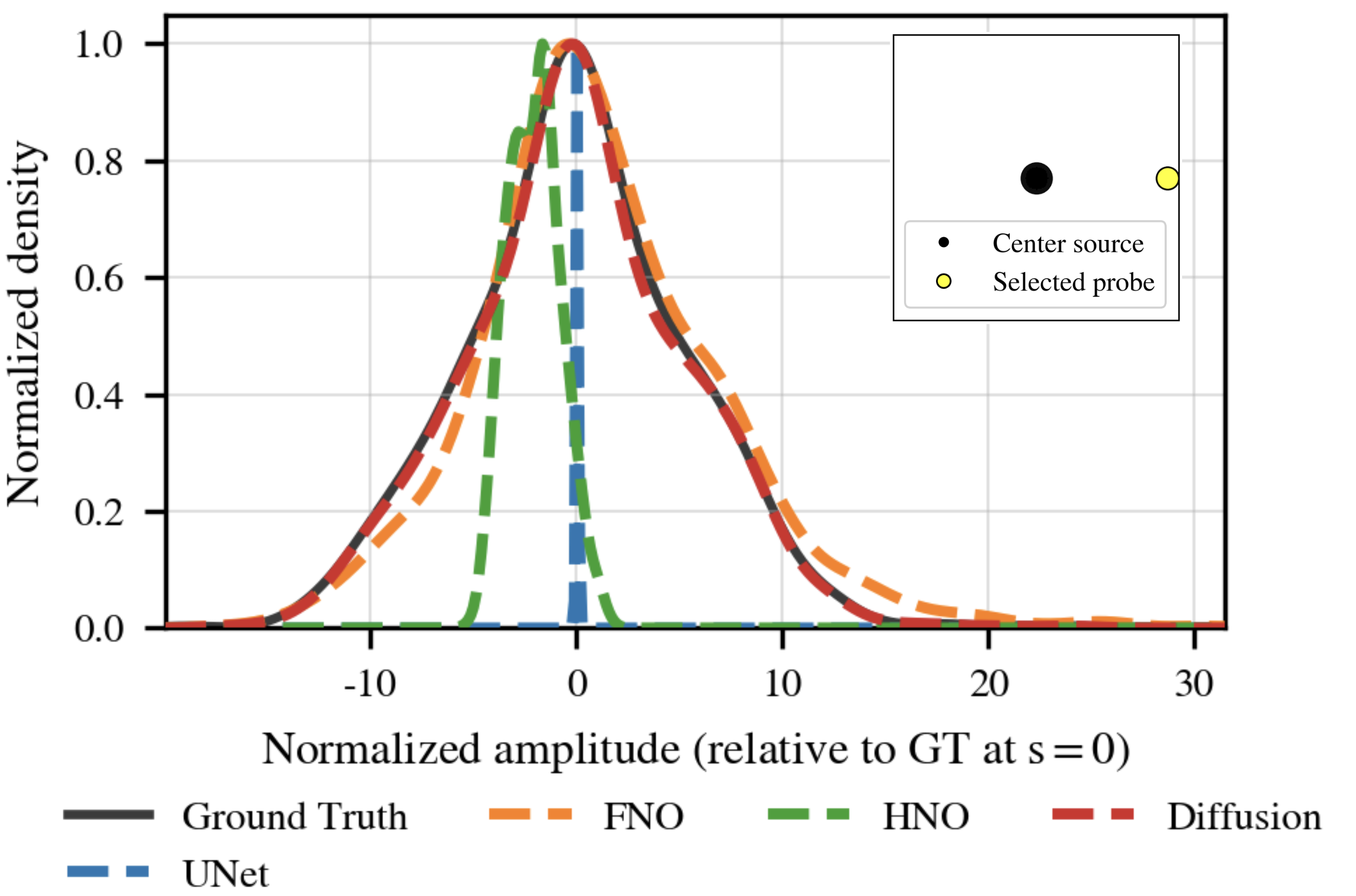}
    \caption*{Near boundary}
  \end{subfigure}

  \caption{\textbf{Kernel density estimates across directions for \(\mathbf{s{=}1}\) (all 4 near vs.\ all 4 far).}
  Left column: near the source. Right column: near the boundary.}
  \label{fig:s9-all}
\end{figure}

In the \emph{density plot} view, at $s{=}0$ all inputs satisfy $c^{(d)}(0)=c_0$, so the solver produces a single spike; deterministic models also produce a spike but it is \emph{biased} relative to the solver’s value, whereas diffusion yields a narrow yet non-degenerate predictive distribution that \emph{covers} the ground-truth spike—reflecting the fact that at high frequency the map $c\!\mapsto\!u$ is effectively one-to-many (phase ambiguity, discretization/model mismatch, and measurement noise make several wavefields plausible for the same inputs). We turn this stochasticity to our benefit as diffusion model learns $p(u\mid x)$ rather than a single point, quantifying the push-forward distribution instead of committing to a potentially biased mean (\cref{fig:s0-all,fig:s1-all,fig:s9-all}). At $s{=}1$, only diffusion captures the ground truth’s broad amplitude distribution across directions, again consistent with the qualitative boundary behavior (\cref{fig:s0-all,fig:s1-all,fig:s9-all,fig:qual-multifreq-2col}). We also note that for all sensitivity plots we use one diffusion sample per input to expose the predicted distribution rather than its mean.

\clearpage

\clearpage

\subsection{Preliminary Extensions to 3D Helmholtz}
Our framework is readily extendable to three-dimensional problem, which we explore in this subsection. The computational challenges arising in three dimensions demand a more efficient diffusion framework. We thus leverage Vision Transformers as the backbone of our diffusion model to demonstrate the superiority of our probabilistic approach, qunatified by our error analysis. 

\begin{figure}[H]
  \centering

  \includegraphics[width=\textwidth]{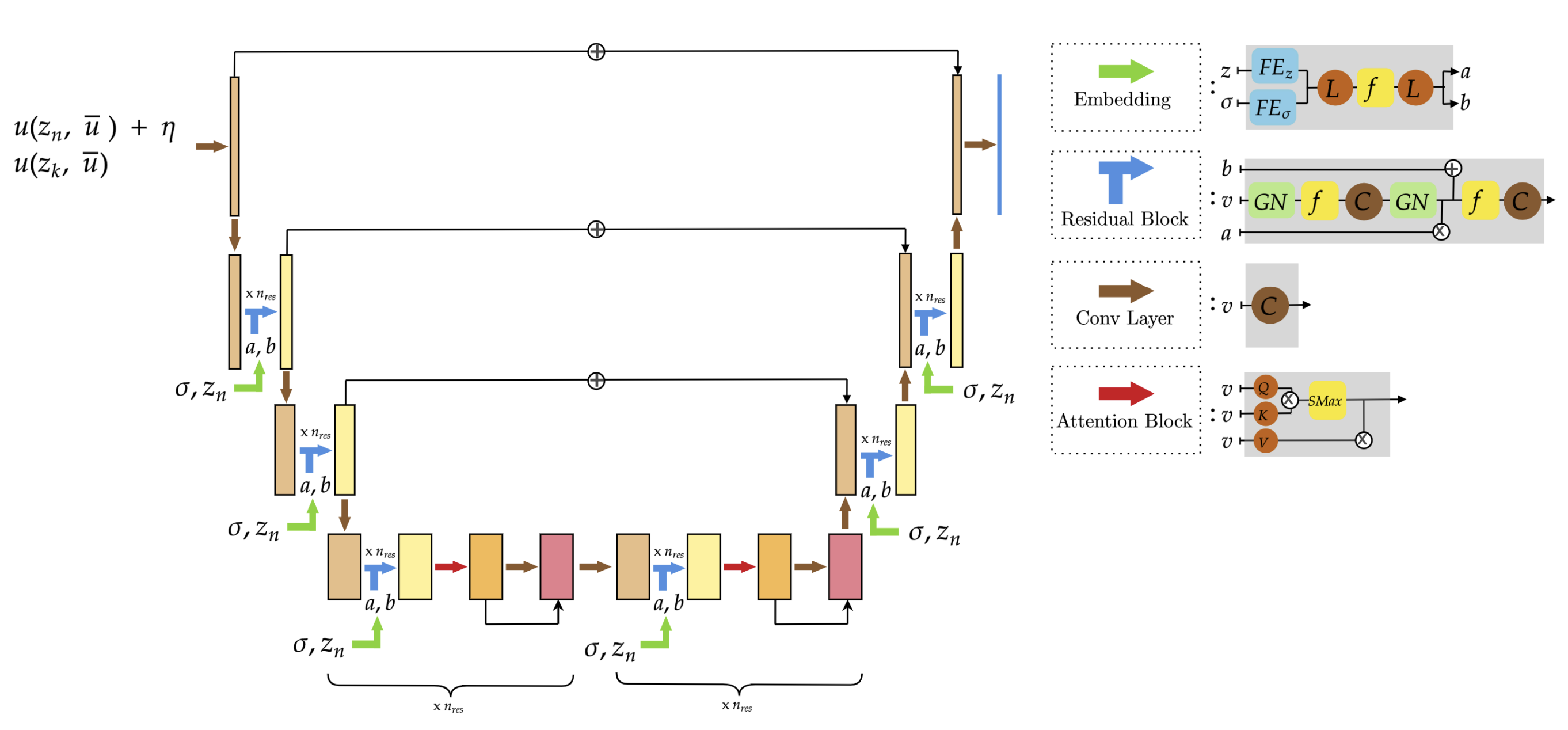}
  \caption{Architecture of the conditional UViT backbone used in our diffusion model.}
  \label{fig:uvit_arch}
\end{figure}

\subsubsection{Network backbone -- UViT}\label{app:3d}
After establishing our main results in 2D, we next extend the framework to 3D Helmholtz wavefields. A key methodological change in this transition is the backbone used within the diffusion model: whereas many prior diffusion-based surrogates rely on U\mbox{-}Net-style CNN backbones, we adopt a U-shaped vision transformer (UViT) to better capture long-range spatial dependencies and global interference patterns that become more pronounced in 3D. Due to the substantially higher data and computational complexity in 3D, we treat these experiments as \emph{preliminary}: in this section we focus on \emph{error analysis} only, and defer a full sensitivity study (as in 2D) to future work.

Our UViT backbone, illustrated in Fig.~\ref{fig:uvit_arch}, is a U-shaped vision transformer inspired by~\cite{molinaro2024_gencfd} and adapted here for Helmholtz wavefields. The network processes multi-scale feature maps in an encoder--decoder hierarchy. At each resolution level, it applies stacks of residual blocks (blue), optionally interleaved with self-attention blocks (red) at selected scales. The downsampling and upsampling paths are linked by skip connections, as in a standard U\mbox{-}Net, to preserve fine-scale information while enabling global context aggregation.

\begin{table}[H]
\centering

\scriptsize
\setlength{\tabcolsep}{4pt}
\caption{\textbf{3D Helmholtz errors on $64^3$ grids.} Relative $L^2$, $H^1$, and energy errors (mean $\pm$ std, clamped to $\leq 1$) over $N=500$ test samples for our frequencies. Diffusion--UViT consistently outperforms Diffusion--UNet and the vanilla U-Net.}
\label{tab:3d_errors}
\begin{tabular}{llccc}
\toprule
\textbf{Frequency (Hz)} & \textbf{Metric} & \textbf{Diffusion--UNet} & \textbf{Diffusion--UViT} & \textbf{U-Net only} \\
\midrule
\multirow{3}{*}{$5\times 10^{5}$}
  & $L^2$   & $0.030$ & $\mathbf{0.013}$ & $0.180$ \\
  & $H^1$   & $0.050$ & $\mathbf{0.019}$ & $0.287$ \\
  & Energy  & $0.041$ & $\mathbf{0.019}$ & $0.102$ \\
\midrule
\multirow{3}{*}{$1.0\times 10^{6}$}
  & $L^2$   & $0.032$ & $\mathbf{0.011}$ & $0.295$ \\
  & $H^1$   & $0.049$ & $\mathbf{0.016}$ & $0.455$ \\
  & Energy  & $0.037$ & $\mathbf{0.012}$ & $0.170$ \\
\midrule
\multirow{3}{*}{$1.5\times 10^{6}$}
  & $L^2$   & $0.063$ & $\mathbf{0.016}$ & $0.874$ \\
  & $H^1$   & $0.091$ & $\mathbf{0.024}$ & $1.000$ \\
  & Energy  & $0.101$ & $\mathbf{0.014}$ & $0.752$ \\
\midrule
\multirow{3}{*}{$2.5\times 10^{6}$}
  & $L^2$   & $0.159$ & $\mathbf{0.078}$ & $0.967$ \\
  & $H^1$   & $0.229$ & $\mathbf{0.112}$ & $1.000$ \\
  & Energy  & $0.178$ & $\mathbf{0.049}$ & $0.793$ \\
\bottomrule
\end{tabular}
\end{table}

Conditioning on the diffusion time \(\sigma\) and the latent code \(z_n\) (green arrows in Fig.~\ref{fig:uvit_arch}) is implemented via feature-wise affine modulation. Specifically, \((z_n,\sigma)\) are embedded by Fourier encoders \(\mathrm{FE}_z\) and \(\mathrm{FE}_\sigma\), then mapped by linear layers to scale and shift vectors \((a,b)\). Within each residual block, these parameters modulate the normalized features through a FiLM-style transformation \(a\odot v + b\), where \(v\) denotes the current feature map. Residual blocks consist of GroupNorm (GN), pointwise nonlinearities \(f\), and convolutional layers \(C\), as indicated on the right of Fig.~\ref{fig:uvit_arch}. Attention blocks use standard multi-head self-attention with query--key--value projections \((Q,K,V)\) and a softmax operator, applied to flattened spatial tokens at the corresponding resolution.

Overall, this backbone combines the locality and multiscale structure of a U\mbox{-}Net with the long-range interactions and flexible conditioning of transformers, making it a strong architectural choice for diffusion modeling of 3D Helmholtz wavefields.

\subsubsection{3D results}
Having introduced the UViT backbone as our primary architectural upgrade for diffusion-based operators, we now evaluate these models on fully 3D Helmholtz problems. We consider $64^3$ grids and four frequencies,
\(5\times 10^{5}\), \(10^{6}\), \(1.5\times 10^{6}\), and \(2.5\times 10^{6}\,\mathrm{Hz}\).
We start from \(5\times 10^{5}\,\mathrm{Hz}\) because, in the 2D experiments, all models already performed comparably well at lower frequencies (e.g., \(1.5\times 10^{5}\) and \(2.5\times 10^{5}\,\mathrm{Hz}\)), whereas performance gaps became pronounced only as frequency increased. Focusing the 3D study on \(5\times 10^{5}\,\mathrm{Hz}\) and above therefore emphasizes the more challenging, high-frequency regime where phase errors and interference effects are amplified.

In this setting, we compare three models: a diffusion model with a U\mbox{-}Net backbone (\emph{Diffusion--UNet}), a diffusion model with a UViT backbone (\emph{Diffusion--UViT}), and a deterministic U\mbox{-}Net trained to predict the wavefield directly. We retain both diffusion variants to isolate the impact of the backbone architecture and directly assess the benefit of moving from a convolutional U\mbox{-}Net backbone to a transformer-based UViT backbone within the same probabilistic (diffusion) framework.

\begin{figure}[H]
  \centering

  \begin{subfigure}[b]{0.24\textwidth}
    \centering
    \includegraphics[width=\linewidth]{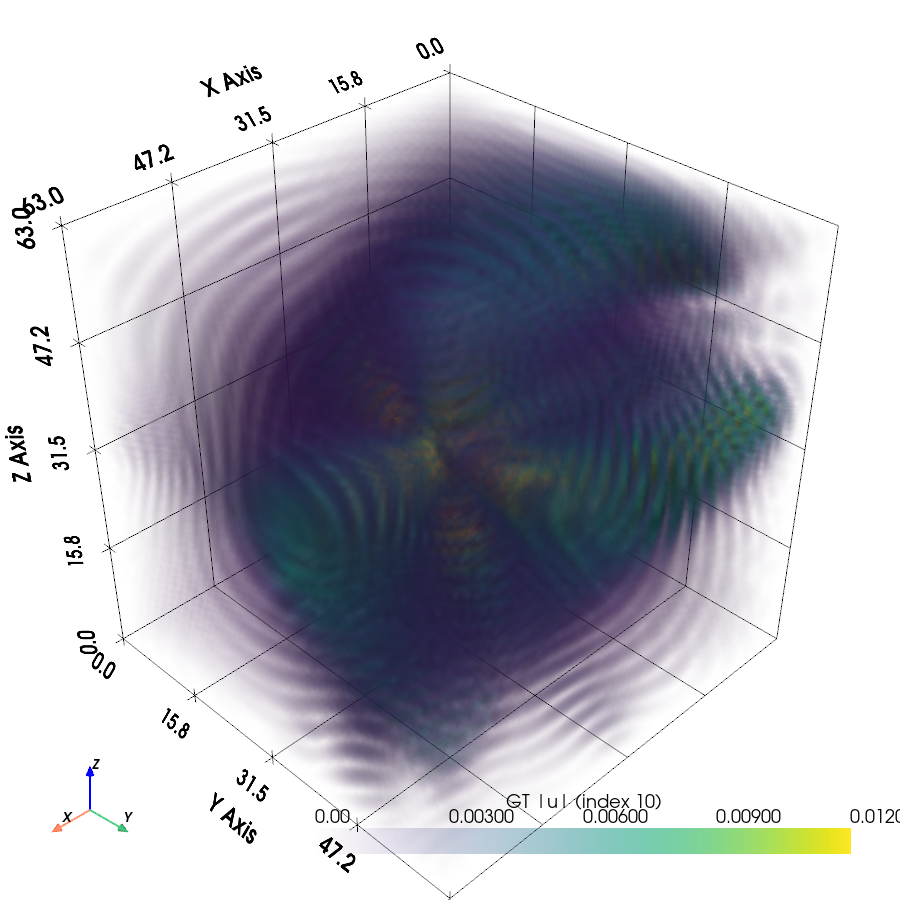}
    \caption{GT (full)}
  \end{subfigure}
  \begin{subfigure}[b]{0.24\textwidth}
    \centering
    \includegraphics[width=\linewidth]{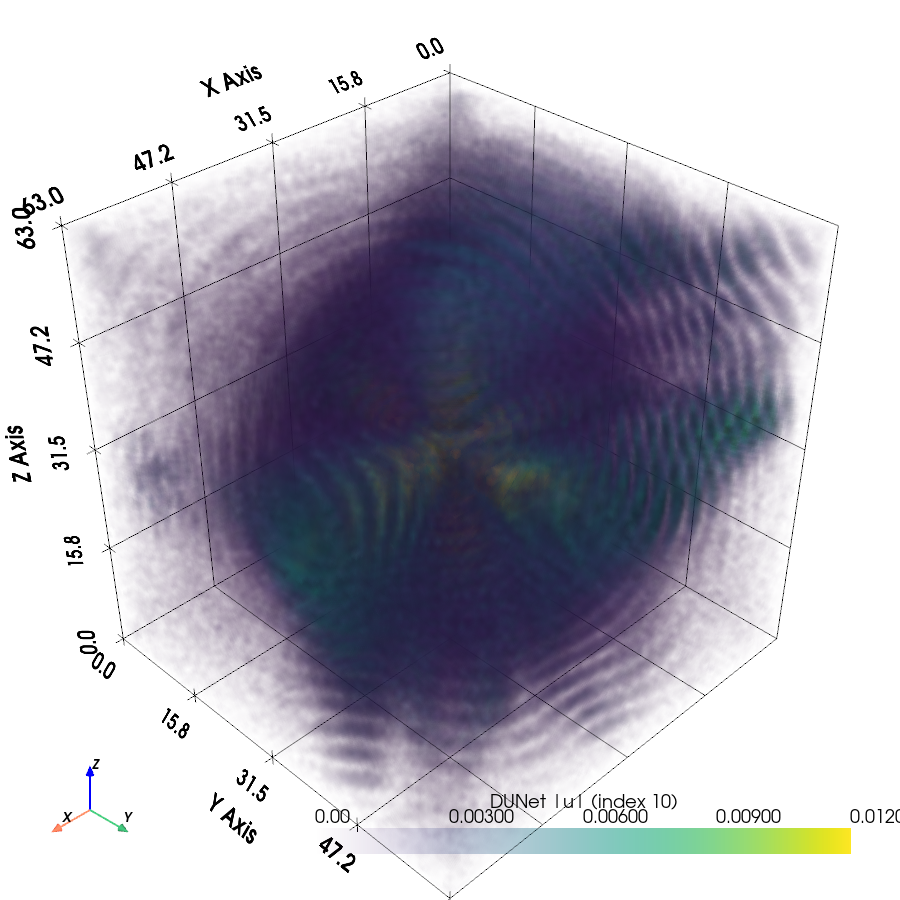}
    \caption{D-UNet (full)}
  \end{subfigure}
  \begin{subfigure}[b]{0.24\textwidth}
    \centering
    \includegraphics[width=\linewidth]{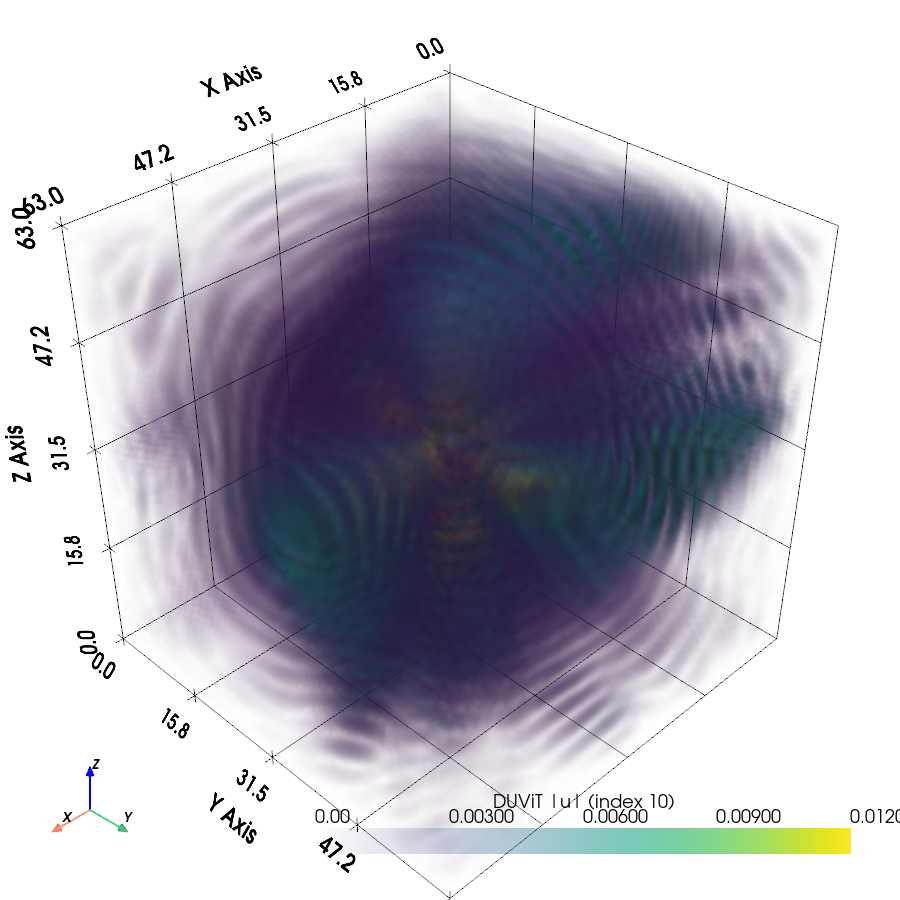}
    \caption{D-UViT (full)}
  \end{subfigure}
  \begin{subfigure}[b]{0.24\textwidth}
    \centering
    \includegraphics[width=\linewidth]{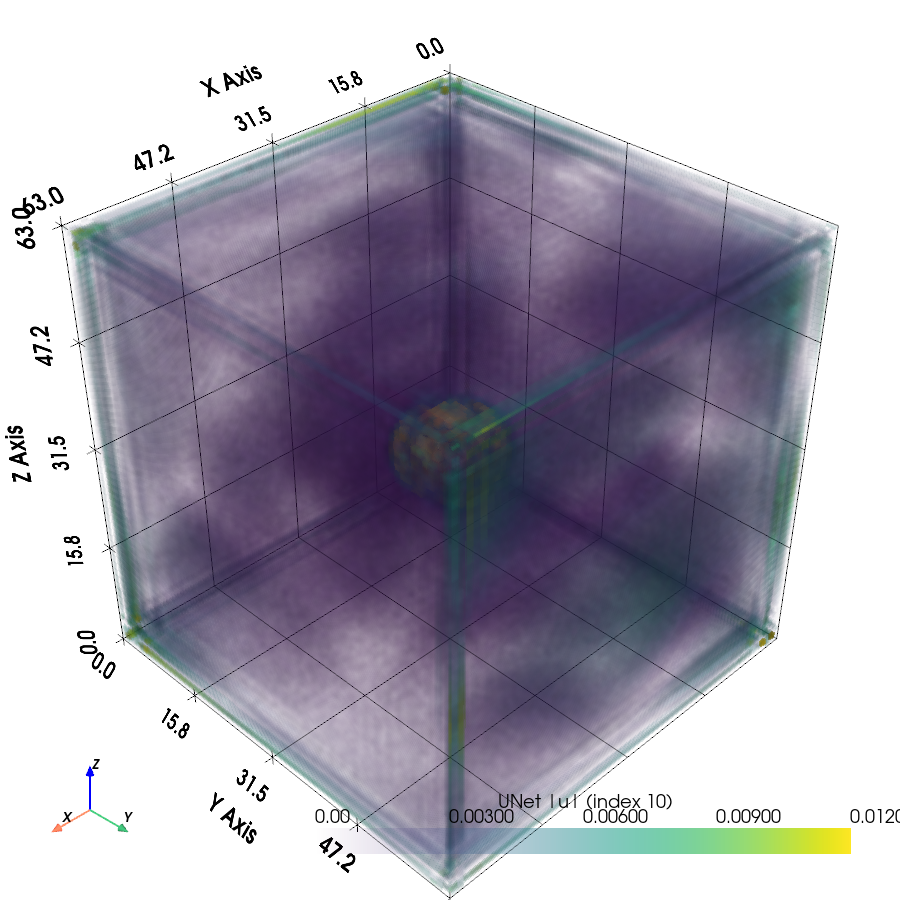}
    \caption{U-Net (full)}
  \end{subfigure}

  \vspace{0.5em}

  \begin{subfigure}[b]{0.24\textwidth}
    \centering
    \includegraphics[width=\linewidth]{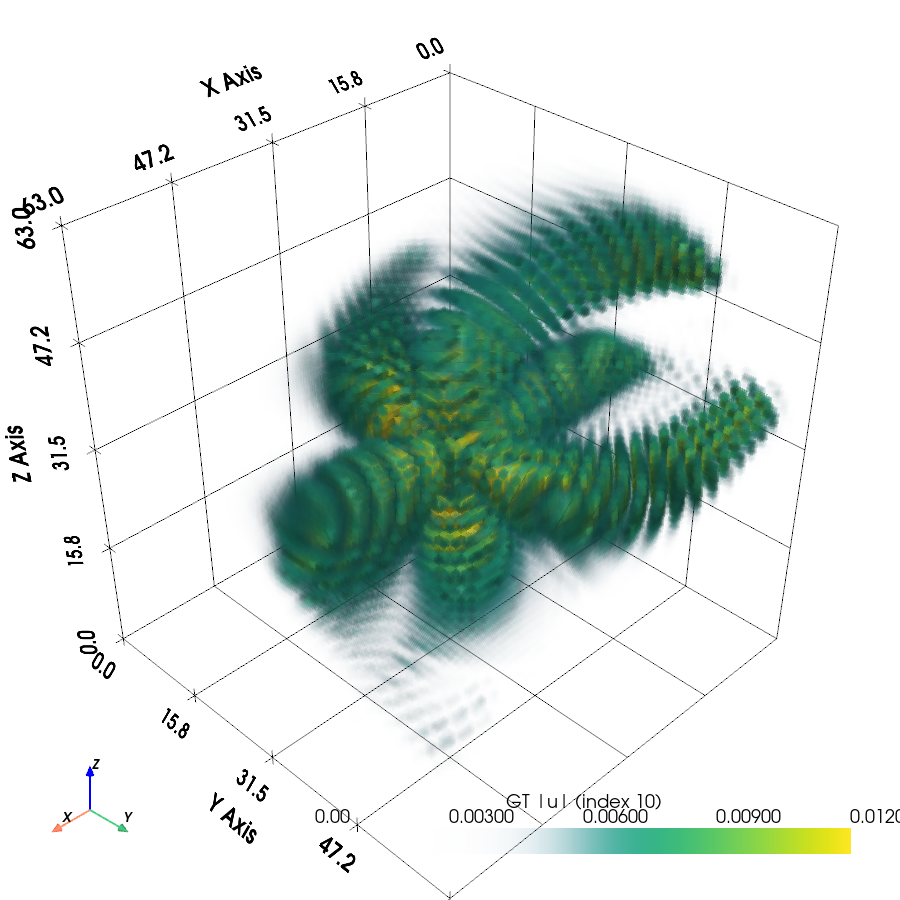}
    \caption{GT (thresholded)}
  \end{subfigure}
  \begin{subfigure}[b]{0.24\textwidth}
    \centering
    \includegraphics[width=\linewidth]{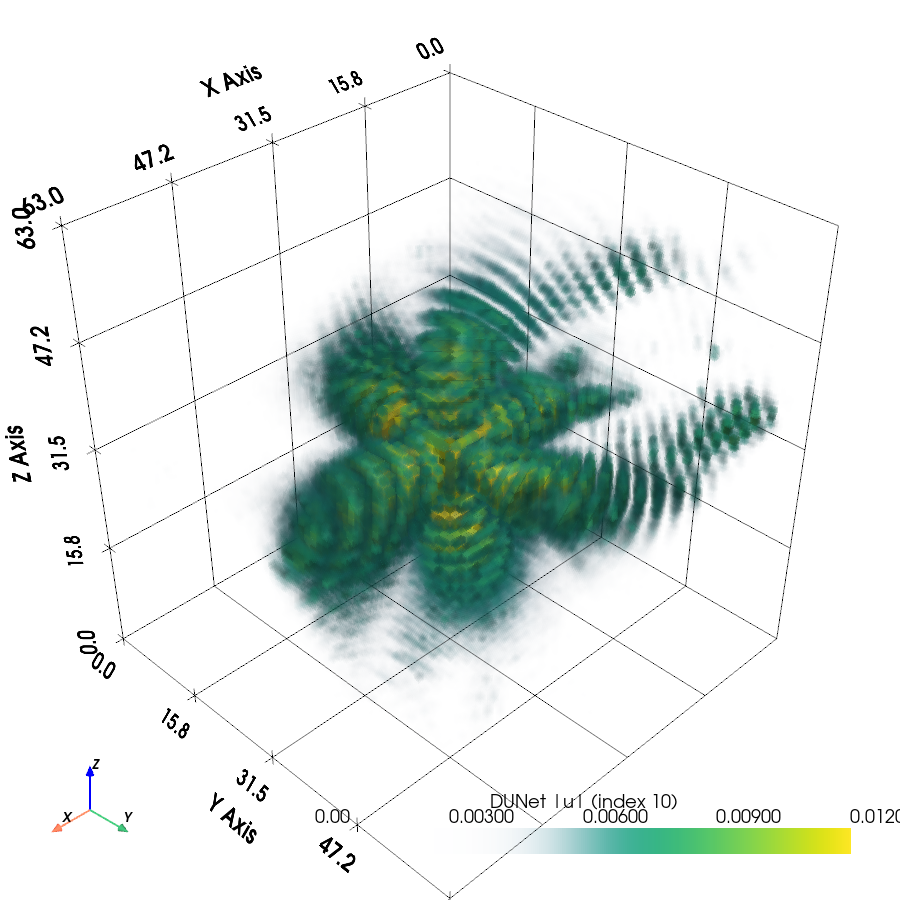}
    \caption{D-UNet (thresholded)}
  \end{subfigure}
  \begin{subfigure}[b]{0.24\textwidth}
    \centering
    \includegraphics[width=\linewidth]{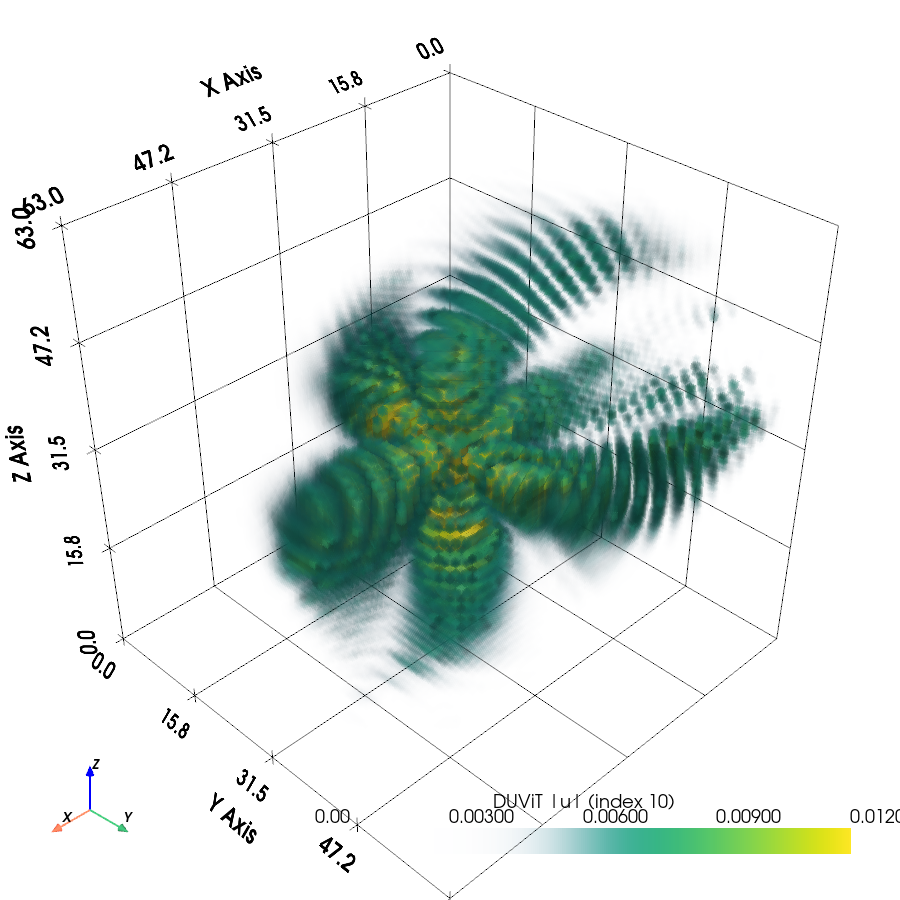}
    \caption{D-UViT (thresholded)}
  \end{subfigure}
  \begin{subfigure}[b]{0.24\textwidth}
    \centering
    \includegraphics[width=\linewidth]{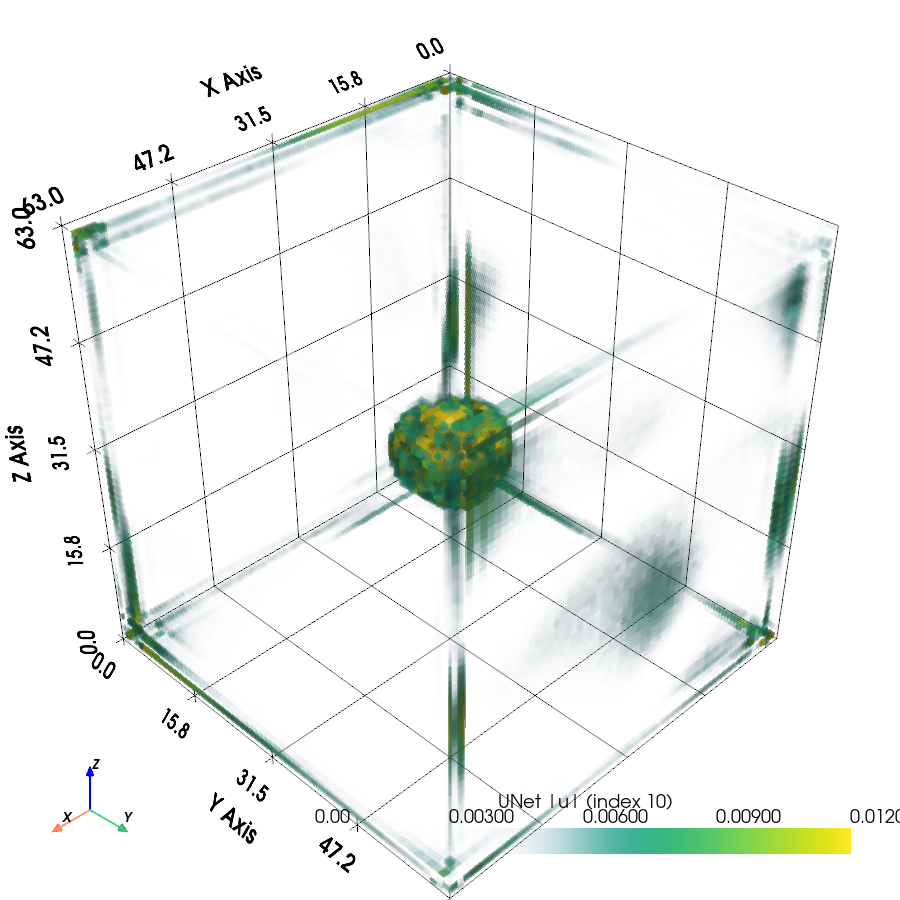}
    \caption{U-Net (thresholded)}
  \end{subfigure}

  \caption{Example 3D Helmholtz wavefield visualization using volume rendering from a fixed viewpoint: top row shows the full 3D visualization; bottom row uses a threshold to focus on center region to elaborate the differences.}
  \label{fig:3d_vis}
\end{figure}

Across all models, relative \(L^{2}\), \(H^{1}\), and energy errors increase with frequency, consistent with the 2D setting. However, \emph{Diffusion--UViT} achieves the lowest errors at every frequency, substantially outperforming both \emph{Diffusion--UNet} and the deterministic U\mbox{-}Net. The qualitative 3D renderings in Fig.~\ref{fig:3d_vis} corroborate these quantitative trends: the deterministic U\mbox{-}Net prediction remains overly concentrated near the domain center, whereas both diffusion models better recover wave structure toward the boundaries. Among them, \emph{Diffusion--UViT} yields the sharpest boundary reconstruction and the most faithful global interference pattern.

\section{Conclusion}

This paper presents a novel probabilistic machine-learning framework for solving the deterministic Helmholtz equation in high-frequency regimes. Using conditional diffusion models as probabilistic operators, we show that learning a distribution over wavefields yields substantially improved accuracy and robustness compared to deterministic neural operators, particularly as frequency increases. Across all tested frequencies, diffusion achieves lower $L^2$, $H^1$, and energy errors, with the performance gap widening markedly in the highest-frequency cases (Table~\ref{tab:all_by_freq}). Qualitative comparisons further show that diffusion preserves interference patterns and far-field structure that deterministic surrogates systematically smooth away (Fig.~\ref{fig:qual-multifreq-2col}).

These results follow from the sensitivity structure of the Helmholtz equation at high frequency. As shown in Sec.~\ref{sec:background}, small perturbations in the sound-speed field accumulate into phase errors that scale with $k$ and distance from source, rendering the solution map sharply varying in high $k$ and far-field. In this regime, point-estimate surrogates trained with mean-squared error effectively average over phase variability, leading to attenuation of oscillations and loss of phase coherence even when the training data contain fully resolved solutions, as obtained in our U-Net, FNO, and HNO predictions. By contrast, diffusion models represent a conditional distribution over solutions rather than committing to a single output. This distinction becomes particularly important for quantities that remain stable under phase variability. In our experiments, energy-type functionals are consistently predicted with high accuracy by diffusion, even at frequencies where deterministic surrogates deteriorate significantly. Averaging energies over samples provides a reliable estimator precisely when pointwise complex fields become fragile.

The sensitivity experiments reinforce this interpretation. Along controlled perturbation paths in coefficient function space, diffusion tracks solver trajectories more faithfully, including non-monotone behavior induced by accumulated phase error, while deterministic models often flatten or drift away from the reference solution. When aggregating across perturbation directions, diffusion captures the widening pushforward variability of the solver response, whereas deterministic operators systematically miss components of the solution variability. Although the Helmholtz equation defines a deterministic map, these results demonstrate that high-frequency prediction in heterogeneous media is effectively one-to-many under unresolved variability and discretization mismatch, and that modeling a conditional law provides a more faithful surrogate.

An important direction to improve our framework is to reduce its computational cost. Achieving the highest accuracy requires many diffusion steps, which increases inference time relative to deterministic operators - though still at a fraction of the cost of direct solvers. Building upon the framework here, we envision accelerating the inference as a natural future work via flow matching training \citep{baldan2025flowpdes}, improved ODE/SDE solvers \citep{zhang2022fast}, or latent or patchwise diffusion. Beyond efficiency, extending the framework to more complex geometries, multi-source configurations, and fully three-dimensional propagation will further test its robustness. Our preliminary 3D results already suggest that combining probabilistic operators with architectures capable of capturing long-range interactions can substantially improve fidelity in this setting.

\section*{Funding}
This research was in part supported by the internal funding provided by the Duke University.

\section*{CRediT authorship contribution statement}
Yicheng Zou: Conceptualization, Methodology, Software, Data curation,
Formal analysis, Investigation, Visualization, Writing -- original draft.
Samuel Lanthaler: Methodology, Formal analysis, Supervision, Writing -- review \& editing.
Hossein Salahshoor: Conceptualization, Methodology, Formal analysis, Supervision, Funding acquisition,
Project administration, Writing -- review \& editing.

\section*{Declaration of competing interest}
The authors declare that they have no known competing financial interests or personal
relationships that could have appeared to influence the work reported in this paper.

\section*{Data availability}
Code and scripts to reproduce all figures and tables are available at:
\url{https://github.com/YichengZou626/Diffusion-Model-for-High-Frequency-Helmholtz}.
The datasets generated for this study, model training, and
post-processing pipelines are documented in the repository.

\appendix

\section{Method Details}\label{app:methods}
\subsection{Gaussian random fields}\label{app:grf}

We synthesize heterogeneous acoustic media by sampling stationary Gaussian random fields (GRFs) on the simulation grid and mapping them to sound speed. Specifically, we construct a mean-zero GRF in the Fourier domain by filtering complex white noise with an exponential spectral envelope and then applying an inverse real FFT (see Algorithm \ref{alg:grf2sound}). This yields families of coefficient fields with controllable smoothness and correlation length, providing ground-truth dataset across frequencies for the Helmholtz operator-learning benchmarks in the main text.

\begin{algorithm}[!ht]
\caption{Sampling heterogeneous sound-speed fields via spectral GRFs}
\label{alg:grf2sound}
\begin{algorithmic}[1]
\Require Grid size $(N_x,N_y)$; background speed $c_{\mathrm{bg}}$; scale $\sigma_c$; bounds $[c_{\min},c_{\max}]$
\State Sample hyperparameters: $\alpha \sim \mathcal{U}(0.5,\,2.5)$, $\;\ell \sim \mathcal{U}(0.35,\,0.7)$
\State Construct frequency grids $k_x=\mathrm{fftfreq}(N_x)$, $k_y=\mathrm{rfftfreq}(N_y)$; form mesh $K=\sqrt{k_x^2 + k_y^2}$
\State Define spectral envelope $\lambda \gets \exp\!\big(-(\ell\,K)^{\alpha}\big)$
\Repeat
  \State Draw complex white noise $\eta \gets \eta_{\mathrm{r}} + i\,\eta_{\mathrm{i}}$, with $\eta_{\mathrm{r}},\eta_{\mathrm{i}} \stackrel{\text{i.i.d.}}{\sim} \mathcal{N}(0,1)$
  \State Spectral field $\hat{u} \gets \lambda \odot \eta$
  \State Realization $u \gets \mathrm{irfft2}(\hat{u};\,N_x,N_y)$; \; mean–center: $u \gets u - \mathrm{mean}(u)$
  \State Map to sound speed: $c(x) \gets c_{\mathrm{bg}} + \sigma_c\,u(x)$
\Until{$c_{\min} < c(x) < c_{\max}$ for all grid points $x$}
\State \Return $c$
\end{algorithmic}
\end{algorithm}

\subsection{Data generation for 1D Helmholtz}\label{app:1d-solver}
We complement the 2D study with a matched 1D Helmholtz experiment that uses the same data–model pipeline but replaces J-Wave for a finite-difference frequency-domain (FDFD) Helmholtz solver. Given samples of the wave speed, which is from the same GRF family as in 2D, we assemble the tridiagonal system for $-u''+k(x)^2u=0$ with a left Dirichlet source and a right Sommerfeld (outgoing) boundary, and solve it in banded form (see Algorithm~\ref{alg:1d-fdfd}). We evaluate 5 frequencies $f\in\{1.5{\times}10^{5},\,2.5{\times}10^{5},\,5{\times}10^{5},\,7.5{\times}10^{5},\,1{\times}10^{6}\}\,\mathrm{Hz}$.

\begin{algorithm}[!ht]
\caption{1D FDFD Helmholtz solver}\label{alg:1d-fdfd}
\begin{algorithmic}[1]
\Require Wave speed samples $c[0{:}N\!-\!1]$ on $[0,L]$, frequency input (angular $\omega\gets 2\pi f$)
\Ensure Complex field $u[0{:}N\!-\!1]$ solving $-u''(x)+k(x)^2u(x)=0$, $k(x){=}\omega/c(x)$
\State $\Delta x \gets L/(N-1)$
\For{$j=0$ to $N-1$}
  \State $k[j] \gets \omega / c[j]$
\EndFor
\State Allocate $\texttt{ab}\in\mathbb{C}^{3\times N}$ and $b\in\mathbb{C}^{N}$; initialize to $0$
\State Interior stencil (implicit via diagonals):
\State $\quad$ Set $\texttt{ab}[0,j{-}1]\gets 1$,\; $\texttt{ab}[1,j]\gets -2 + (k[j]\Delta x)^2$,\; $\texttt{ab}[2,j{+}1]\gets 1$ for $j=1,\dots,N-2$
\State \textbf{Left BC (Dirichlet)} $u(0){=}\omega$:
\State $\quad \texttt{ab}[1,0]\gets 1$;\; $\texttt{ab}[0,1]\gets 0$;\; $b[0]\gets \omega$
\State \textbf{Right BC (Sommerfeld)} $u'(L)+i\,k(L)u(L){=}0$:
\State $\quad \texttt{ab}[1,N{-}1]\gets i\,k[N{-}1]\Delta x - 1$;\; $\texttt{ab}[2,N{-}1]\gets 1$;\; $b[N{-}1]\gets 0$
\State Solve $\texttt{ab}\cdot u = b$ with a tridiagonal banded solver (e.g., \texttt{solve\_banded((1,1), ab, b)})
\State \Return $u$
\end{algorithmic}
\end{algorithm}

\subsection{Denoiser architecture (conditional U\,-Net)}
\paragraph{Convolution (Conv1d/Conv2d)}
Let \(z\in\mathbb{R}^{H\times W\times C}\) be an input tensor and \(K\in\mathbb{R}^{k\times k\times C\times \widehat C}\) a kernel bank producing \(\widehat C\) output channels. For stride \(s\in\mathbb{N}\), the discrete multi\-channel convolution \(\mathcal{C}:\mathbb{R}^{H\times W\times C}\to\mathbb{R}^{\widehat H\times \widehat W\times \widehat C}\) is
\begin{equation}
\label{eq:conv}
\bigl(\mathcal{C}(z)\bigr)[i,j,\widehat\ell]
=\sum_{m=0}^{k-1}\sum_{n=0}^{k-1}\sum_{\ell=1}^{C}
K[m,n,\ell,\widehat\ell]\; z[i\,s+m,\ j\,s+n,\ \ell],
\quad
\begin{aligned}
&i=0,\dots,\widehat H{-}1,\\[-2pt]
&j=0,\dots,\widehat W{-}1,\\[-2pt]
&\widehat\ell=1,\dots,\widehat C.
\end{aligned}
\end{equation}
(To handle out\-of\-bounds indices, we use \emph{circular} padding to mitigate boundary artifacts in phase\-like Helmholtz fields.) The encoder downsamples with \(k{=}3\), stride \(s{=}2\) (no stride at the first scale), while the decoder mirrors this via nearest\-neighbor upsampling by a factor \(2\) followed by a \(k{=}3\) refinement convolution.

\paragraph{Activation functions (SiLU)}
We use the SiLU nonlinearity with \(\beta{=}1\). For a scalar \(z\),
\[
\mathrm{SiLU}(z)\;=\;z\,\sigma(z)\;=\;\frac{z}{1+e^{-z}},
\]
where \(\sigma(z)\) is the logistic sigmoid. SiLU is smooth and \emph{self-gating} (the input modulates its own pass-through), and unlike ReLU it preserves small negative activations.

\paragraph{Normalization}
To stabilize training with small batches typical in PDE settings, we normalize activations either by \emph{group normalization} (GN) or \emph{layer normalization} (LN).

\emph{Group Normalization (GN).}
Given \(z\in\mathbb{R}^{H\times W\times C}\), split the channel axis into \(G\) groups of size \(C/G\). Let \(\mathcal{G}_g\) be the index set of group \(g\) with \(m=HW\cdot(C/G)\). Per–group statistics and normalization are
\begin{equation}
\label{eq:gn-stats}
\mu_g=\frac{1}{m}\!\sum_{(h,w,c)\in\mathcal{G}_g}\!z_{h,w,c},\qquad
\sigma_g^2=\frac{1}{m}\!\sum_{(h,w,c)\in\mathcal{G}_g}\!(z_{h,w,c}-\mu_g)^2,
\end{equation}
\begin{equation}
\label{eq:gn}
\widehat z_{h,w,c}=\frac{z_{h,w,c}-\mu_g}{\sqrt{\sigma_g^2+\varepsilon}},\quad (h,w,c)\in\mathcal{G}_g,
\qquad
\widetilde z_{h,w,c}=\gamma_c\,\widehat z_{h,w,c}+\beta_c,
\end{equation}
with learnable per–channel scale \(\gamma_c\) and shift \(\beta_c\).

\emph{Layer Normalization (LN).}
LN normalizes \emph{all} features of a sample jointly (channel\(+\)space). Let \(\mathcal{I}=\{(h,w,c):1\!\le\!h\!\le\!H,1\!\le\!w\!\le\!W,1\!\le\!c\!\le\!C\}\) and \(M=HWC\). Then
\begin{equation}
\label{eq:ln-stats}
\mu=\frac{1}{M}\!\sum_{(h,w,c)\in\mathcal{I}}\!z_{h,w,c},\qquad
\sigma^2=\frac{1}{M}\!\sum_{(h,w,c)\in\mathcal{I}}\!(z_{h,w,c}-\mu)^2,
\end{equation}
\begin{equation}
\label{eq:ln}
\widehat z_{h,w,c}=\frac{z_{h,w,c}-\mu}{\sqrt{\sigma^2+\varepsilon}},
\qquad
\widetilde z_{h,w,c}=\gamma_c\,\widehat z_{h,w,c}+\beta_c.
\end{equation}

For simplicity, in our code we use LayerNorm by default.

\paragraph{Diffusion time embedding}
The denoiser conditions on diffusion time \(t\in[0,1]\) via a sinusoidal Fourier feature map followed by a small MLP:
\begin{equation}
\label{eq:time-feats}
\begin{aligned}
\phi(t)
&=
\Big[\cos(\omega_1 t),\dots,\cos(\omega_{64} t),\ \sin(\omega_1 t),\dots,\sin(\omega_{64} t)\Big]\in\mathbb{R}^{128}, \\
\omega_r
&=\tfrac{\pi}{2}\,\cdot\,10^{3\,\frac{r-1}{63}}.
\end{aligned}
\end{equation}

This produces multi–scale time features spanning several orders of magnitude. Another MLP then maps \(\phi(t)\) to the per–block conditioning width:
\begin{equation}
\label{eq:time-mlp}
e(t)\;=\;\mathrm{Linear}_{128\to 256}\bigl(\phi(t)\bigr)\xrightarrow{\mathrm{SiLU}}\mathrm{Linear}_{256\to d}\;\in\;\mathbb{R}^{d},
\end{equation}
after which a per–scale linear projection broadcasts \(e(t)\) over space and injects it additively into each \emph{context residual block} (FiLM–style conditioning) prior to the two \(k{=}3\) convolutions. This ensures that denoising decisions are time–aware at every resolution.

\paragraph{Positional encodings for coordinates}
In addition to time conditioning, we concatenate sinusoidal \emph{spatial} encodings to the input channels. In 2D,
\begin{multline}
\big\{\sin(2^\ell \pi x),\ \cos(2^\ell \pi x),\ \sin(2^\ell \pi y),\ \cos(2^\ell \pi y)\big\}_{\ell=0}^{L-1}.
\end{multline}
These multi--frequency positional cues help align phase and interference patterns across the domain,
particularly at high Helmholtz wavenumber \(k\).

\paragraph{U\,-Net layout} 
Here we specify implementation choices that were only summarized in the paper. The backbone is an encoder–decoder with skip connections and \emph{context residual blocks} that inject the diffusion time embedding \(e(t)\) at every depth. At scale \(s\) with channel width \(C_s\), a block applies FiLM–style additive conditioning and two same–width convolutions with normalization and SiLU:
\[
z \leftarrow z \;+\; \underbrace{\Big(\mathrm{Conv}_{3}\!\big(\sigma(\mathrm{Norm}(z+\Pi_s e(t)))\big)\xrightarrow{}\mathrm{Conv}_{3}\Big)}_{\text{residual branch}},
\]
where \(\Pi_s:\mathbb{R}^{d}\!\to\!\mathbb{R}^{C_s}\) is a linear projection of \(e(t)\) followed by unflattening broadcast over space, \(\mathrm{Conv}_{3}\) denotes a kernel-\(3\) circularly padded convolution (Conv1d/2d depending on dimensionality), \(\mathrm{Norm}\) is LayerNorm by default, and \(\sigma\) is SiLU. Downsampling “heads’’ are strided convolutions (kernel \(3\), stride \(2\) at all but the first scale); upsampling “tails’’ perform \(\mathrm{Norm}\!\to\) nearest–neighbor upsample \(\times 2\!\to\) kernel-\(3\) convolution and are added to the aligned encoder feature (\emph{skip connection}). Circular padding is used throughout to reduce boundary artifacts for Helmholtz phase fields.

\paragraph{Training}
We train the conditional U\,-Net as a score network within a variance–preserving SDE (VPSDE) using mini-batch denoising score matching on tensors \(X\in\mathbb{R}^{B\times C\times H\times W}\) with \(H\!=\!W\!=\!256\) and a small batch size \(B=32\); the channel stack is a \emph{single} solution channel that is noised, and \emph{three} clean conditioning channels (e.g., sound speed, source mask, positional encodings). At each iteration we sample \(t\sim\mathrm{Unif}(0,1)\), obtain \(u_t\) via the VPSDE perturbation kernel implemented in our forward function, and minimize a mean-squared denoising objective,
\begin{equation}
\label{eq:vpsde-loss-app}
\mathcal{L}_{\text{VPSDE}}(\theta)
=\mathbb{E}_{u_0,\,z,\,t}\Big[w(t)\,\big\|\,s_\theta\!\big(u_t, z, t\big)-\tilde{y}(u_t,t)\big\|_2^2\Big],
\end{equation}
where \(s_\theta\) is the U\,-Net denoiser, \(\tilde{y}(u_t,t)\) is the VPSDE target produced by forward diffusion, and \(w(t)\equiv 1\) in our runs; in code this appears as \(\texttt{mse}=(\texttt{net}(u_t,z,t)-\texttt{output})^2\) averaged over spatial/channel dimensions, followed by \(\texttt{mean}()\) over the batch. Each epoch iterates over mini-batches on the selected GPU device, updates an EMA of parameters for more stable sampling with a standard learning-rate scheduler, and evaluates the same loss on a held-out validation split without gradients. At inference, we use different sampling methods (see more details in Appendix~\ref{app:sample}).

\subsection{Fourier Neural Operator (FNO)}
We implement a 2D \emph{tensorized} Fourier neural operator as a strong deterministic baseline. Let \(z\in\mathcal{Z}\subset\mathbb{R}^{H\times W\times d_u}\) denote the input stack of channels (here \(d_z{=}3\): sound speed, source mask, positional encodings) and \(G_\phi:\mathcal{Z}\to\mathcal{Y}\) the learned operator producing the target wavefield. The FNO is composed as
\[
G_\phi \;=\; Q \circ L_L \circ \cdots \circ L_1 \circ R,
\]
with a \emph{lifting} \(R:\mathbb{R}^{d_u}\to\mathbb{R}^{d_v}\) (pointwise \(1{\times}1\) conv) that increases channel width to \(d_v\), followed by \(L\) Fourier layers \(L_\ell\), and a \emph{projection} \(Q:\mathbb{R}^{d_v}\to\mathbb{R}^{d_y}\) (pointwise head) back to output channels. Each Fourier layer is
\begin{equation}
\label{eq:fno-layer}
v_{\ell+1}(z)
\;=\;
\sigma\!\Big(
W_\ell v_\ell(z) \;+\; \mathcal{F}^{-1}\!\big[\,P_\ell(k)\, \cdot\, \mathcal{F}[v_\ell](k)\,\big](z)
\Big),
\end{equation}
where \(v_\ell\in\mathbb{R}^{H\times W\times d_v}\), \(W_\ell\) is a learned \emph{local} (linear/skip) operator, \(\mathcal{F}\) is the 2D FFT, and \(P_\ell(k)\in\mathbb{C}^{d_v\times d_v}\) are learned complex multipliers applied only on a truncated band of modes. Here we retain \((n_{\text{modes}}^{h},n_{\text{modes}}^{w})=(64,64)\) in every Fourier layer and use GeLU nonlinearity (\texttt{preactivation}=0); the spectral update is the factorized/tensorized variant that parameterizes \(P_{\ell}(k)\) with low-rank factors (\texttt{rank}=1.0), reducing memory and improving stability at \(256\times256\). After each spectral block, a lightweight channel MLP (\texttt{use\_channel\_mlp}=1, expansion=\(0.5\), dropout=\(0\)) mixes features in the spatial domain. We employ group normalization and linear residual skips, while domain padding is off by default. The architecture uses \(L=4\) Fourier layers with hidden width \(d_v=\)32 and projection ratio \(2\times\) inside the spectral block, taking \texttt{data\_channels}=3 as input and producing a single output channel per component. Optimization follows standard operator-learning practice: \(1000\) epochs with AdamW (lr \(=5\times10^{-3}\), weight decay \(=10^{-4}\)), batch size \(32\), \texttt{StepLR} scheduler (step size=\(60\), \(\gamma=0.5\)), and an \(H^{1}\) training loss to encourage gradient fidelity. Overall, this \texttt{FNO} baseline provides a fair deterministic comparator focused on frequency-domain accuracy under controlled memory/compute.

\subsection{Helmholtz Neural Operator (HNO)}
Our HNO baseline follows a UNO–style spectral operator with U–shaped skip connections tailored to \(256{\times}256\) grids. Inputs are tensors \(z\in\mathbb{R}^{B\times H\times W\times C}\) with \(H{=}W{=}256\) and \(C{=}3\) channels (sound speed, source mask, positional features); outputs are \(u\in\mathbb{R}^{B\times H\times W\times 1}\). Then, we concatenate a coordinate grid (linear \([0,1]\)) to the input, lift with two pointwise MLPs (\texttt{fc\_n1}, \texttt{fc0}), permute to \((B,C,H,W)\), and pass through eight \emph{operator blocks} with encoder–decoder topology and skip concatenations. Each block is
\[
\begin{aligned}
\text{OperatorBlock2D:}\quad
&\underbrace{\text{SpectralConv2d\_Uno}}_{\text{FFT }\to\text{ low-rank spectral map }\to\text{ iFFT}}
\;+\;
\underbrace{\text{pointwise\_op\_2D}}_{\text{1{\small x}1 conv + bicubic resize}}
\\[2pt]
&\xrightarrow{\ \mathrm{GELU}\ }\,
\mathrm{MLP(1{\small x}1)}\;
\xrightarrow{\ \mathrm{LayerNorm}\ }\,
\text{residual add}\;
\xrightarrow{\ \mathrm{GELU}\ }\, .
\end{aligned}
\]

The spectral layer performs \(z\mapsto\mathcal{F}^{-1}\!\big(P(\mathbf{k})\odot\mathcal{F}z\big)\) with two learnable complex weight tensors (\texttt{weights1}, \texttt{weights2}) applied to the positive and negative vertical bands of retained modes. In parallel, a pointwise 1{\small x}1 convolution is up-/down–scaled via bicubic interpolation to the block’s output resolution and added to the spectral path, improving locality and stabilizing high–\(k\) content. Each block ends with an MLP (two 1{\small x}1 convs with GeLU) and \texttt{LayerNorm} over \([\text{H},\text{W}]\), followed by a residual addition with the pointwise branch and GeLU. The U–shaped pathway uses resolutions \((256,256)\!\to\!(128,128)\!\to\!(64,64)\!\to\!(32,32)\) and back, with mode budgets matched per scale: \texttt{conv1}: \((\text{dim}=256,256;\ \text{modes}=192,96)\), \texttt{conv2}: \((128,128;\ 128,64)\), \texttt{conv3}: \((64,64;\ 64,32)\), \texttt{conv4}: \((32,32;\ 32,16)\), then symmetric values on the way up. After the decoder, features are concatenated with the lifted input, projected by a kernel MLP to output a per–pixel scalar kernel, and contracted with features via a normalized Einstein sum, yielding a per–pixel reduction. A final two–layer head with \(\tanh\) maps to outputs.

During the training, we instantiate \texttt{UNO2D} with \texttt{width}=16, \texttt{in\_channels}=3, \texttt{out\_channels}=1 and train on \(256{\times}256\) fields using Adam (lr \(10^{-3}\), weight decay \(10^{-5}\)), \texttt{StepLR} (step=30, \(\gamma{=}0.5\)), batch size \(32\), for up to \(1000\) epochs. The loss combines \(0.9\,L^1+0.1\,L^2\) on the real component trained per pass. This HNO baseline thus realizes a frequency–domain, multi–scale spectral operator that couples learned Fourier multipliers with local pointwise updates and U–shaped skips, providing a strong deterministic comparator for our probabilistic diffusion operator on Helmholtz problems.

\section{Sampler ablation}\label{app:sampler}
Building on Sec.~5.1, which motivates DDPM as our default choice, we now present the three sampling procedures considered in the sampler ablation. Let $\{\beta_t\}_{t=1}^{T}$ be a discrete noise schedule (linear or cosine), $\alpha_t:=1-\beta_t$, and $\bar\alpha_t:=\prod_{s=1}^{t}\alpha_s$. The forward process is $q(z_t\!\mid\!z_0)=\mathcal{N}\!\big(\sqrt{\bar\alpha_t}\,z_0,(1-\bar\alpha_t)I\big)$ and the network predicts noise $\varepsilon_\theta(z_t,t,c)$ given the condition $c$. DDPM (ancestral): We use the standard Gaussian reverse transition
\begin{equation}
\begin{aligned}
p_\theta(z_{t-1}\mid z_t,c)
  &= \mathcal{N}\!\Big(z_{t-1};\,\mu_\theta(z_t,t,c),\,\tilde\beta_t I\Big),\\
\mu_\theta(z_t,t,c)
  &= \frac{1}{\sqrt{\alpha_t}}\!\left(z_t-\frac{\beta_t}{\sqrt{1-\bar\alpha_t}}\,
     \varepsilon_\theta(z_t,t,c)\right).
\end{aligned}
\end{equation}
with posterior variance $\tilde\beta_t=\frac{1-\bar\alpha_{t-1}}{1-\bar\alpha_t}\,\beta_t$. We ablate \emph{linear} and \emph{cosine} schedules and step budgets $T\in\{10,50,100,1000\}$. Over $T$ steps, the update draws $z\!\sim\!\mathcal{N}(0,I)$ at each step ($t>1$) and sets $z_{t-1}=\mu_\theta+\sigma_t z$ with $\sigma_t=\sqrt{\tilde\beta_t}$. DDIM (implicit): Using the same schedule, define the predicted clean sample $\hat z_0(z_t,t,c)=\frac{1}{\sqrt{\bar\alpha_t}}\big(z_t-\sqrt{1-\bar\alpha_t}\,\varepsilon_\theta(z_t,t,c)\big)$. For a decreasing sequence $t_1{>}t_2{>}\cdots{>}t_S$, DDIM updates deterministically ($\eta{=}0$) as
\begin{equation}
z_{t_{i+1}}\;=\;\sqrt{\bar\alpha_{t_{i+1}}}\,\hat z_0\;+\;\sqrt{1-\bar\alpha_{t_{i+1}}}\,\varepsilon_\theta(z_{t_i},t_i,c),
\end{equation}
or stochastically with $\eta\!\in\![0,1]$ by decomposing the second term into an $\eta z$ component plus a $(1-\eta^2)^{1/2}\varepsilon_\theta$ component. In our ablation we use the \emph{cosine} schedule and report the $\eta{=}0$ (deterministic) case. Score-based SDE (continuous-time): In the variance-preserving formulation $dz=\!-{\tfrac{1}{2}}\beta(t)\,z\,dt+\sqrt{\beta(t)}\,dW_t$, the closed-form mean/variance maps from $t$ to $t-\Delta t$ can be written as $z_{t-\Delta t}=r(t,\Delta t)\,z_t+\big(\sigma(t\!-\!\Delta t)-r(t,\Delta t)\sigma(t)\big)\,\varepsilon_\theta(z_t,t,c)$, where $r(t,\Delta t)=\mu(t-\Delta t)/\mu(t)$ and $(\mu,\sigma)$ are the analytic solution scalings of the VP–SDE\footnote{In code, \texttt{mu(\,)} and \texttt{sigma(\,)} implement these scalings, and we integrate $t$ linearly from $1\!\to\!0$ with step $1/T$.}. This implements an ancestral sampler in continuous time, and we use the \emph{cosine} schedule for $\beta(t)$ via its discretized counterpart and the same step budgets as above. Across methods, DDPM introduces stepwise Gaussian noise (higher diversity, slightly higher variance at small $T$), DDIM provides a deterministic path given the same $\varepsilon_\theta$ (lower variance and faster convergence at small $T$), and the SDE sampler follows the continuous-time ancestral update governed by $(\mu,\sigma)$. Our ablation (Table~\ref{tab:sampler_ablation}) compares these choices under identical networks and conditions.

\begin{table}[t]
\centering
\scriptsize
\setlength{\tabcolsep}{3.5pt}
\caption{\textbf{Sampler–step ablation at $f=2.5\times10^6$ Hz} (relative error; lower is better). 
Cosine schedule unless noted. Values rounded to 3 decimals; best per metric in \textbf{bold}.}
\label{tab:sampler_ablation}
\begin{tabular}{lcccccccccccc}
\toprule
& \multicolumn{4}{c}{$\mathcal{E}_{L^2}\!\downarrow$} & \multicolumn{4}{c}{$\mathcal{E}_{H^1}\!\downarrow$} & \multicolumn{4}{c}{$\mathcal{E}_{\text{energy}}\!\downarrow$} \\
\cmidrule(lr){2-5}\cmidrule(lr){6-9}\cmidrule(lr){10-13}
\textbf{Sampler} & \textbf{10} & \textbf{50} & \textbf{100} & \textbf{1000} & \textbf{10} & \textbf{50} & \textbf{100} & \textbf{1000} & \textbf{10} & \textbf{50} & \textbf{100} & \textbf{1000} \\
\midrule
DDPM (linear) & 0.999 & 0.956 & 0.942 & 0.918 & 0.999 & 0.797 & 0.720 & 0.631 & 0.999 & 0.949 & 0.932 & 0.901 \\
DDPM (cosine) & 0.457 & 0.268 & 0.234 & \textbf{0.188} & 0.180 & 0.106 & 0.095 & \textbf{0.078} & 0.138 & 0.100 & 0.093 & \textbf{0.082} \\
DDIM (cosine) & 0.470 & 0.333 & 0.305 & 0.284 & 0.189 & 0.139 & 0.129 & 0.122 & 0.137 & 0.122 & 0.124 & 0.125 \\
SDE (cosine)  & 0.458 & 0.373 & 0.358 & 0.336 & 0.190 & 0.151 & 0.146 & 0.138 & 0.168 & 0.126 & 0.124 & 0.129 \\
\bottomrule
\end{tabular}
\end{table}

\begin{table}[H]
\centering
\scriptsize
\setlength{\tabcolsep}{3.5pt}
\caption{\textbf{Sampler--step ablation at $f=2.5\times10^6$ Hz }(wall-clock time per 500 samples in seconds).} 
\label{tab:sampler_ablation_time}
\begin{tabular}{lcccc}
\toprule
& \multicolumn{4}{c}{Time (s)} \\
\cmidrule(lr){2-5}
\textbf{Sampler} & \textbf{10} & \textbf{50} & \textbf{100} & \textbf{1000} \\
\midrule
DDPM (linear) & 8.927  & 46.892 & 94.534 & 948.698 \\
DDPM (cosine) & 19.541 & 57.521 & 95.430 & 951.771 \\
DDIM (cosine) & 10.041 & 47.731 & 95.023 & 948.228 \\
SDE (cosine)  & 9.752  & 47.713 & 94.918 & 947.771 \\
\bottomrule
\end{tabular}
\end{table}

In addition to accuracy, we also report the wall-clock time required to generate a single sample for each sampler and step budget in Tab.~\ref{tab:sampler_ablation_time}. Even though our current sampling procedure is computationally expensive, recent work has proposed substantially faster alternatives—such as ODE-based samplers and exponential-integrator (EI) schemes for diffusion models (e.g., the exponential integrator method of~\cite{zhang2022fast})—which we plan to explore in future work. For all remaining experiments, we therefore adopt a cosine-schedule DDPM sampler with 1000 steps as our default, as it offers the best accuracy in the high-frequency Helmholtz regime.

\begin{table}[H]
\centering
\scriptsize
\setlength{\tabcolsep}{4pt}
\caption{\textbf{Model size, evaluation cost, and accuracy at $f=2.5\times10^6$ Hz.} 
Parameter counts are total learnable parameters; times are wall-clock evaluation times for a single sample;
errors are relative. The direct J-Wave solver is treated as the reference with zero relative error.}
\label{tab:model_params_time}
\begin{tabular}{lrrrrr}
\toprule
\textbf{Model} 
& \textbf{\# Parameters} 
& \textbf{Eval. time (s)} 
& $\mathcal{E}_{L^2}$ 
& $\mathcal{E}_{H^1}$ 
& $\mathcal{E}_{\text{energy}}$ \\
\midrule
U-Net                 & 91.9M  & 0.01  & 0.767 & 1.000 & 0.514 \\
FNO                   & 15.2M  & 0.07  & 0.412 & 0.494 & 0.141 \\
HNO                   & 73.1M  & 0.15  & 0.802 & 0.996 & 0.590 \\
Diffusion (DDPM/1000) & 92.6M  & 1.83  & \textbf{0.095} & \textbf{0.135} & \textbf{0.036} \\
Direct solver (J-Wave) & ---   & 23.84 & --- & --- & --- \\
\bottomrule
\end{tabular}
\end{table}

To contextualize this choice, Tab.~\ref{tab:model_params_time} compares parameter counts and wall-clock time per sample across the baseline models and our chosen diffusion configuration.
Even with 1000 denoising steps, the diffusion model remains within a practical runtime range relative to the deterministic neural operators, and still operates orders of magnitude faster than the numerical Helmholtz solver, which requires several minutes of computation per sample on the same hardware.

\section{Additional Results}\label{app:add-figures}
This section provides extended qualitative and quantitative results that complement the main text.

\subsection{Data example}
\begin{figure}[!ht]
  \centering
  \setlength{\tabcolsep}{2pt}

  \begin{subfigure}[t]{0.32\linewidth}
    \centering
    \caption*{\small $f=1.5\times10^{5}$ Hz}
    \includegraphics[width=\linewidth]{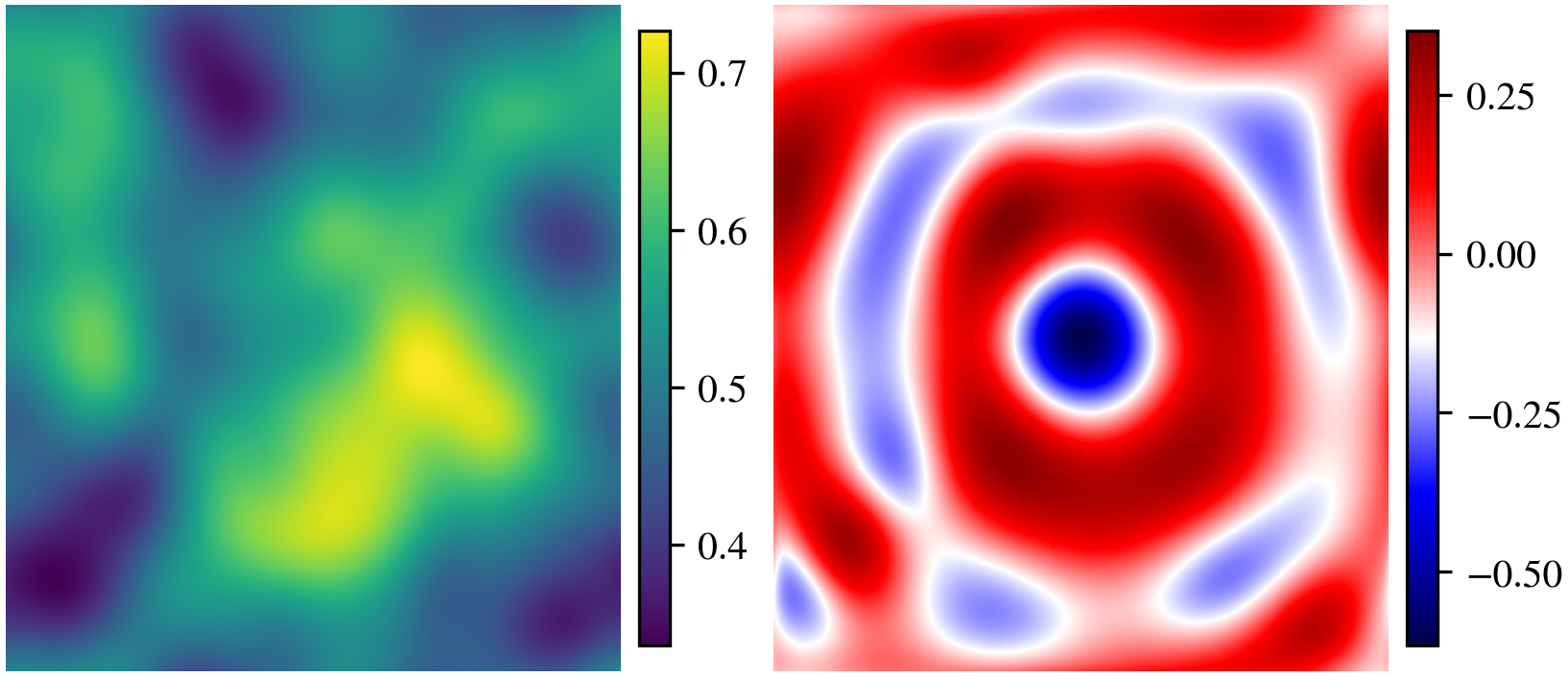}
  \end{subfigure}\hfill
  \begin{subfigure}[t]{0.32\linewidth}
    \centering
    \caption*{\small $f=2.5\times10^{5}$ Hz}
    \includegraphics[width=\linewidth]{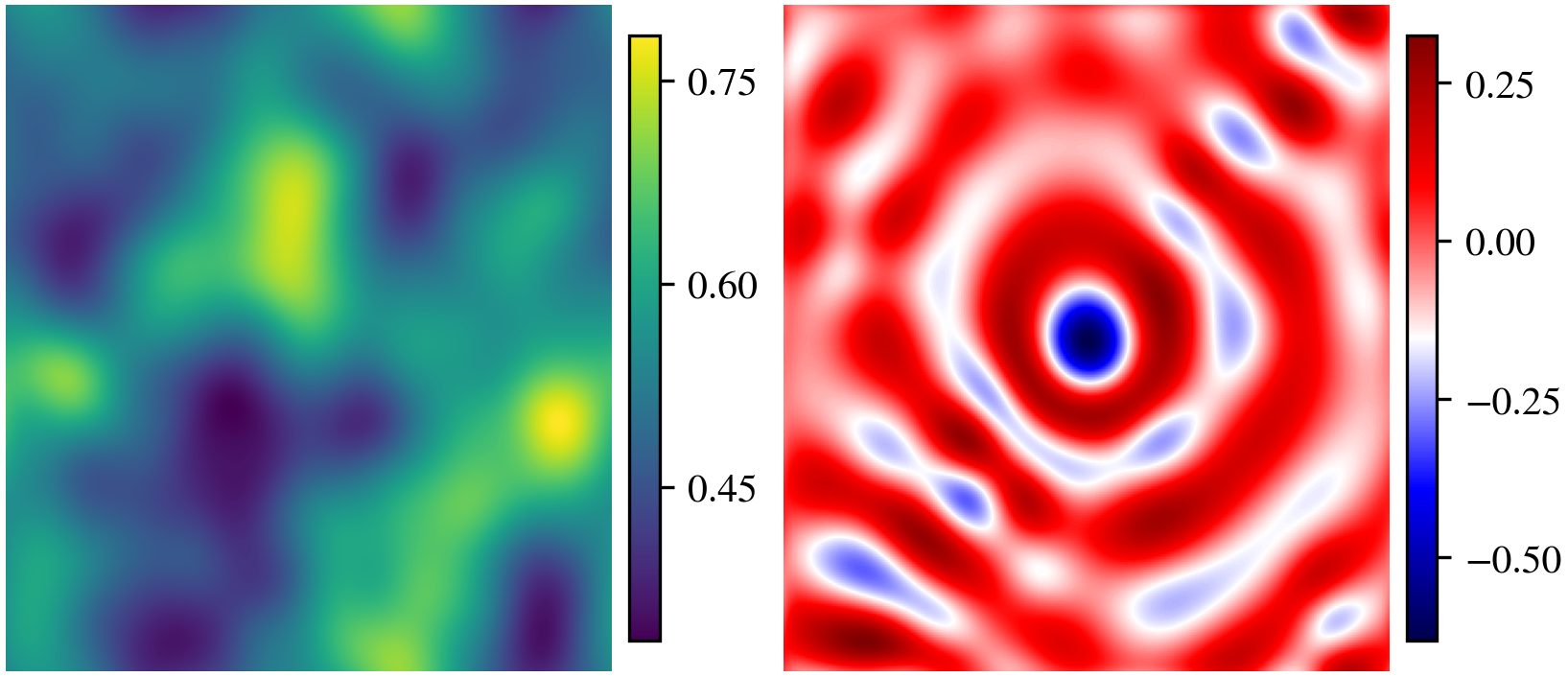}
  \end{subfigure}\hfill
  \begin{subfigure}[t]{0.32\linewidth}
    \centering
    \caption*{\small $f=5\times10^{5}$ Hz}
    \includegraphics[width=\linewidth]{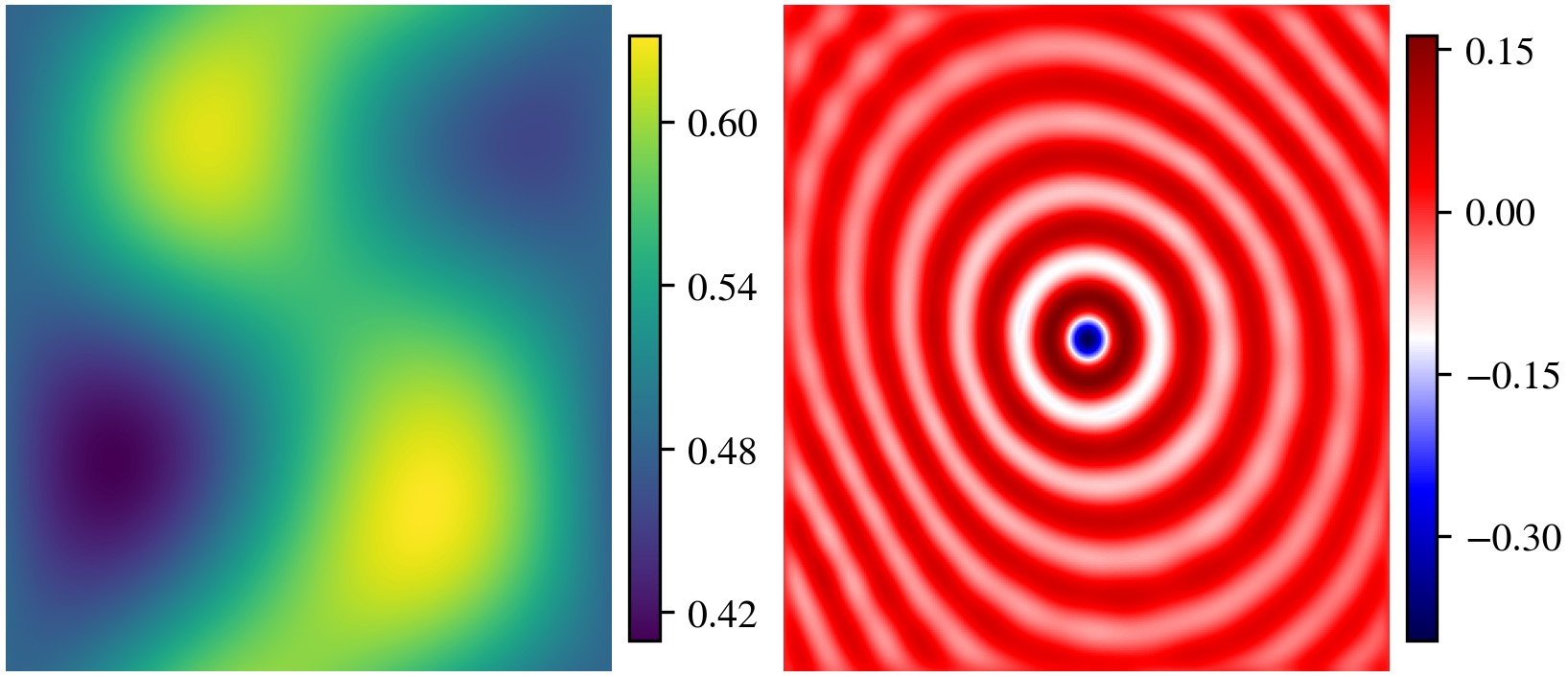}
  \end{subfigure}

  \vspace{0.6em}

  \begin{subfigure}[t]{0.32\linewidth}
    \centering
    \caption*{\small $f=1\times10^{6}$ Hz}
    \includegraphics[width=\linewidth]{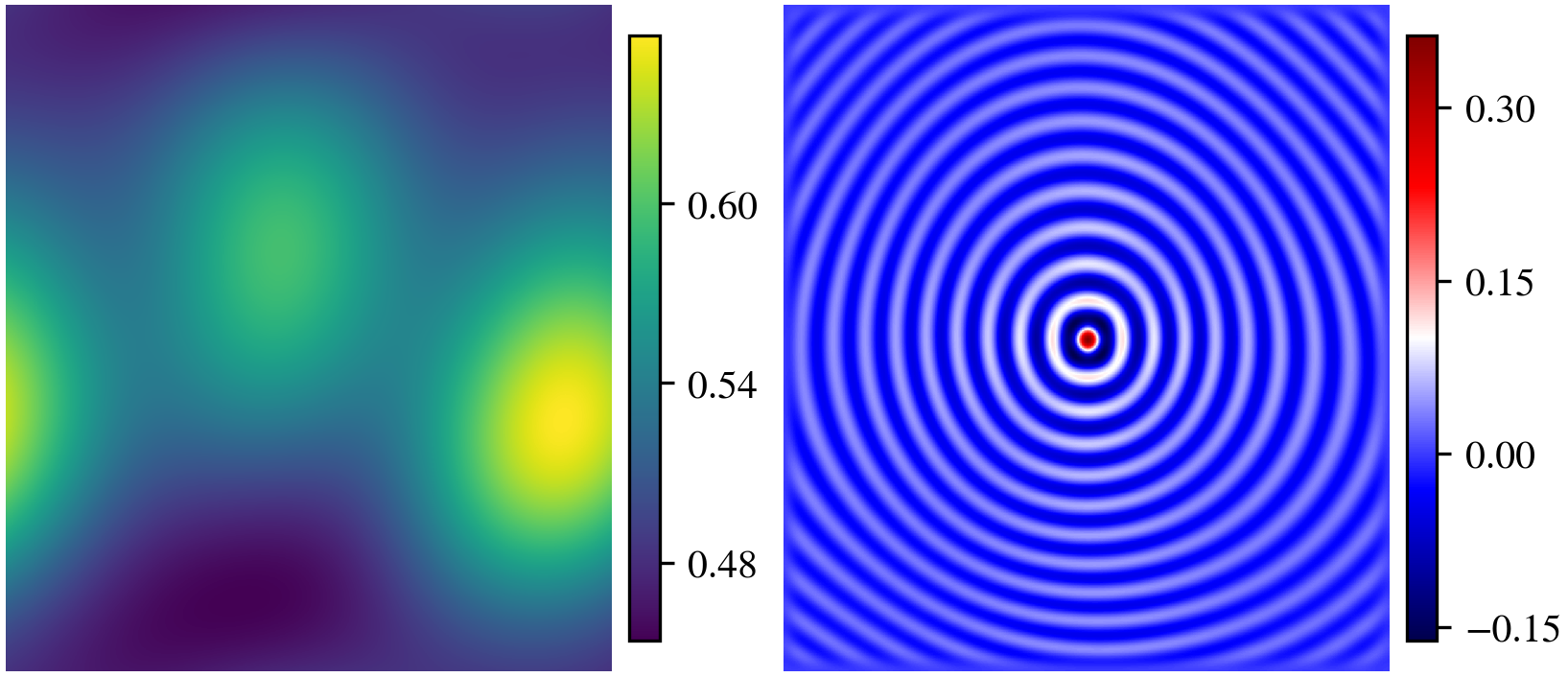}
  \end{subfigure}\hfill
  \begin{subfigure}[t]{0.32\linewidth}
    \centering
    \caption*{\small $f=1.5\times10^{6}$ Hz}
    \includegraphics[width=\linewidth]{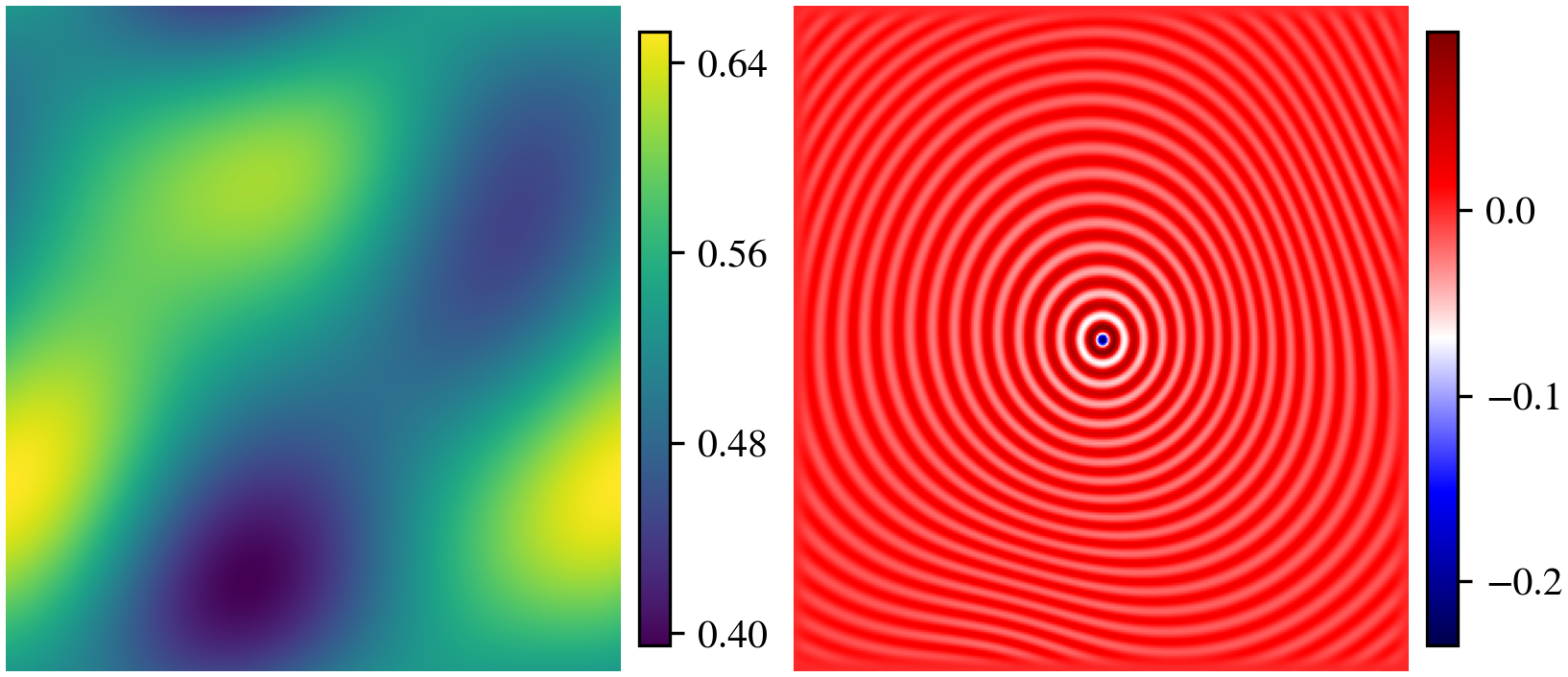}
  \end{subfigure}\hfill
  \begin{subfigure}[t]{0.32\linewidth}
    \centering
    \caption*{\small $f=2.5\times10^{6}$ Hz}
    \includegraphics[width=\linewidth]{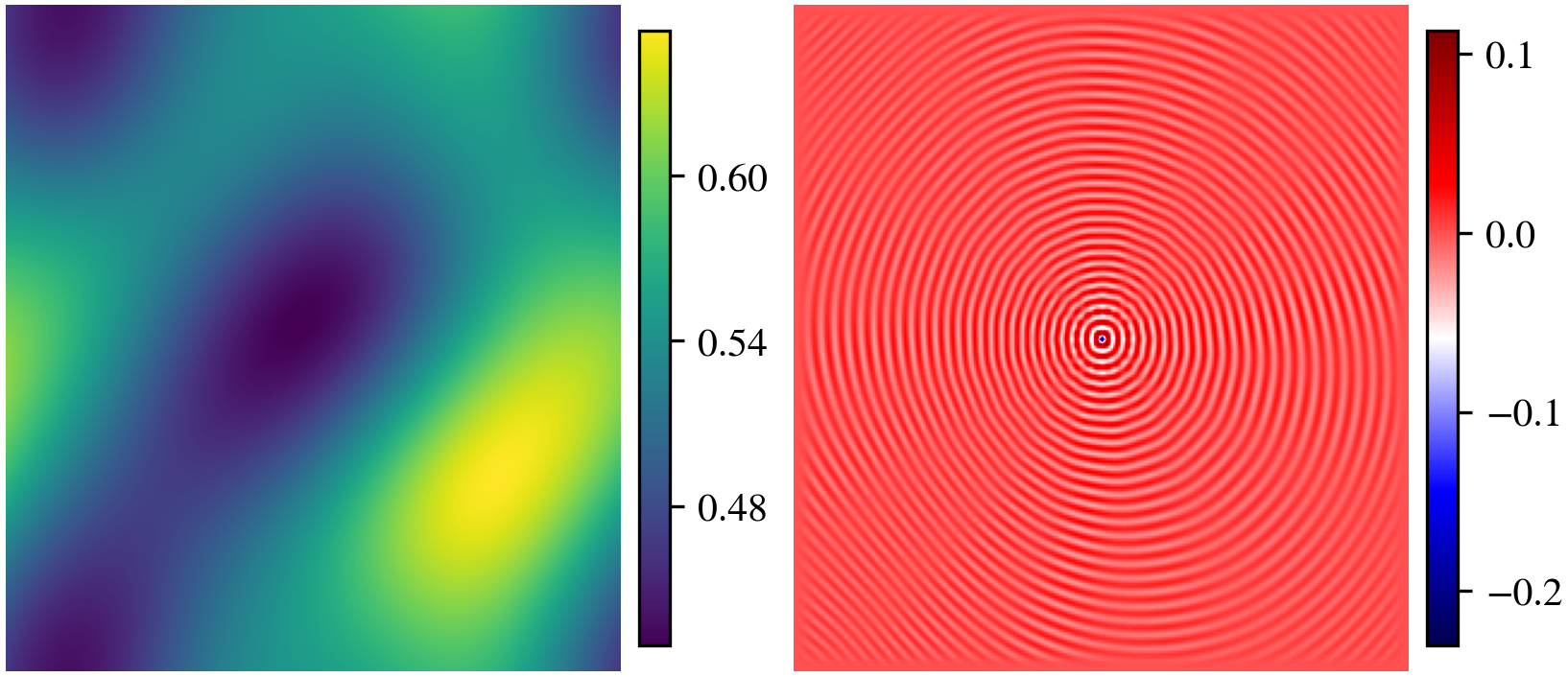}
  \end{subfigure}

  \caption{\textbf{Input–output data pairs by frequency.}
  Each tile shows one random test example: left—coefficient field $c$; right—solution $P$. Color scales are set per panel.}
  \label{fig:data-example}
\end{figure}

Figure~\ref{fig:data-example} shows representative input--output pairs across frequencies: as the driving rate increases, the Helmholtz solution exhibits denser interference fringes and faster phase oscillations. Using angular frequency $\omega=2\pi f$ and local sound speed $c(x)$, the wavelength is
\[
\lambda(x)\;=\;\frac{2\pi\,c(x)}{\omega}.
\]
With an average speed $\bar c\!\approx\!2000$ (simulation units) and our largest angular frequency $\omega_{\max}=2.5\times10^{6}\ \mathrm{rad/s}$, the shortest wavelength is
\[
\lambda_{\min}\;=\;\frac{2\pi\,\bar c}{\omega_{\max}}
\;=\;\frac{2\pi\times 2000}{2.5\times10^{6}}
\;\approx\;5\times10^{-3}.
\]
We use a uniform grid with $\Delta x=\Delta y=10^{-3}$, yielding $\lambda_{\min}/\Delta x\approx 5$ samples per shortest wavelength, which is adequate to resolve the fine oscillatory structure in the ground-truth wavefields.

\subsection{More frequency results}\label{app:qual}
\begin{figure}[t]
  \centering
  \captionsetup{font=small,skip=6pt}
  \captionsetup[subfigure]{labelformat=empty,justification=centering,aboveskip=2pt,belowskip=2pt}

  \setlength{\rowH}{0.20\textheight}

  \begin{subfigure}[t]{0.98\textwidth}
    \centering    \includegraphics[width=\linewidth,height=\rowH,keepaspectratio]{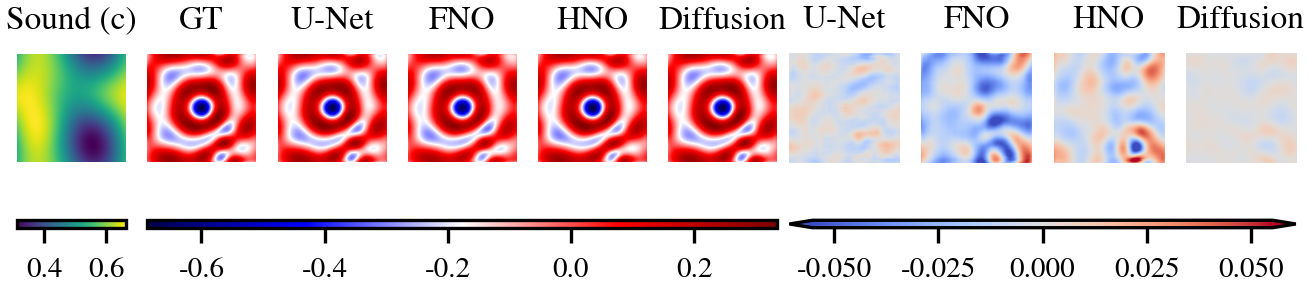}
  \end{subfigure}
  \vspace{4pt}
  
  \begin{subfigure}[t]{0.98\textwidth}
    \centering    \includegraphics[width=\linewidth,height=\rowH,keepaspectratio]{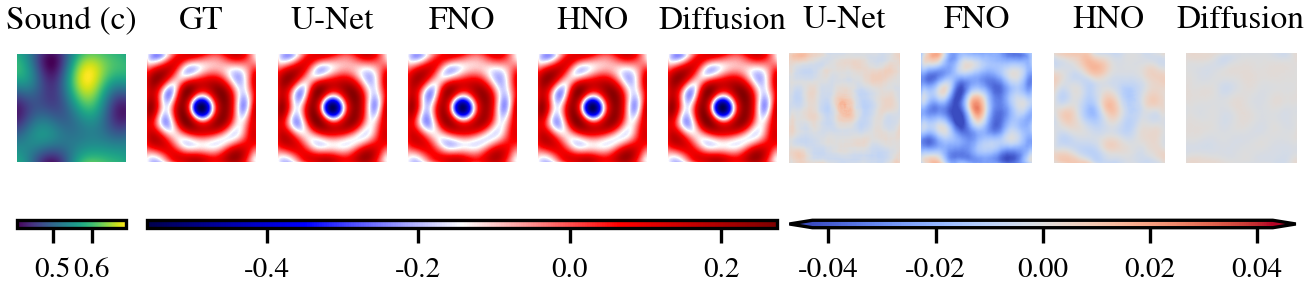}
  \end{subfigure}
  \vspace{4pt}

  \begin{subfigure}[t]{0.98\textwidth}
    \centering    \includegraphics[width=\linewidth,height=\rowH,keepaspectratio]{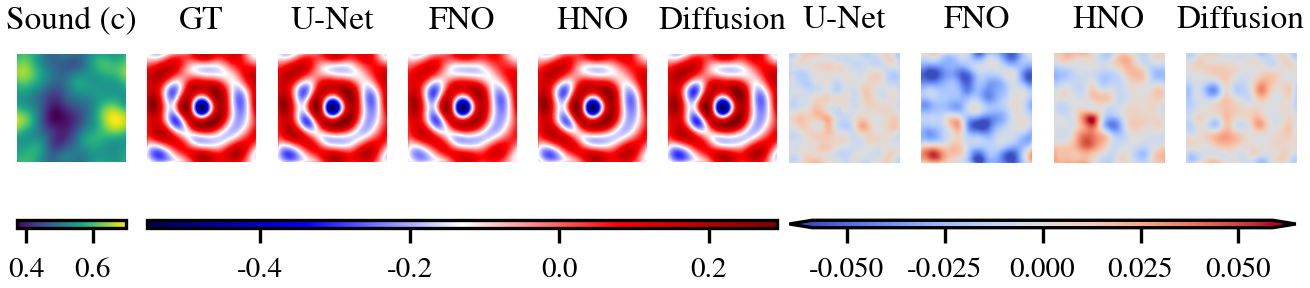}
  \end{subfigure}
  \vspace{4pt}

  \begin{subfigure}[t]{0.98\textwidth}
    \centering    \includegraphics[width=\linewidth,height=\rowH,keepaspectratio]{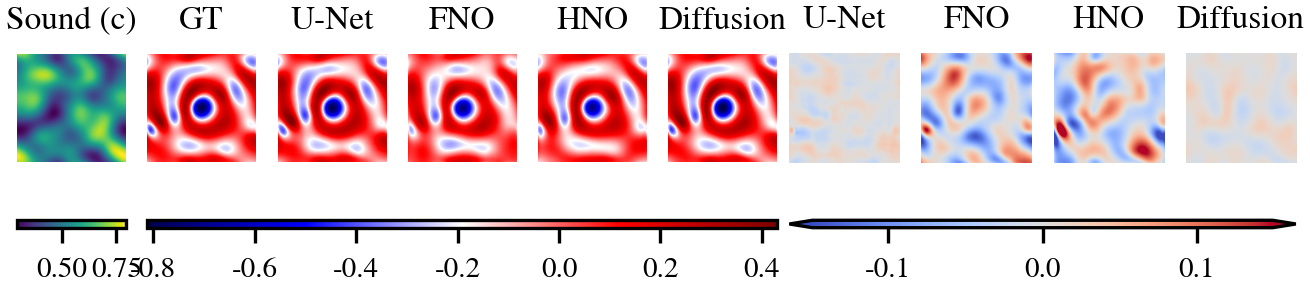}
  \end{subfigure}
  \vspace{4pt}

  \begin{subfigure}[t]{0.98\textwidth}
    \centering    \includegraphics[width=\linewidth,height=\rowH,keepaspectratio]{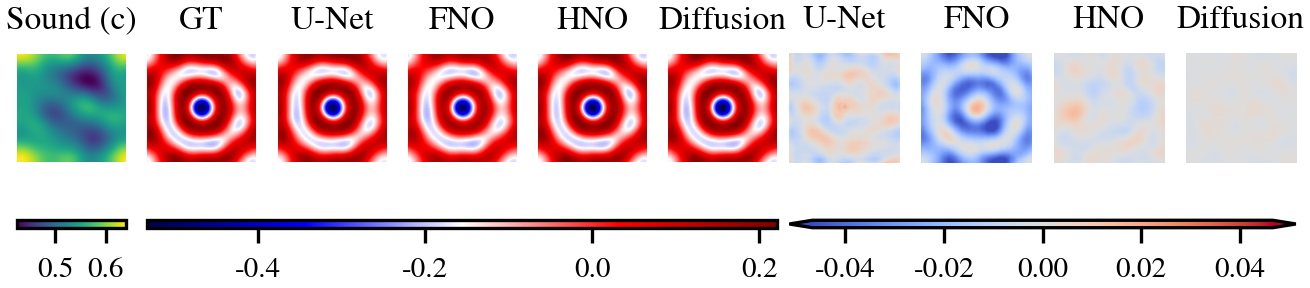}
  \end{subfigure}
  \vspace{4pt}

  \begin{subfigure}[t]{0.98\textwidth}
    \centering    \includegraphics[width=\linewidth,height=\rowH,keepaspectratio]{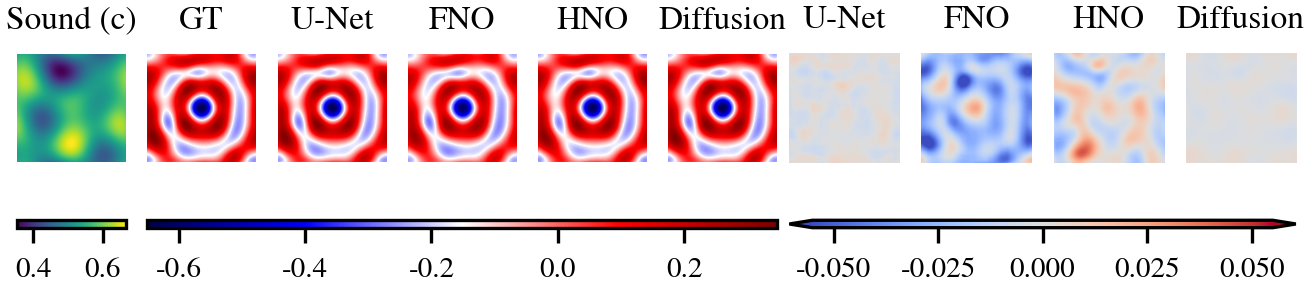}
  \end{subfigure}
  \vspace{4pt}
  
  \begin{subfigure}[t]{0.98\textwidth}
    \centering    \includegraphics[width=\linewidth,height=\rowH,keepaspectratio]{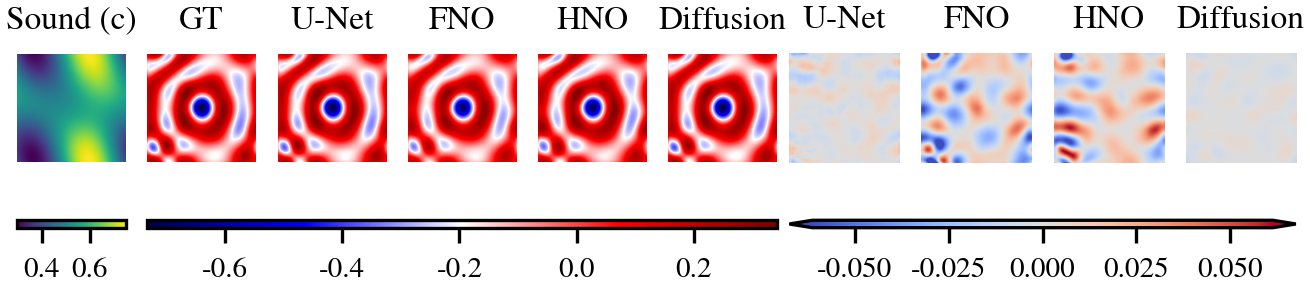}
    \vspace{2pt}
    \subcaption*{\small%
    \makebox[0pt][l]{\hspace*{0.22\linewidth}\emph{Prediction vs.\ Ground Truth}}%
    \makebox[\linewidth][l]{\hspace*{0.70\linewidth}\emph{Residual (Pred$-$GT)}}%
    }
    \vspace{-5pt}
  \end{subfigure}

  \caption{\textbf{Qualitative comparisons at $f=1.5\times10^{5}$ Hz.}
  }
  \label{fig:qual-supp-f1}
\end{figure}

\begin{figure}[t]
  \centering
  \captionsetup{font=small,skip=6pt}
  \captionsetup[subfigure]{labelformat=empty,justification=centering,aboveskip=2pt,belowskip=2pt}

  \setlength{\rowH}{0.20\textheight}

  \begin{subfigure}[t]{0.98\textwidth}
    \centering    \includegraphics[width=\linewidth,height=\rowH,keepaspectratio]{figs/additional_figures/qual/example_2d_2.5e5_idx11_with_sound.png}
  \end{subfigure}
  \vspace{4pt}
  
  \begin{subfigure}[t]{0.98\textwidth}
    \centering    \includegraphics[width=\linewidth,height=\rowH,keepaspectratio]{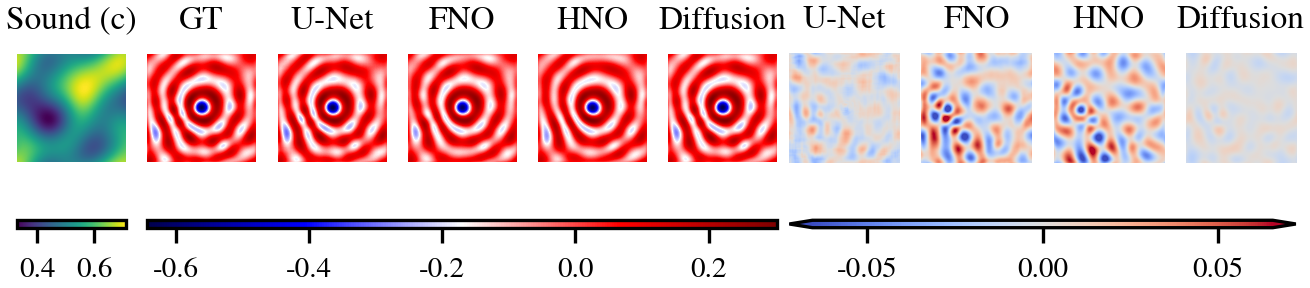}
  \end{subfigure}
  \vspace{4pt}

  \begin{subfigure}[t]{0.98\textwidth}
    \centering    \includegraphics[width=\linewidth,height=\rowH,keepaspectratio]{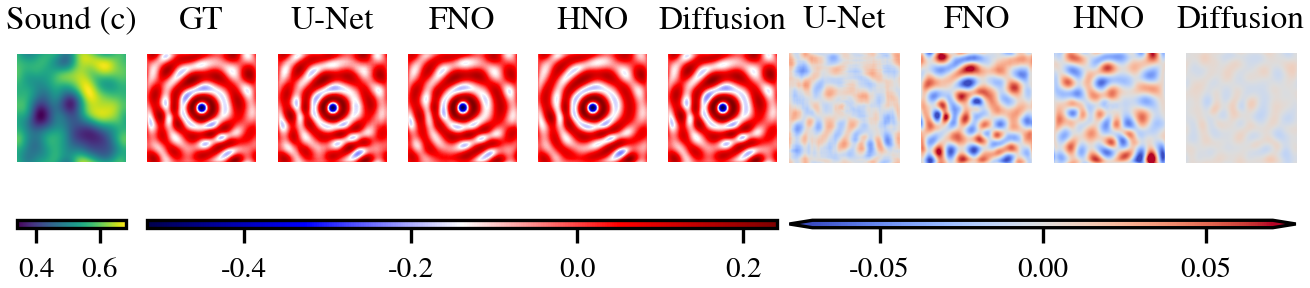}
  \end{subfigure}
  \vspace{4pt}

  \begin{subfigure}[t]{0.98\textwidth}
    \centering    \includegraphics[width=\linewidth,height=\rowH,keepaspectratio]{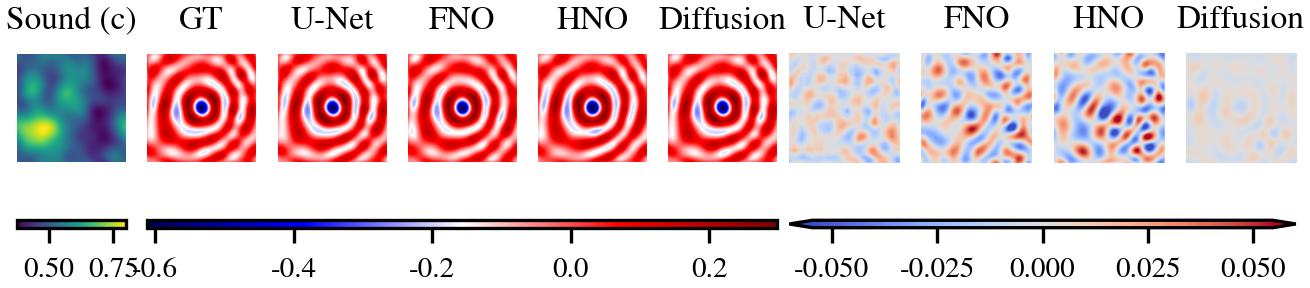}
  \end{subfigure}
  \vspace{4pt}

  \begin{subfigure}[t]{0.98\textwidth}
    \centering    \includegraphics[width=\linewidth,height=\rowH,keepaspectratio]{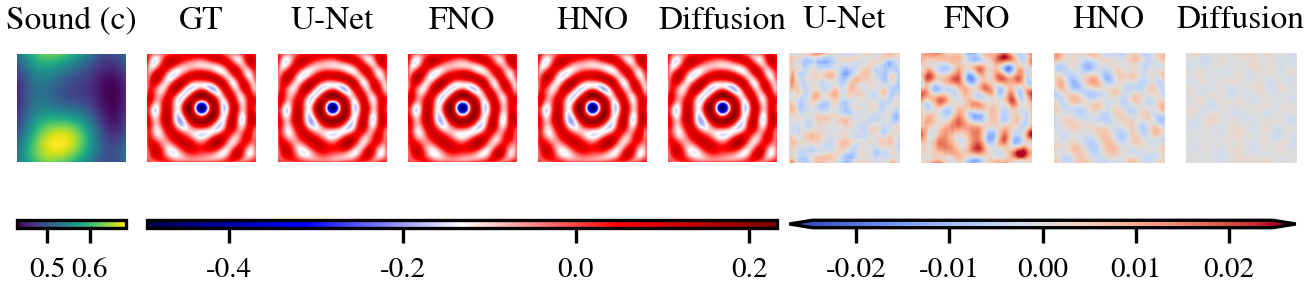}
  \end{subfigure}
  \vspace{4pt}

  \begin{subfigure}[t]{0.98\textwidth}
    \centering    \includegraphics[width=\linewidth,height=\rowH,keepaspectratio]{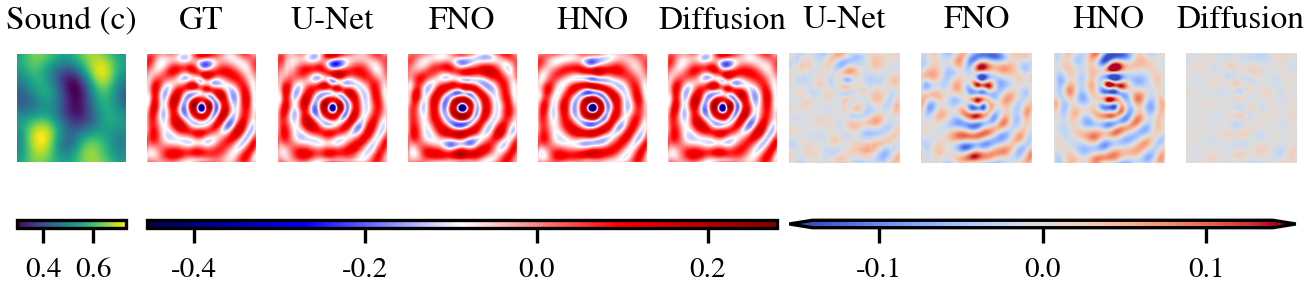}
  \end{subfigure}
  \vspace{4pt}
  
  \begin{subfigure}[t]{0.98\textwidth}
    \centering    \includegraphics[width=\linewidth,height=\rowH,keepaspectratio]{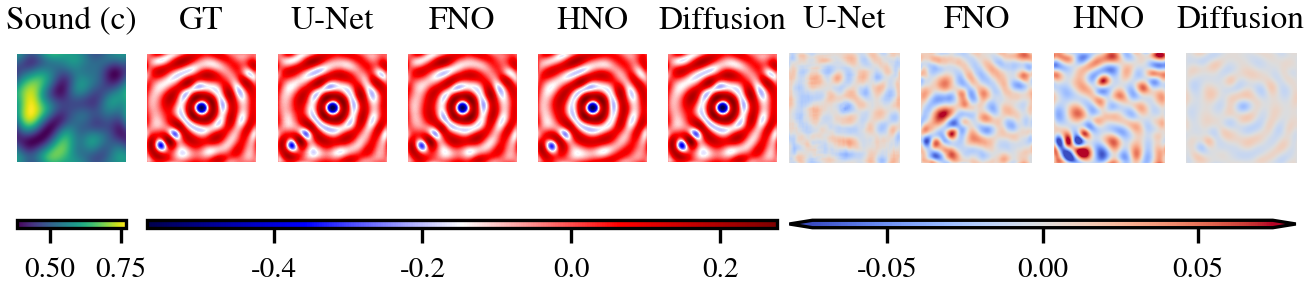}
    \vspace{2pt}
    \subcaption*{\small%
    \makebox[0pt][l]{\hspace*{0.22\linewidth}\emph{Prediction vs.\ Ground Truth}}%
    \makebox[\linewidth][l]{\hspace*{0.70\linewidth}\emph{Residual (Pred$-$GT)}}%
    }
    \vspace{-5pt}
  \end{subfigure}

  \caption{\textbf{Qualitative comparisons at $f=2.5\times10^{5}$ Hz.}
  }
  \label{fig:qual-supp-f2}
\end{figure}

\begin{figure}[t]
  \centering
  \captionsetup{font=small,skip=6pt}
  \captionsetup[subfigure]{labelformat=empty,justification=centering,aboveskip=2pt,belowskip=2pt}

  \setlength{\rowH}{0.20\textheight}

  \begin{subfigure}[t]{0.98\textwidth}
    \centering    \includegraphics[width=\linewidth,height=\rowH,keepaspectratio]{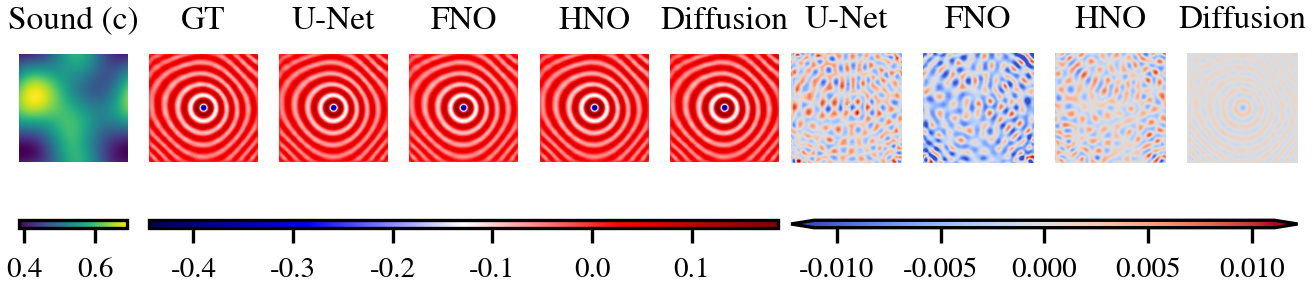}
  \end{subfigure}
  \vspace{4pt}
  
  \begin{subfigure}[t]{0.98\textwidth}
    \centering    \includegraphics[width=\linewidth,height=\rowH,keepaspectratio]{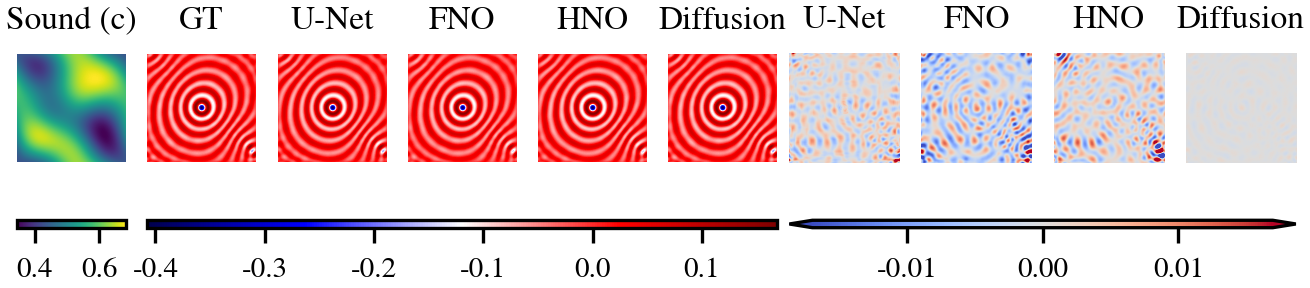}
  \end{subfigure}
  \vspace{4pt}

  \begin{subfigure}[t]{0.98\textwidth}
    \centering    \includegraphics[width=\linewidth,height=\rowH,keepaspectratio]{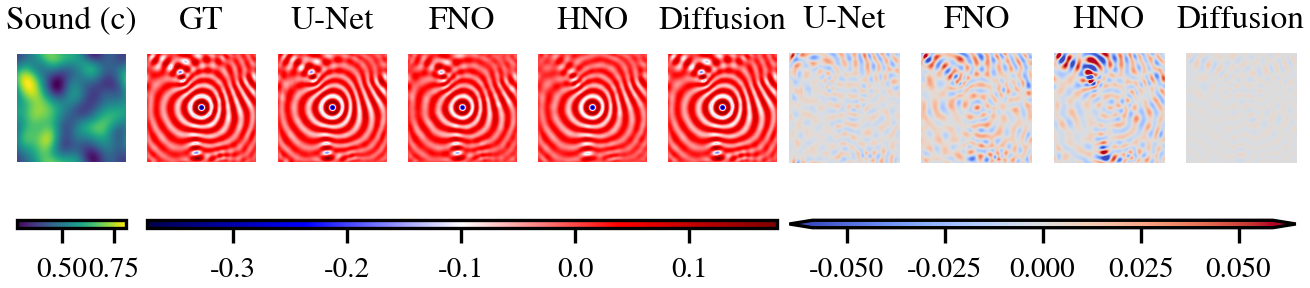}
  \end{subfigure}
  \vspace{4pt}

  \begin{subfigure}[t]{0.98\textwidth}
    \centering    \includegraphics[width=\linewidth,height=\rowH,keepaspectratio]{figs/additional_figures/qual/example_2d_5e5_idx356_with_sound.png}
  \end{subfigure}
  \vspace{4pt}

  \begin{subfigure}[t]{0.98\textwidth}
    \centering    \includegraphics[width=\linewidth,height=\rowH,keepaspectratio]{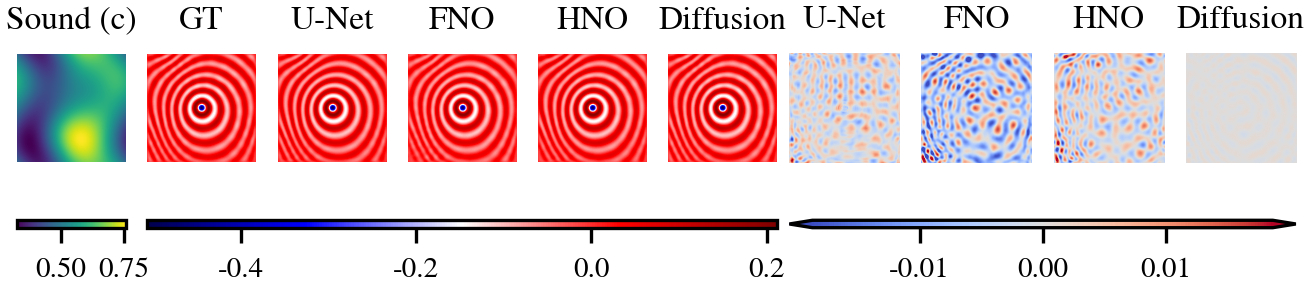}
  \end{subfigure}
  \vspace{4pt}

  \begin{subfigure}[t]{0.98\textwidth}
    \centering    \includegraphics[width=\linewidth,height=\rowH,keepaspectratio]{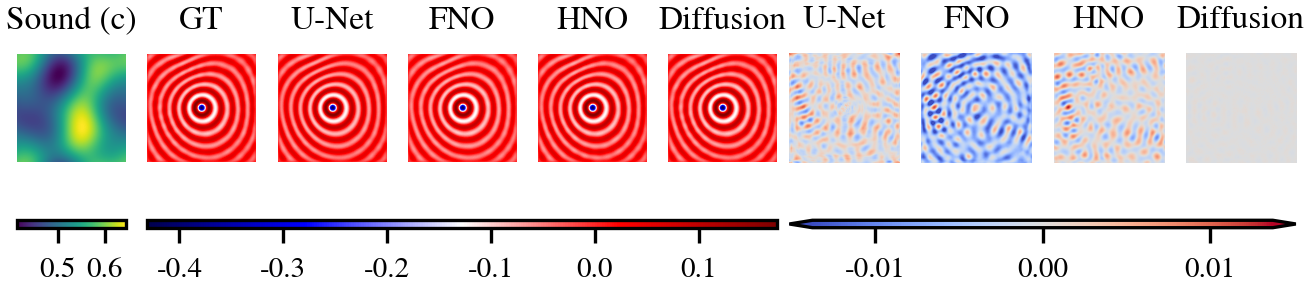}
  \end{subfigure}
  \vspace{4pt}
  
  \begin{subfigure}[t]{0.98\textwidth}
    \centering    \includegraphics[width=\linewidth,height=\rowH,keepaspectratio]{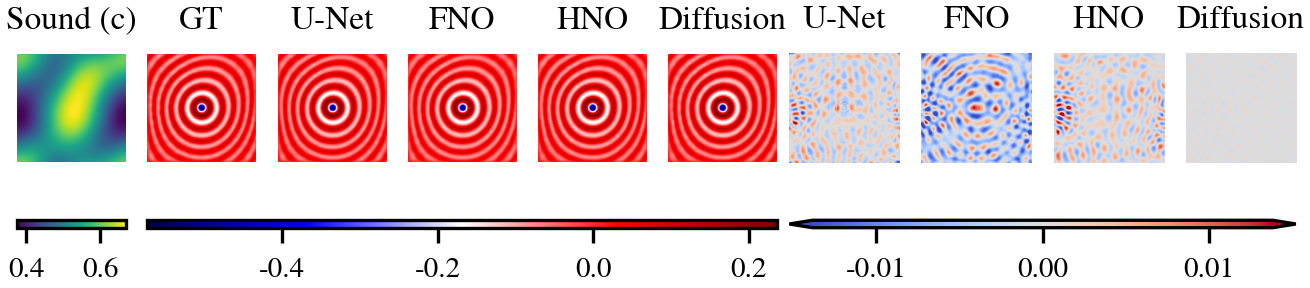}
    \vspace{2pt}
    \subcaption*{\small%
    \makebox[0pt][l]{\hspace*{0.22\linewidth}\emph{Prediction vs.\ Ground Truth}}%
    \makebox[\linewidth][l]{\hspace*{0.70\linewidth}\emph{Residual (Pred$-$GT)}}%
    }
    \vspace{-5pt}
  \end{subfigure}

  \caption{\textbf{Qualitative comparisons at $f=5\times10^{5}$ Hz.}
  }
  \label{fig:qual-supp-f3}
\end{figure}

\begin{figure}[t]
  \centering
  \captionsetup{font=small,skip=6pt}
  \captionsetup[subfigure]{labelformat=empty,justification=centering,aboveskip=2pt,belowskip=2pt}

  \setlength{\rowH}{0.20\textheight}

  \begin{subfigure}[t]{0.98\textwidth}
    \centering    \includegraphics[width=\linewidth,height=\rowH,keepaspectratio]{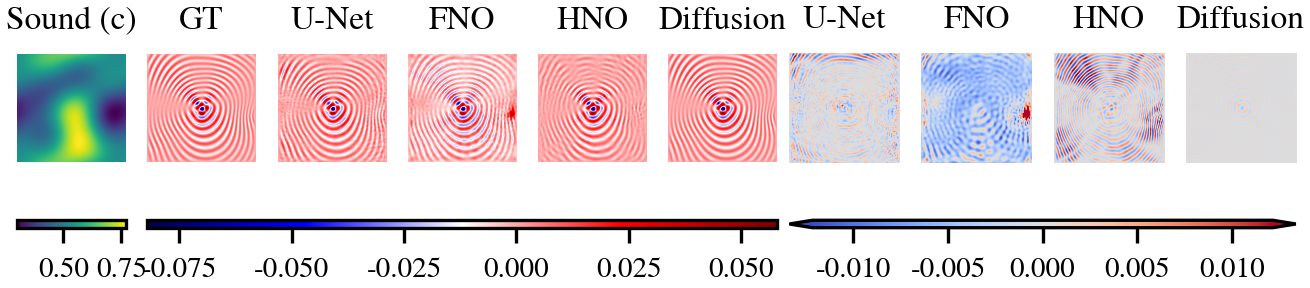}
  \end{subfigure}
  \vspace{4pt}
  
  \begin{subfigure}[t]{0.98\textwidth}
    \centering    \includegraphics[width=\linewidth,height=\rowH,keepaspectratio]{figs/additional_figures/qual/example_2d_1e6_idx167_with_sound.png}
  \end{subfigure}
  \vspace{4pt}

  \begin{subfigure}[t]{0.98\textwidth}
    \centering    \includegraphics[width=\linewidth,height=\rowH,keepaspectratio]{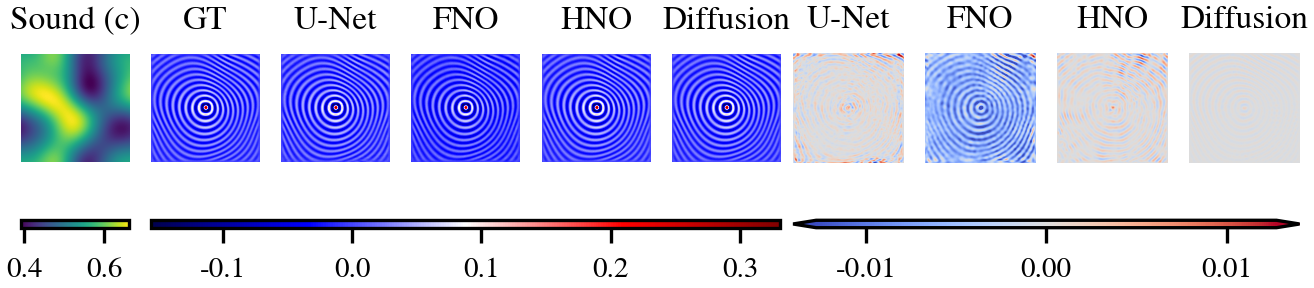}
  \end{subfigure}
  \vspace{4pt}

  \begin{subfigure}[t]{0.98\textwidth}
    \centering    \includegraphics[width=\linewidth,height=\rowH,keepaspectratio]{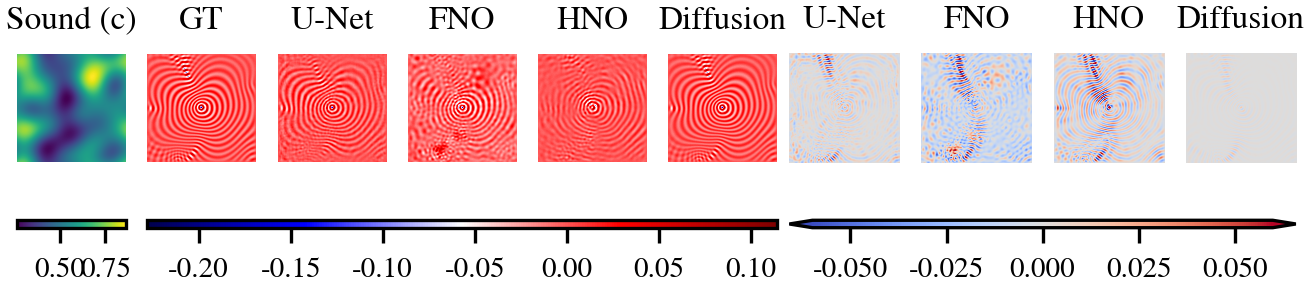}
  \end{subfigure}
  \vspace{4pt}

  \begin{subfigure}[t]{0.98\textwidth}
    \centering    \includegraphics[width=\linewidth,height=\rowH,keepaspectratio]{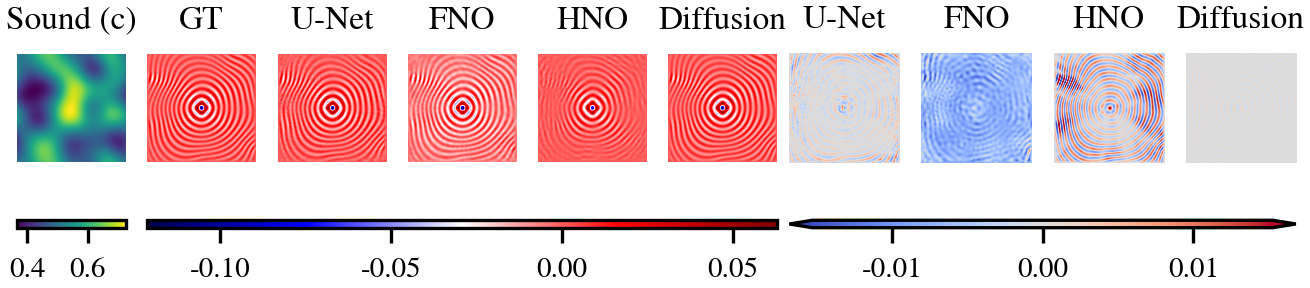}
  \end{subfigure}
  \vspace{4pt}

  \begin{subfigure}[t]{0.98\textwidth}
    \centering    \includegraphics[width=\linewidth,height=\rowH,keepaspectratio]{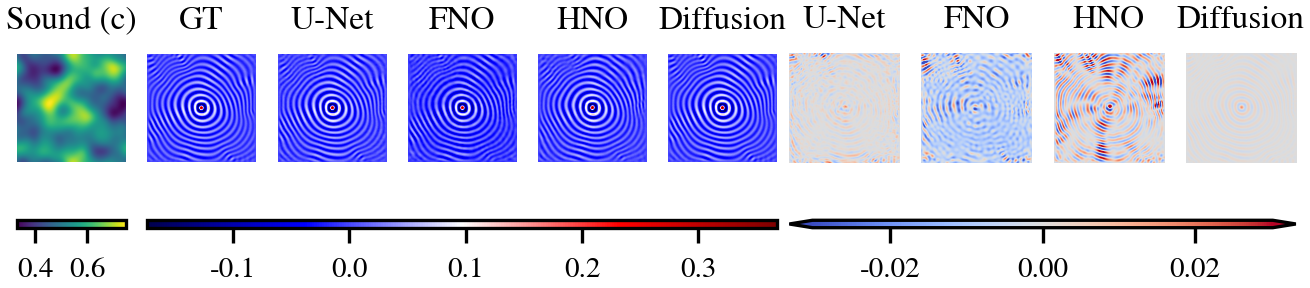}
  \end{subfigure}
  \vspace{4pt}
  
  \begin{subfigure}[t]{0.98\textwidth}
    \centering    \includegraphics[width=\linewidth,height=\rowH,keepaspectratio]{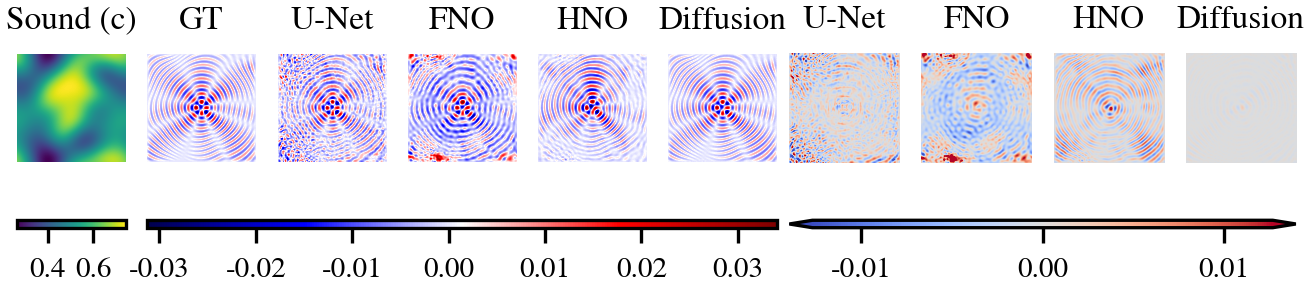}
    \vspace{2pt}
    \subcaption*{\small%
    \makebox[0pt][l]{\hspace*{0.22\linewidth}\emph{Prediction vs.\ Ground Truth}}%
    \makebox[\linewidth][l]{\hspace*{0.70\linewidth}\emph{Residual (Pred$-$GT)}}%
    }
    \vspace{-5pt}
  \end{subfigure}

  \caption{\textbf{Qualitative comparisons at $f=1\times10^{6}$ Hz.}
  }
  \label{fig:qual-supp-f4}
\end{figure}

\begin{figure}[t]
  \centering
  \captionsetup{font=small,skip=6pt}
  \captionsetup[subfigure]{labelformat=empty,justification=centering,aboveskip=2pt,belowskip=2pt}

  \setlength{\rowH}{0.20\textheight}

  \begin{subfigure}[t]{0.98\textwidth}
    \centering    \includegraphics[width=\linewidth,height=\rowH,keepaspectratio]{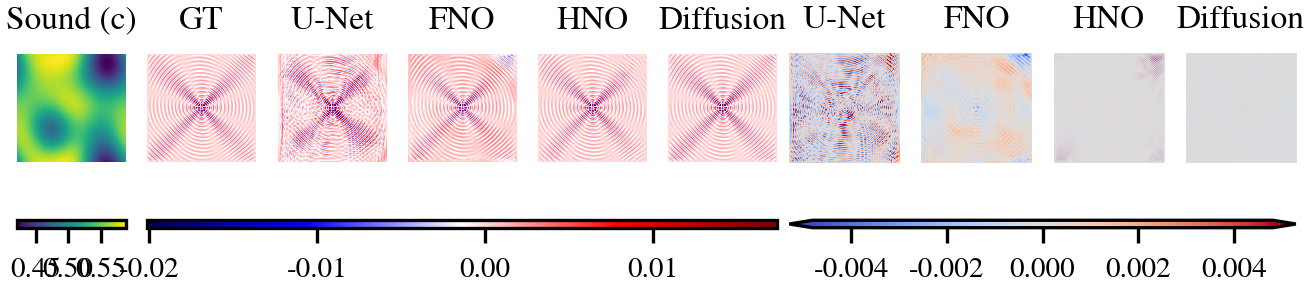}
  \end{subfigure}
  \vspace{4pt}
  
  \begin{subfigure}[t]{0.98\textwidth}
    \centering    \includegraphics[width=\linewidth,height=\rowH,keepaspectratio]{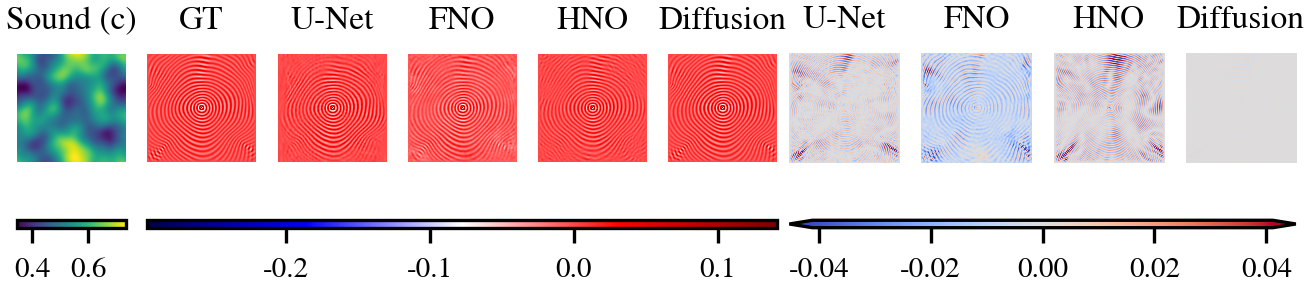}
  \end{subfigure}
  \vspace{4pt}

  \begin{subfigure}[t]{0.98\textwidth}
    \centering    \includegraphics[width=\linewidth,height=\rowH,keepaspectratio]{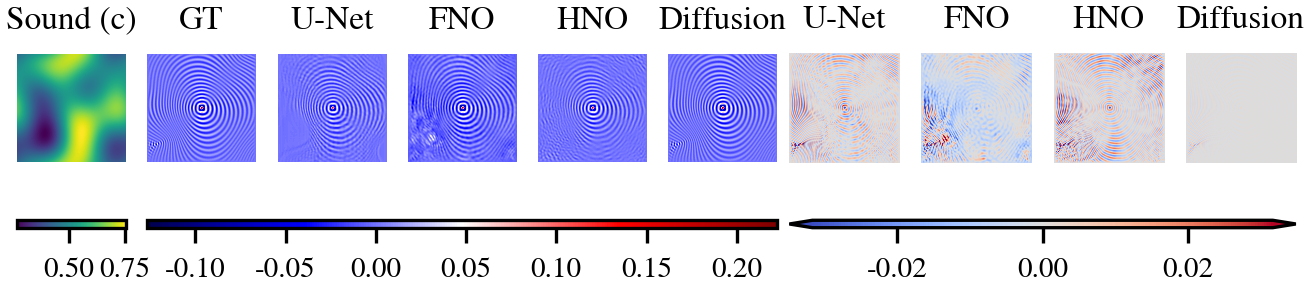}
  \end{subfigure}
  \vspace{4pt}

  \begin{subfigure}[t]{0.98\textwidth}
    \centering    \includegraphics[width=\linewidth,height=\rowH,keepaspectratio]{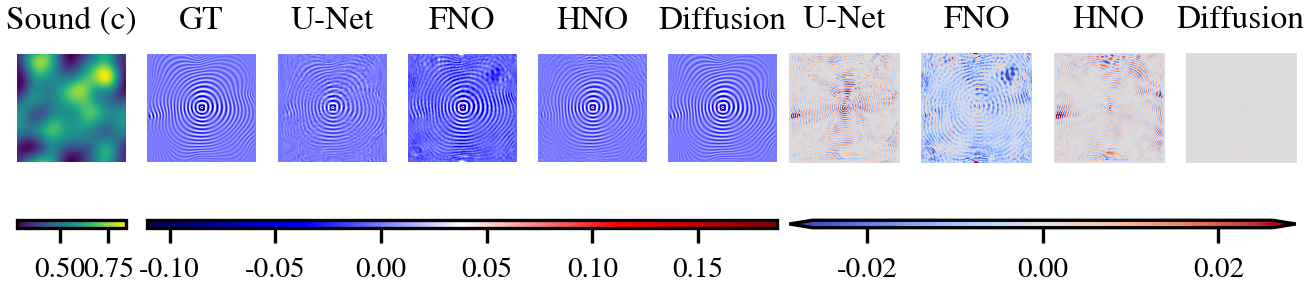}
  \end{subfigure}
  \vspace{4pt}

  \begin{subfigure}[t]{0.98\textwidth}
    \centering    \includegraphics[width=\linewidth,height=\rowH,keepaspectratio]{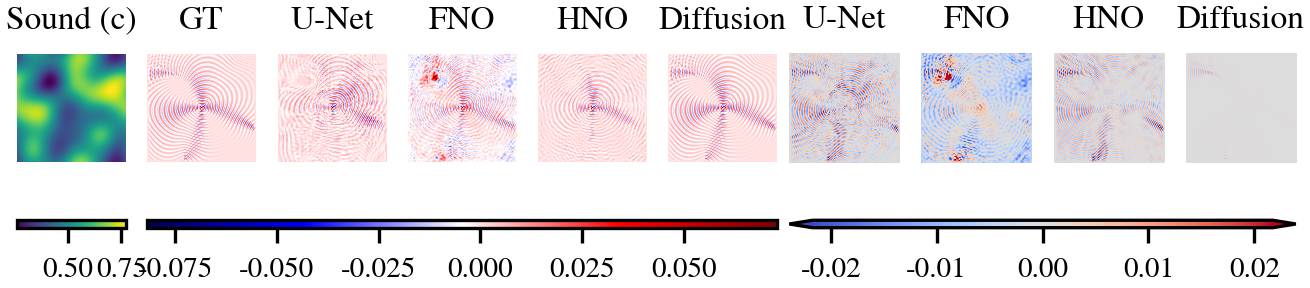}
  \end{subfigure}
  \vspace{4pt}

  \begin{subfigure}[t]{0.98\textwidth}
    \centering    \includegraphics[width=\linewidth,height=\rowH,keepaspectratio]{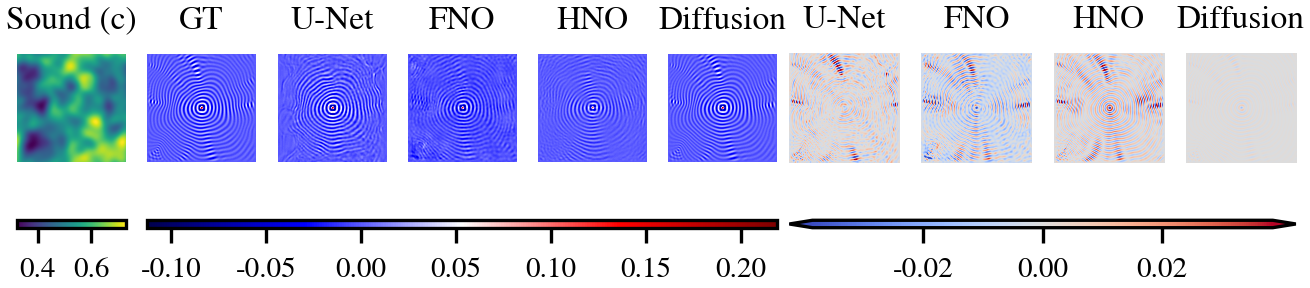}
  \end{subfigure}
  \vspace{4pt}
  
  \begin{subfigure}[t]{0.98\textwidth}
    \centering    \includegraphics[width=\linewidth,height=\rowH,keepaspectratio]{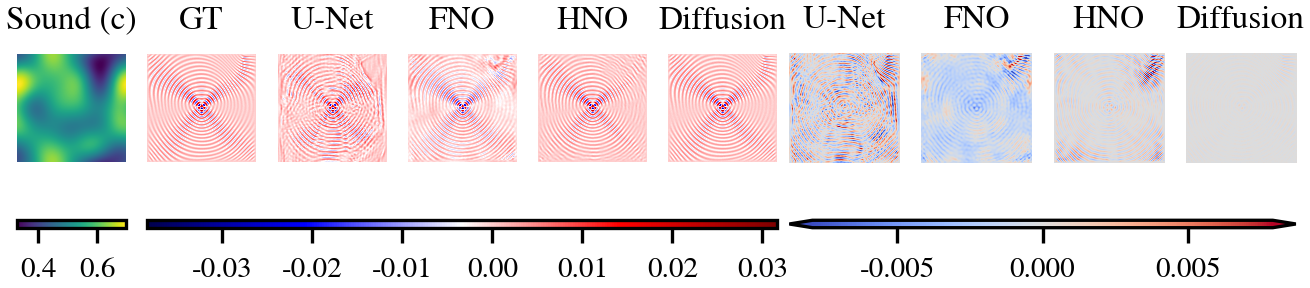}
    \vspace{2pt}
    \subcaption*{\small%
    \makebox[0pt][l]{\hspace*{0.22\linewidth}\emph{Prediction vs.\ Ground Truth}}%
    \makebox[\linewidth][l]{\hspace*{0.70\linewidth}\emph{Residual (Pred$-$GT)}}%
    }
    \vspace{-5pt}
  \end{subfigure}

  \caption{\textbf{Qualitative comparisons at $f=1.5\times10^{6}$ Hz.}
  }
  \label{fig:qual-supp-f5}
\end{figure}

\begin{figure}[t]
  \centering
  \captionsetup{font=small,skip=6pt}
  \captionsetup[subfigure]{labelformat=empty,justification=centering,aboveskip=2pt,belowskip=2pt}

  \setlength{\rowH}{0.20\textheight}

  \begin{subfigure}[t]{0.98\textwidth}
    \centering    \includegraphics[width=\linewidth,height=\rowH,keepaspectratio]{figs/additional_figures/qual/example_2d_2.5e6_idx132_with_sound.png}
  \end{subfigure}
  \vspace{4pt}
  
  \begin{subfigure}[t]{0.98\textwidth}
    \centering    \includegraphics[width=\linewidth,height=\rowH,keepaspectratio]{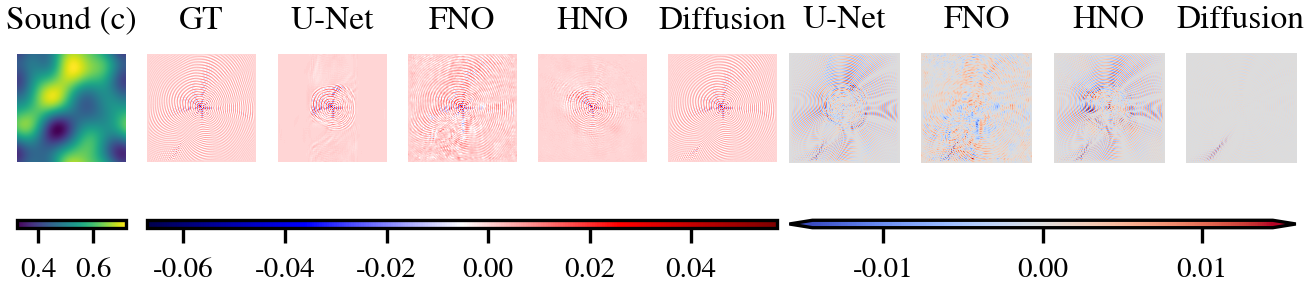}
  \end{subfigure}
  \vspace{4pt}

  \begin{subfigure}[t]{0.98\textwidth}
    \centering    \includegraphics[width=\linewidth,height=\rowH,keepaspectratio]{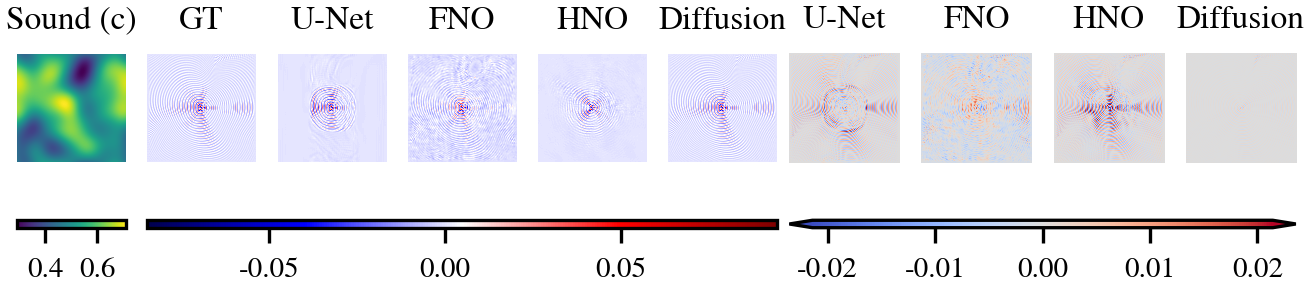}
  \end{subfigure}
  \vspace{4pt}

  \begin{subfigure}[t]{0.98\textwidth}
    \centering    \includegraphics[width=\linewidth,height=\rowH,keepaspectratio]{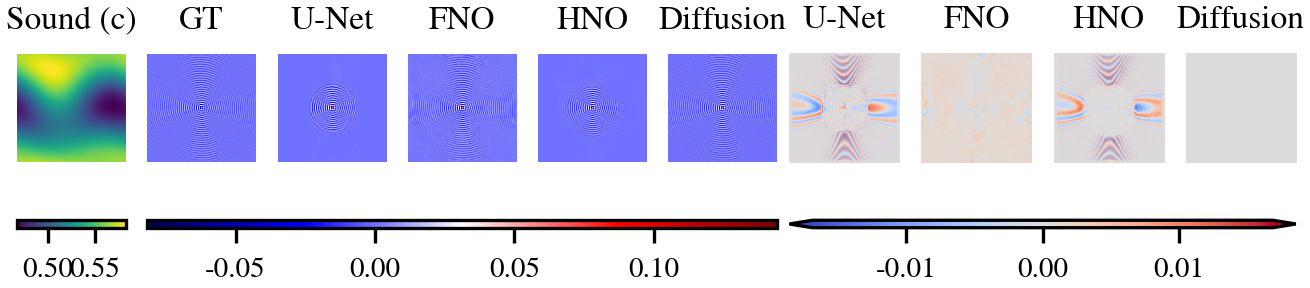}
  \end{subfigure}
  \vspace{4pt}

  \begin{subfigure}[t]{0.98\textwidth}
    \centering    \includegraphics[width=\linewidth,height=\rowH,keepaspectratio]{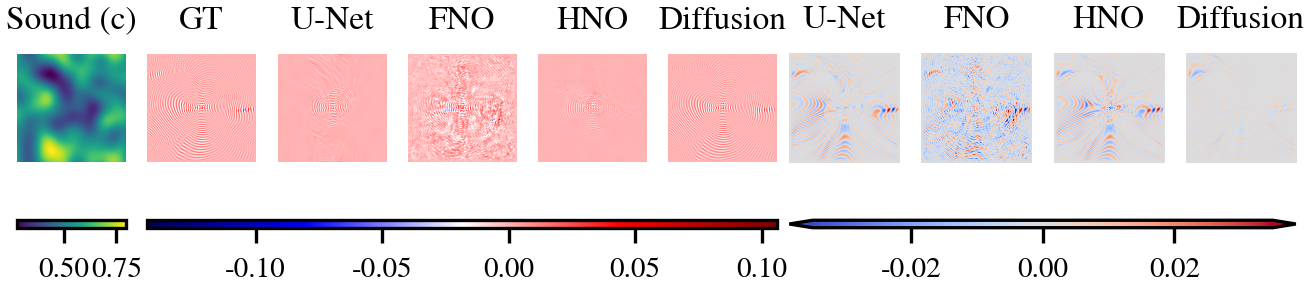}
  \end{subfigure}
  \vspace{4pt}

  \begin{subfigure}[t]{0.98\textwidth}
    \centering    \includegraphics[width=\linewidth,height=\rowH,keepaspectratio]{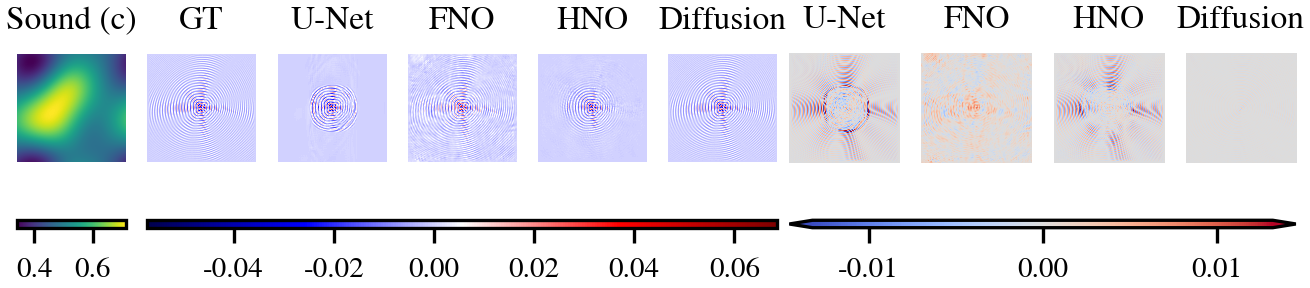}
  \end{subfigure}
  \vspace{4pt}
  
  \begin{subfigure}[t]{0.98\textwidth}
    \centering    \includegraphics[width=\linewidth,height=\rowH,keepaspectratio]{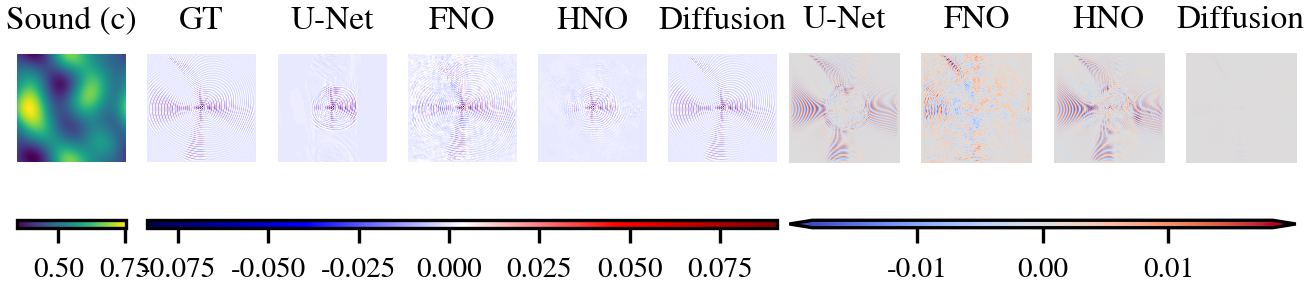}
    \vspace{2pt}
    \subcaption*{\small%
    \makebox[0pt][l]{\hspace*{0.22\linewidth}\emph{Prediction vs.\ Ground Truth}}%
    \makebox[\linewidth][l]{\hspace*{0.70\linewidth}\emph{Residual (Pred$-$GT)}}%
    }
    \vspace{-5pt}
  \end{subfigure}

  \caption{\textbf{Qualitative comparisons at $f=2.5\times10^{6}$ Hz.}
  }
  \label{fig:qual-supp-f6}
\end{figure}

Here we provide an expanded set of visual comparisons across all evaluated frequencies and multiple randomly drawn sound–speed maps (see App.~Figs.~\ref{fig:qual-supp-f1}–\ref{fig:qual-supp-f6}). For each case, we show GT, U\mbox{-}Net, FNO, HNO, and Diffusion predictions alongside per-pixel error maps. The trends observed in the main text persist: the Diffusion model consistently resolves interference patterns and high-frequency details with reduced artifacts, while deterministic operators exhibit smoothing and phase misalignment—especially in far-field regions and at higher $k$.

\clearpage

\clearpage

\subsection{Supplementary results}\label{app:supp-results}
This section provides additional figures that complement the main text. We include: (i) 1D sensitivity diagnostics that mirror the 2D setup, showing both direction-wise trajectories and distributional variability across directions; (ii) domain-averaged variance curves across frequencies to summarize global sensitivity as a function of interpolation level \(s\); (iii) qualitative sampler ablations at the highest frequency to illustrate step-budget effects; and (iv) low-frequency KDE comparisons at \(s{=}0\) to confirm that the same qualitative sensitivity trends persist beyond the highest-frequency setting.

\begin{figure}[!ht]
\begingroup
  \setlength{\abovecaptionskip}{2pt}
  \setlength{\belowcaptionskip}{0pt}
  \captionsetup{skip=2pt,font=small}

  \centering
  \includegraphics[width=\linewidth,height=0.25\textheight,
                   keepaspectratio]{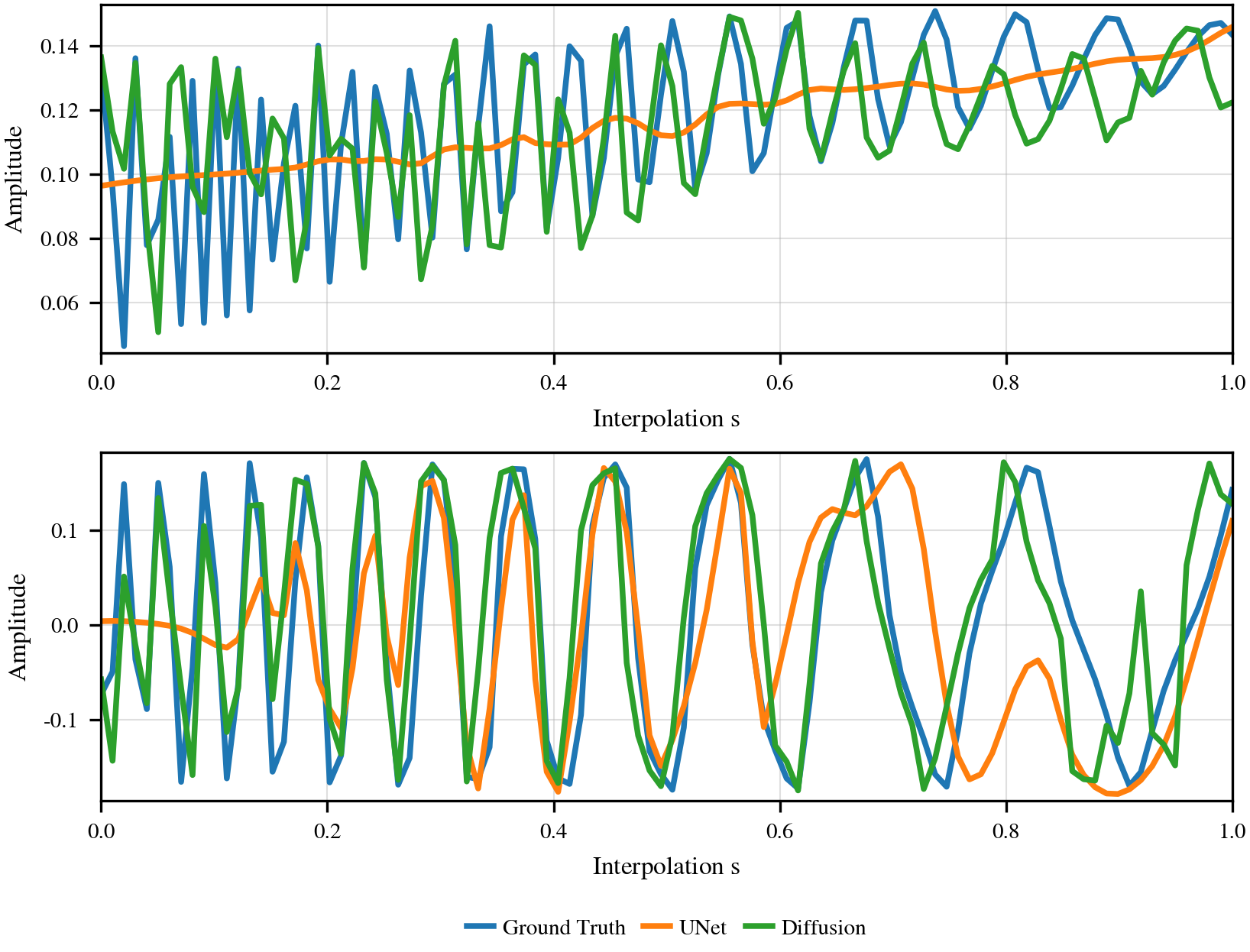}
  \caption{\textbf{Sampling along a linear path in coefficient space (1D).}
  Example curves from \(c_0\) to the target sound-speed map along direction \(d{=}10\) with \(s\!\in\![0,1]\) (100 samples).
  \emph{Top:} near the source. \emph{Bottom:} far from the source.}
  \label{fig:cross-sampling-1d}
\endgroup
\end{figure}

\begin{figure}[!ht]
\begingroup
  \setlength{\abovecaptionskip}{2pt}
  \setlength{\belowcaptionskip}{0pt}
  \captionsetup{skip=2pt,font=small}

  \centering
  \includegraphics[width=\linewidth,height=0.25\textheight,
                   keepaspectratio]{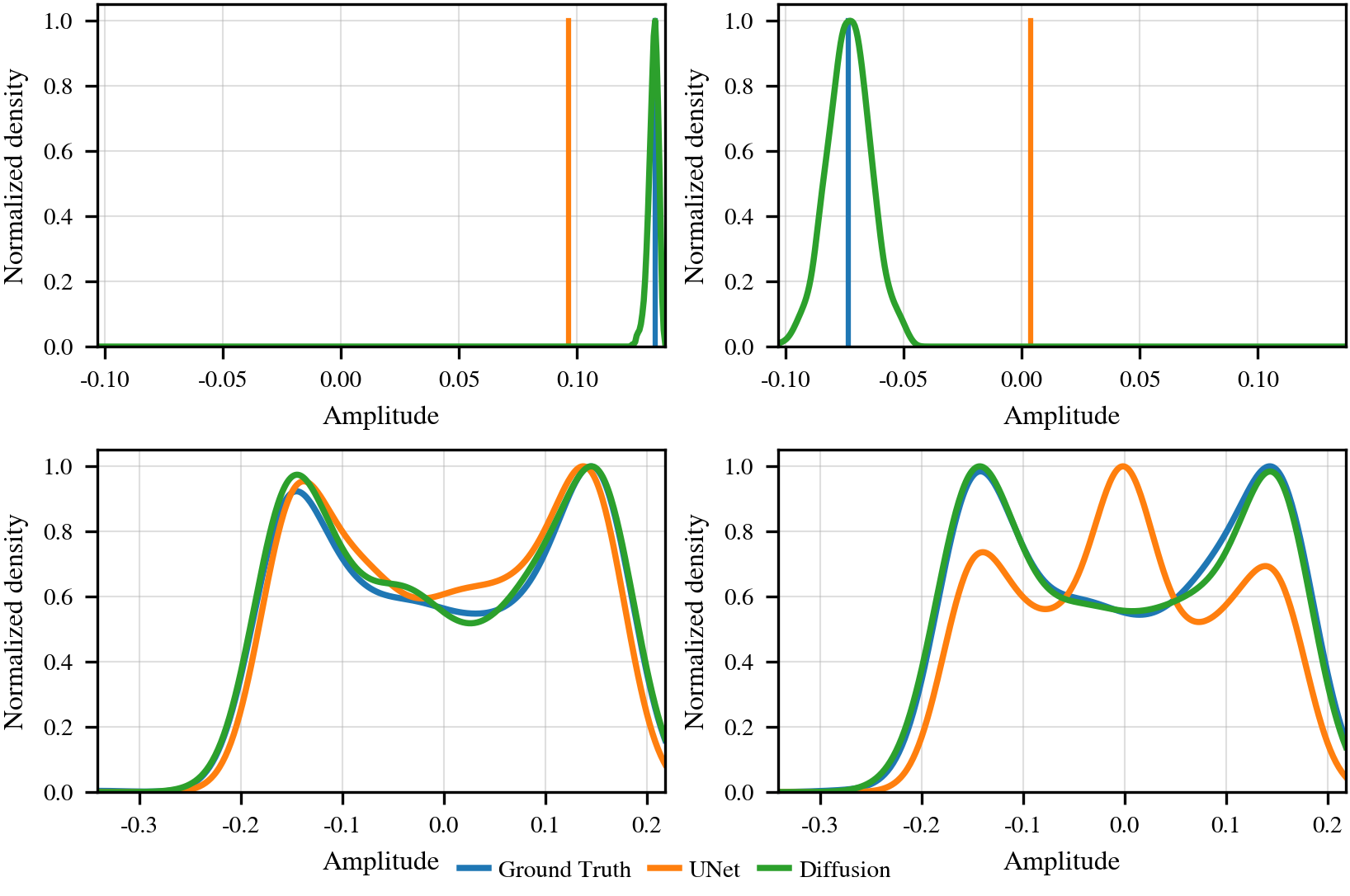}
  \caption{\textbf{Density plot (1D).}
  Kernel density estimates of \(\{\,u_M(s,d;y,x)\,\}_{d=1}^{100}\) at \(s=0\) (top) and \(s=1\) (bottom),
  shown for a \emph{near} point (left) and a \emph{far} point (right).
  At \(s=0\) all inputs equal \(c_0\), so deterministic models collapse to a spike, whereas diffusion exhibits calibrated spread; at \(s\approx1\), all methods broaden, with diffusion better capturing the high-variance regime.}
  \label{fig:density-1d}
\endgroup
\end{figure}

\begin{figure}[!ht]
\begingroup
  \setlength{\abovecaptionskip}{2pt}
  \setlength{\belowcaptionskip}{0pt}
  \captionsetup{skip=2pt}
  \captionsetup[sub]{aboveskip=1pt,belowskip=1pt}

  \centering
  \begin{subfigure}[t]{0.32\linewidth}
    \includegraphics[width=\linewidth,height=0.18\textheight,keepaspectratio]{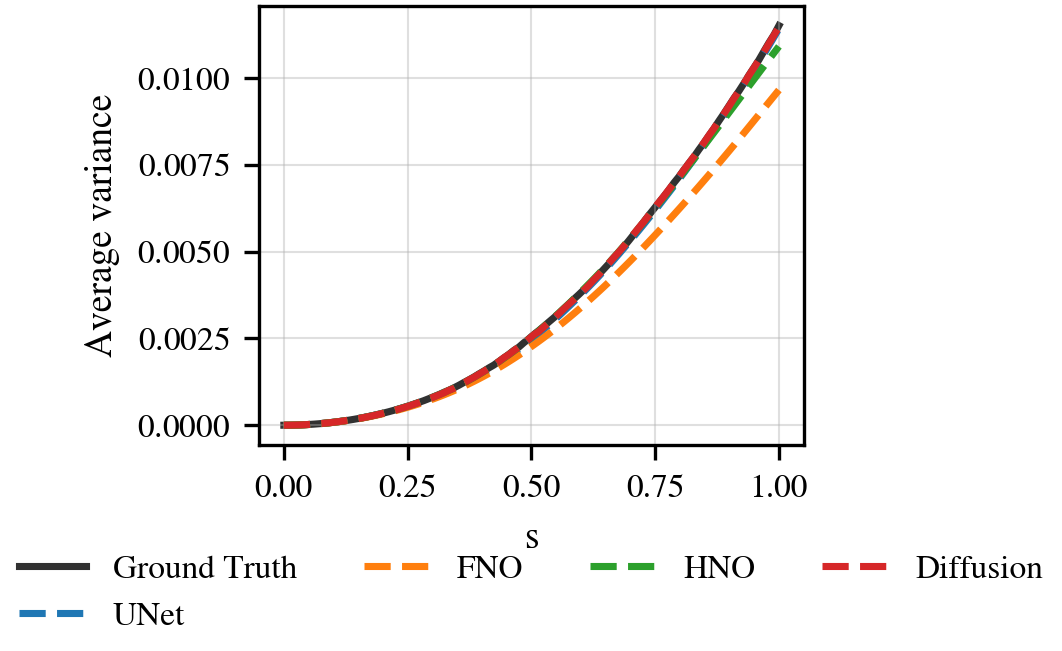}
    \caption*{\small $f=1.5\times10^{5}$ Hz}
  \end{subfigure}\hfill
  \begin{subfigure}[t]{0.32\linewidth}
    \includegraphics[width=\linewidth,height=0.18\textheight,keepaspectratio]{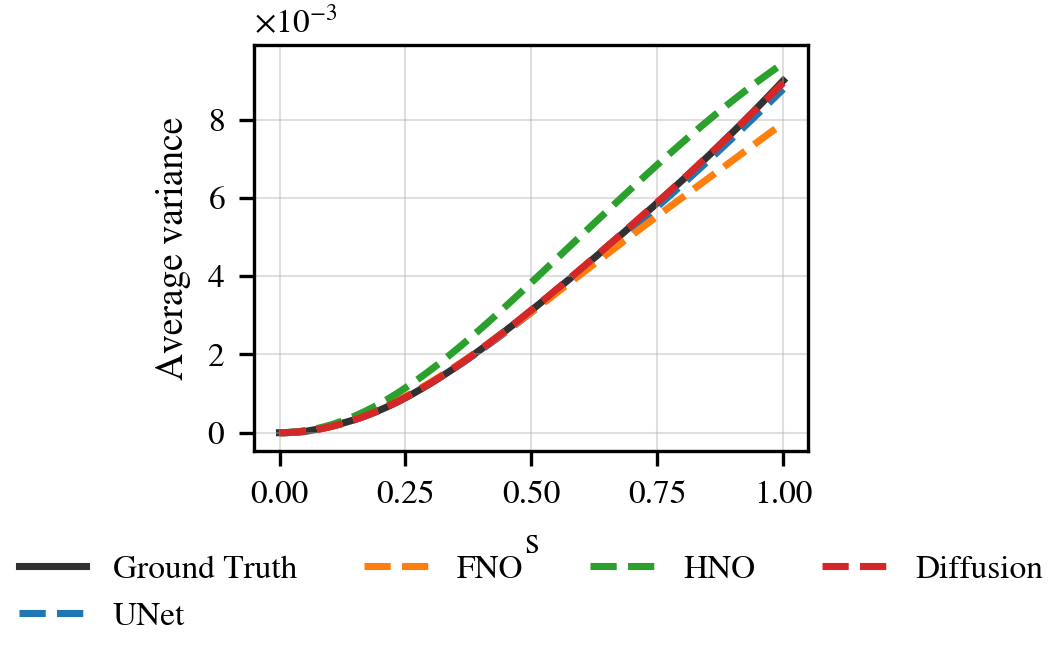}
    \caption*{\small $f=2.5\times10^{5}$ Hz}
  \end{subfigure}\hfill
  \begin{subfigure}[t]{0.32\linewidth}
    \includegraphics[width=\linewidth,height=0.18\textheight,keepaspectratio]{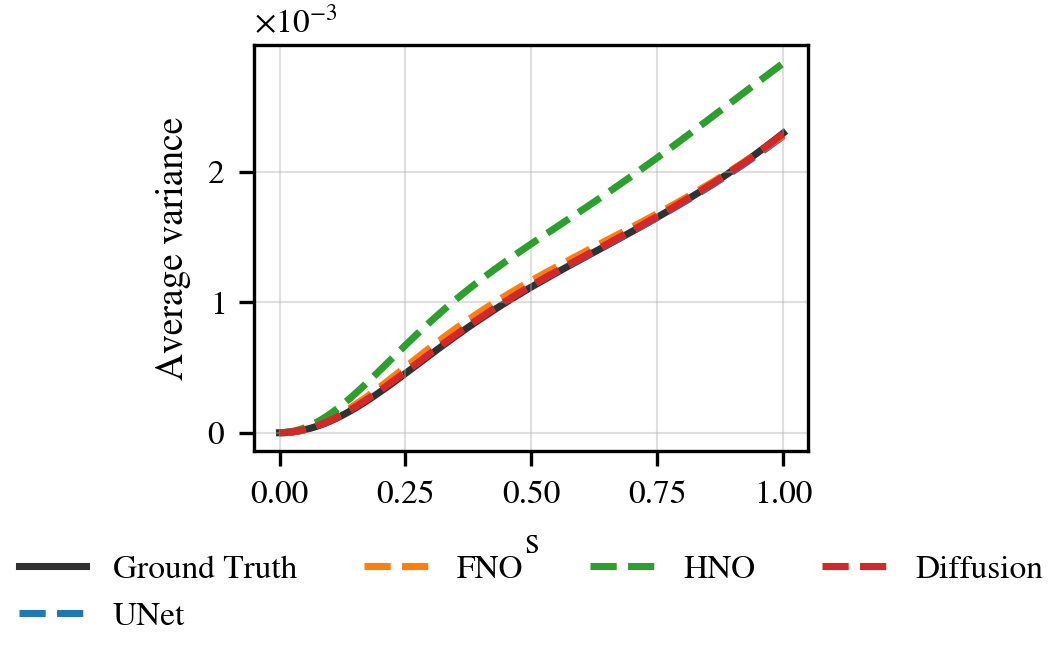}
    \caption*{\small $f=5\times10^{5}$ Hz}
  \end{subfigure}

  \vspace{2pt}

  \begin{subfigure}[t]{0.32\linewidth}
    \includegraphics[width=\linewidth,height=0.18\textheight,keepaspectratio]{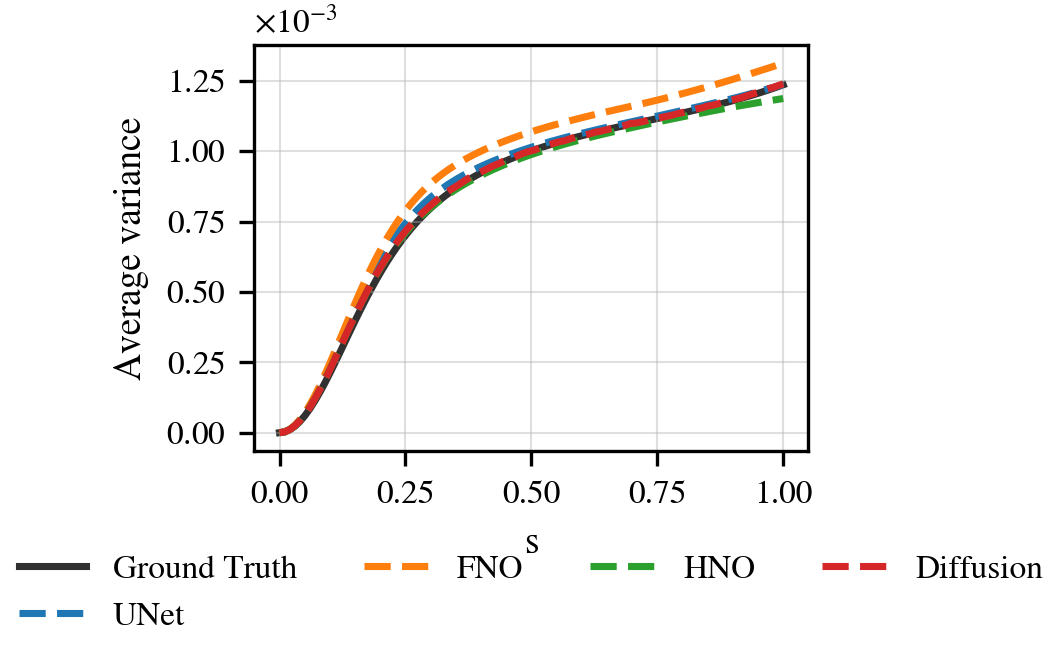}
    \caption*{\small $f=1\times10^{6}$ Hz}
  \end{subfigure}\hfill
  \begin{subfigure}[t]{0.32\linewidth}
    \includegraphics[width=\linewidth,height=0.18\textheight,keepaspectratio]{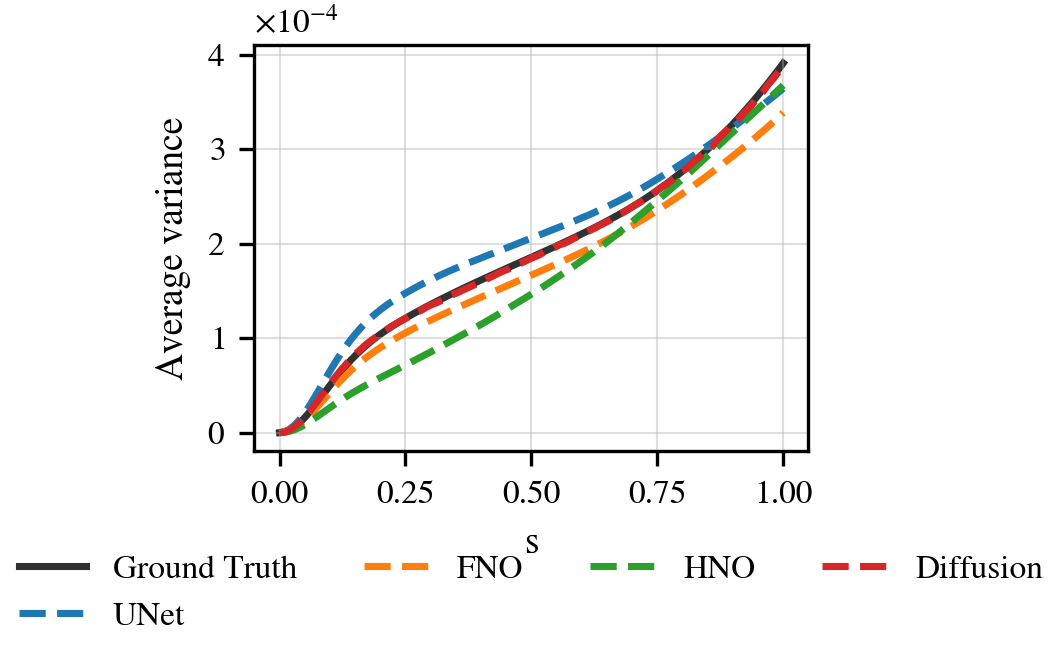}
    \caption*{\small $f=1.5\times10^{6}$ Hz}
  \end{subfigure}\hfill
  \begin{subfigure}[t]{0.32\linewidth}
    \includegraphics[width=\linewidth,height=0.18\textheight,keepaspectratio]{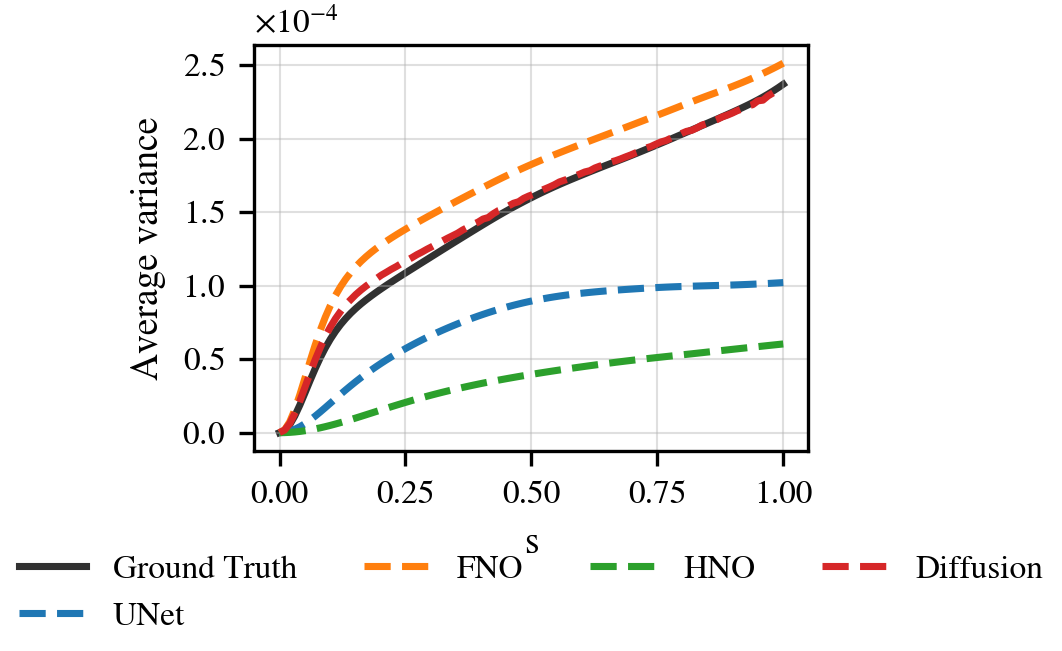}
    \caption*{\small $f=2.5\times10^{6}$ Hz}
  \end{subfigure}

  \caption{\textbf{Average directional variance vs.\ interpolation \(s\) across frequencies.}
  Each panel shows domain-averaged variance across 100 directions as a function of \(s\).}
  \label{fig:avg-var-vs-s}
\endgroup
\end{figure}

\begin{figure}[p]
  \centering
  \includegraphics[width=\linewidth,height=0.9\textheight,keepaspectratio]{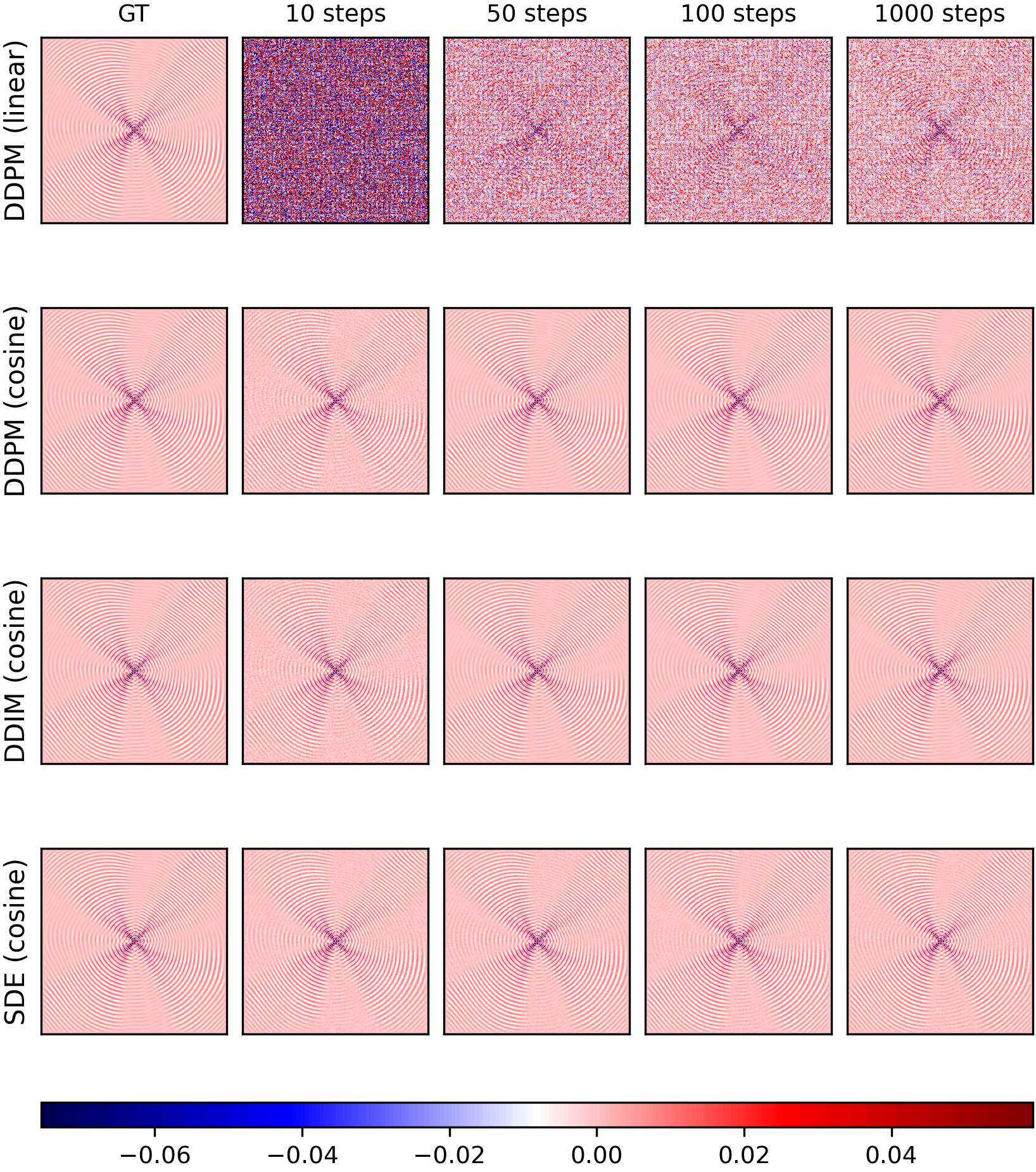}
  \caption{Comparison of sampler behavior across step budgets at $f = 2.5\times10^6$ Hz.}
  \label{fig:sampler_ablation_examples}
\end{figure}

\begin{figure}[t]
  \centering
  \begin{subfigure}[b]{0.7\linewidth}
    \centering
    \includegraphics[width=\linewidth]{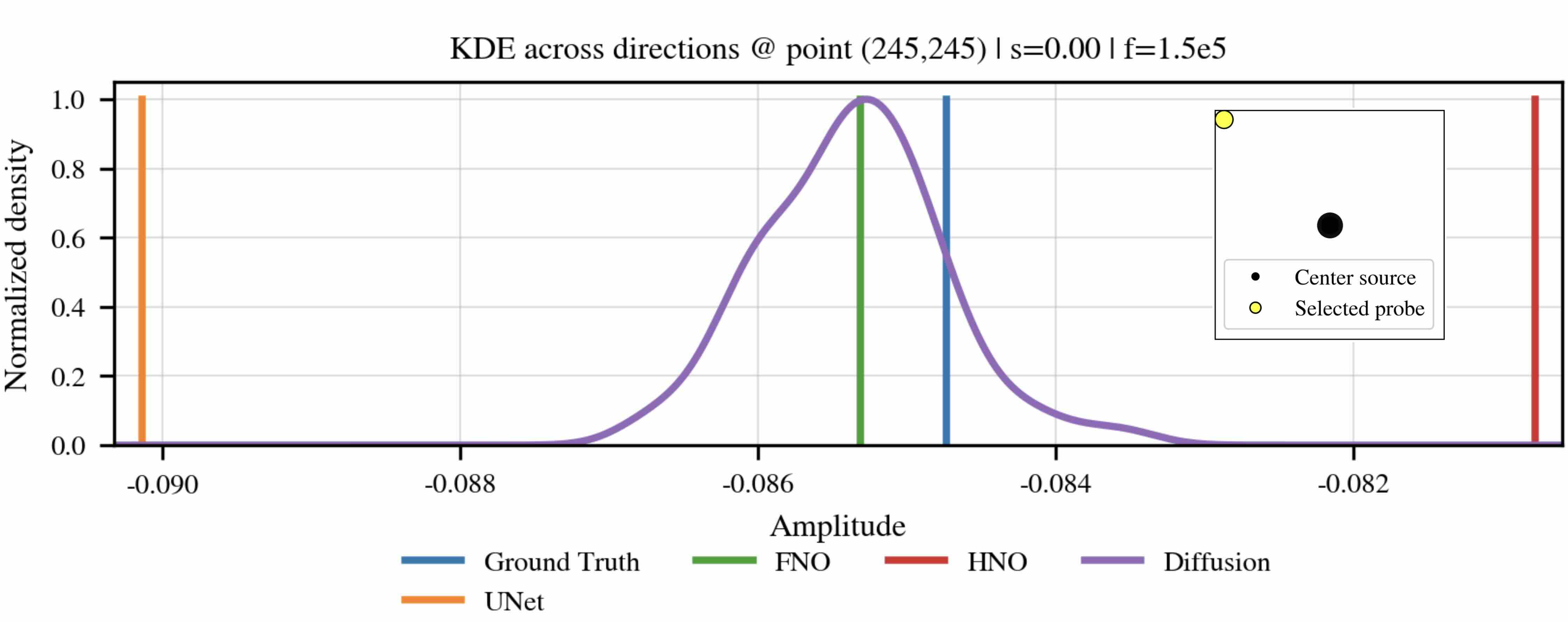}
    \caption{\(f = 1.5\times 10^{5}\,\text{Hz}\)}
  \end{subfigure}
  \vspace{0.3em}

  \begin{subfigure}[b]{0.7\linewidth}
    \centering
    \includegraphics[width=\linewidth]{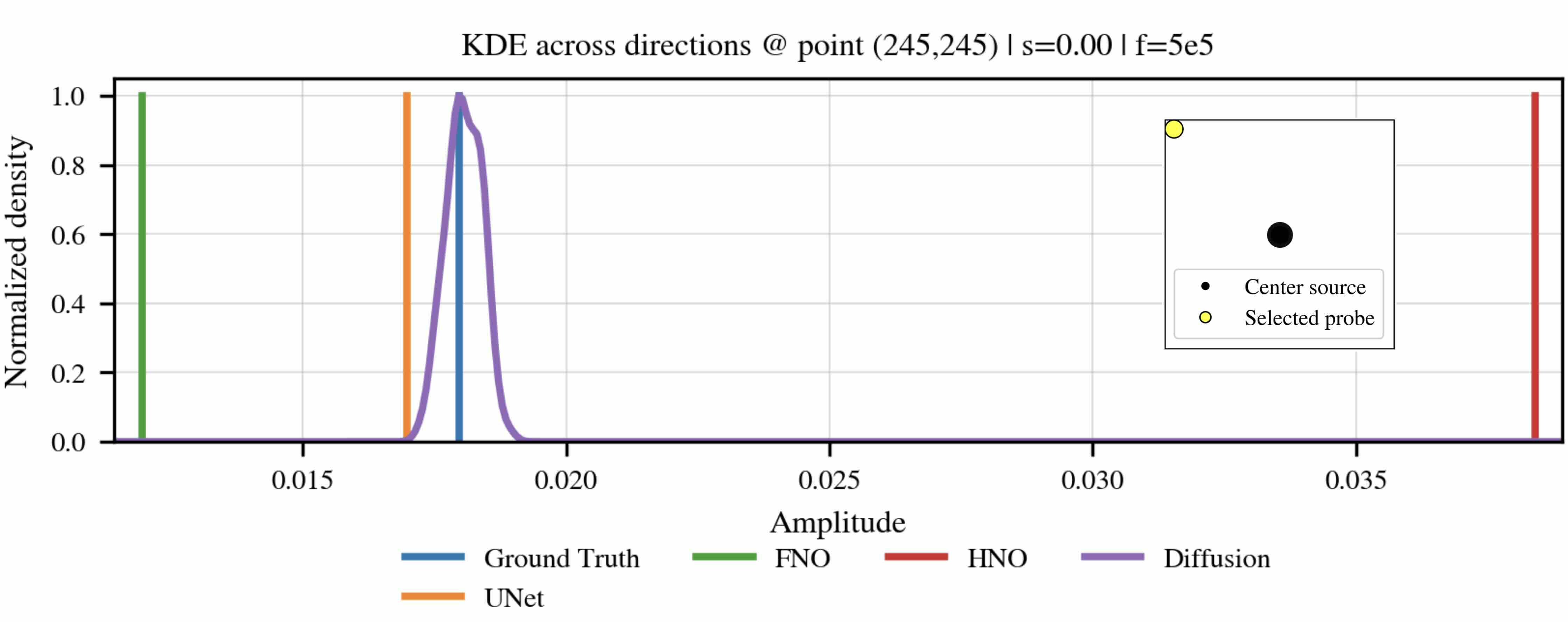}
    \caption{\(f = 5\times 10^{5}\,\text{Hz}\)}
  \end{subfigure}
  \vspace{0.3em}

  \begin{subfigure}[b]{0.7\linewidth}
    \centering
    \includegraphics[width=\linewidth]{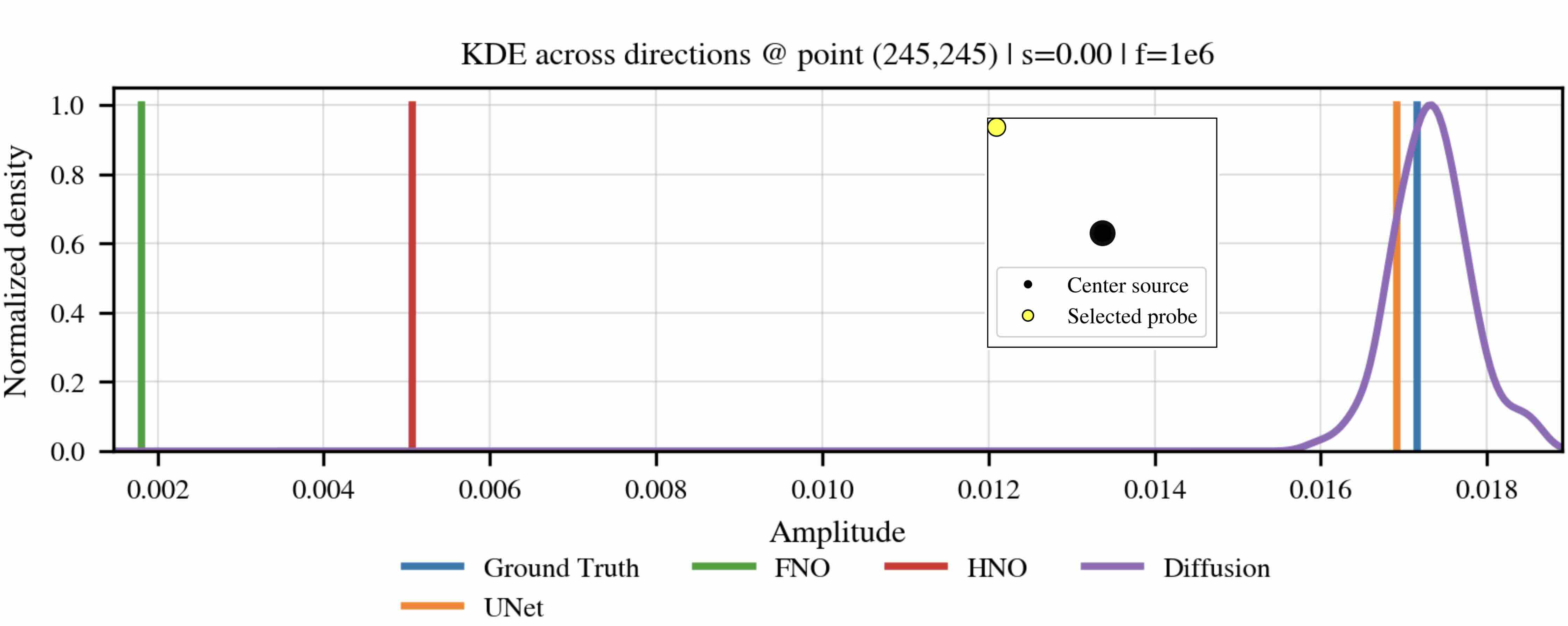}
    \caption{\(f = 10^{6}\,\text{Hz}\)}
  \end{subfigure}
  \vspace{0.3em}

  \begin{subfigure}[b]{0.7\linewidth}
    \centering
    \includegraphics[width=\linewidth]{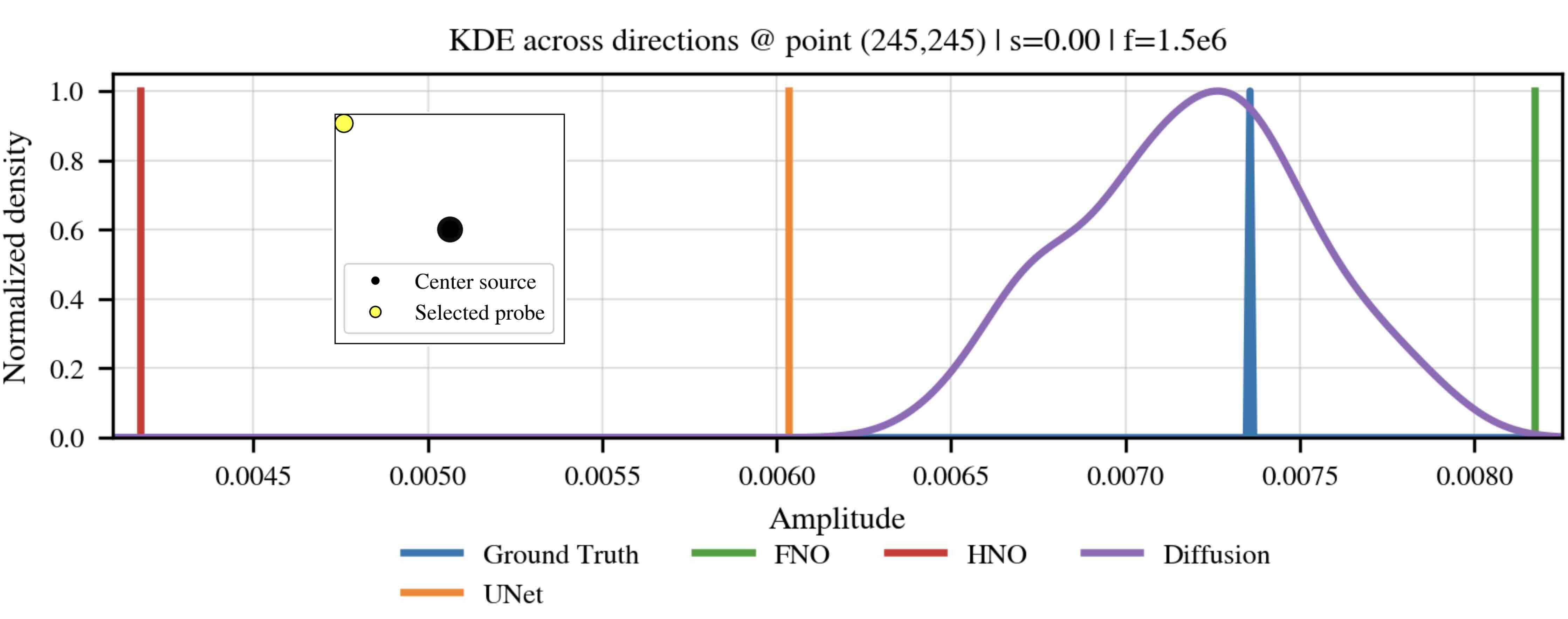}
    \caption{\(f = 1.5\times 10^{6}\,\text{Hz}\)}
  \end{subfigure}

  \caption{KDEs of sensitivity responses at \(s{=}0\) across GRF directions for four frequencies. At all frequencies, deterministic operators yield narrow, concentrated densities, while the diffusion model exhibits broader, multi-modal responses, indicating preserved sensitivity to small perturbations in the coefficient field.}
  \label{fig:kde-s0-lowfreq}
\end{figure}

Specifically, Fig.~\ref{fig:cross-sampling-1d} shows representative 1D cross-sampling curves along a single interpolation path (near vs.\ far probes), while Fig.~\ref{fig:density-1d} summarizes the corresponding variability across GRF directions via KDEs at selected \(s\)-values. Fig.~\ref{fig:avg-var-vs-s} reports the domain-averaged directional variance as a function of \(s\) for multiple frequencies, providing an aggregate sensitivity summary. Fig.~\ref{fig:sampler_ablation_examples} visualizes how sampler behavior changes with step budgets at \(f=2.5\times10^{6}\,\mathrm{Hz}\). Finally, Fig.~\ref{fig:kde-s0-lowfreq} shows KDEs at \(s{=}0\) across several lower frequencies, where deterministic operators remain overly concentrated compared to the diffusion model.

\begin{table}[!ht]
  \centering
  \scriptsize
  \setlength{\tabcolsep}{4pt}
  \caption{\textbf{1D relative errors vs.\ frequency.} Diffusion reports mean$\pm$std over $K{=}10$ samples.}
  \label{tab:1d_errors}
  \begin{tabular}{lcc|cc|cc}
    \toprule
    & \multicolumn{2}{c}{$L^2$} & \multicolumn{2}{c}{$H^1$} & \multicolumn{2}{c}{Energy} \\
    \cmidrule(lr){2-3}\cmidrule(lr){4-5}\cmidrule(lr){6-7}
    \textbf{Freq (Hz)} & \textbf{Diffusion} & \textbf{U\mbox{-}Net} & \textbf{Diffusion} & \textbf{U\mbox{-}Net} & \textbf{Diffusion} & \textbf{U\mbox{-}Net} \\
    \midrule
    $1.5\mathrm{e}5$ & \textbf{0.022}$\pm$0.0003 & 0.065 & \textbf{0.036}$\pm$0.0003 & 0.108 & \textbf{0.014}$\pm$0.0001 & 0.032 \\
    $2.5\mathrm{e}5$ & \textbf{0.030}$\pm$0.001  & 0.087 & \textbf{0.046}$\pm$0.001  & 0.130 & \textbf{0.010}$\pm$0.0002 & 0.025 \\
    $5\mathrm{e}5$   & \textbf{0.091}$\pm$0.002  & 0.174 & \textbf{0.114}$\pm$0.002 & 0.242 & \textbf{0.011}$\pm$0.0003 & 0.044 \\
    $7.5\mathrm{e}5$ & \textbf{0.093}$\pm$0.004  & 0.295 & \textbf{0.117}$\pm$0.004 & 0.389 & \textbf{0.018}$\pm$0.001 & 0.096 \\
    $1\mathrm{e}6$   & \textbf{0.215}$\pm$0.003  & 0.395 & \textbf{0.264}$\pm$0.004 & 0.510 & \textbf{0.050}$\pm$0.002 & 0.181 \\
    \bottomrule
  \end{tabular}
\end{table}

\FloatBarrier
\bibliographystyle{elsarticle-num}
\bibliography{references} 

\end{document}